%% file: 00-main.tex
\documentclass[runningheads,a4paper]{llncs}
\usepackage{geometry}
\usepackage{amssymb}
\usepackage{graphicx}
\usepackage{amsmath}
\usepackage{amssymb}
\usepackage{amsfonts}
\usepackage[mathscr]{euscript}
\usepackage{mathtools}
\usepackage{caption}
\usepackage{subfig}
\usepackage{url}
\usepackage{wrapfig}

\urldef{\mailsa}\path|payamsiyari@gatech.edu, dilkina@usc.edu, |
\urldef{\mailsb}\path|constantine@gatech.edu|

\begin{document}

\mainmatter  

\title{Emergence and Evolution of Hierarchical Structure in Complex Systems}

\titlerunning{Emergence and Evolution of Hierarchical Structure in Complex Systems}

%
%
\author{Payam Siyari*%
\and Bistra Dilkina** \and Constantine Dovrolis*
\\
\mailsa
\\
\mailsb \emph{(Corresponding Author)}}
\authorrunning{P.\@ Siyari, B.\@ Dilkina, C.\@ Dovrolis}

\institute{$*$ School of Computer Science, Georgia Institute of Technology, Atlanta, GA\\
$**$ Department of Computer Science, University of Southern California, Los Angeles, CA
}

%
%

\toctitle{Dynamics on and of Complex Networks}
\maketitle

\begin{abstract}
It is well known that many complex systems, both in technology and nature, exhibit hierarchical modularity: smaller modules, each of them providing a certain function, are used within larger modules that perform more complex functions. What is not well understood however is how this hierarchical structure (which is fundamentally a network property) emerges, 
and how it evolves over time. 
\newline\indent
We propose a modeling framework, referred to as Evo-Lexis, that provides insight to some fundamental questions about evolving hierarchical systems. Evo-Lexis models the most elementary modules of the system as symbols (``sources'') and the modules at the highest level of the hierarchy as sequences of those symbols (``targets''). Evo-Lexis computes the optimized adjustment of a given hierarchy when the set of targets changes over time by additions and removals (a process referred to as ``incremental design'').
\newline\indent
In this paper we use computation modeling to show that:
\begin{itemize}
    \item 
    Low-cost and deep hierarchies emerge when the population of target sequences evolves through tinkering and mutation. 
    \item
    Strong selection on the cost of new candidate targets results in reuse of more complex (longer) nodes in an optimized hierarchy.
    \item
    The bias towards reuse of complex nodes results in an ``hourglass architecture'' (i.e., few intermediate nodes that cover almost all source-target paths).
    \item
    With such bias, the core nodes are conserved for relatively long time periods although still being vulnerable to major transitions and punctuated equilibria.
    \item
    Finally, we analyze the differences in terms of cost and structure between incrementally designed hierarchies and the corresponding ``clean-slate'' hierarchies which result when the system is designed from scratch after a change.
\end{itemize}
\end{abstract}

\input{01-introduction}
\clearpage
\input{03-lexis-background}
\input{04-evo-modeling}
\input{05-target-models}
\clearpage
\input{06-evolvability}
\input{07-transitions}
\input{08-overhead}
\input{02-related-work}
\input{09-conclusion}

\clearpage
\bibliographystyle{plain}

\input{00-main.bbl}
%
%
%
%
%
%
%
\end{document}

%% file: 01-introduction.tex
\section{Introduction}
It is well known that many complex systems, both in technology and nature, exhibit modularity: independent modules, each of them providing a certain function, are combined together to perform more complex functions \cite{Baldwin}. Additionally,  modular systems are also organized in a hierarchical way: smaller modules are used within larger modules recursively \cite{Ravasz}.  Examples of such systems exist in a wide range of environments: in natural systems, it is believed that hierarchical modularity enhances evolvability (the ability of the system to adapt to new environments with minimal changes) and robustness (the ability to maintain the current status in the presence of internal or external variations) \cite{clune2016,sabrin2016}. In the technological world, hierarchically modular designs are preferred in terms of design and development cost, easier maintenance and agility (e.g. less effort in producing future versions of a software), and better abstraction of the system design \cite{myers}.

There are many hypotheses in the literature regarding the factors that contribute to either the hierarchy or modularity properties. Local resource constraints in social networks and ecosystems \cite{Miller2008}, modularly varying goals \cite{clune2012,kashtan2005,kashtan2007}, selection for more robust phenotypes \cite{clune-robust-pheno2,clune-robust-pheno1}, and selection for lower connection costs in a network \cite{clune2016} are some of the mechanisms that have been previously explored and shown to lead to hierarchically modular systems. The main hypothesis that we follow in this paper is along the lines of \cite{clune2016},  which assumes that systems in both nature and technology care to minimize the cost of their interconnections or dependencies between modules.

\begin{figure}[h]
\centering
\includegraphics[trim = 0cm 0cm 0cm 0cm,clip,width=.7\textwidth]{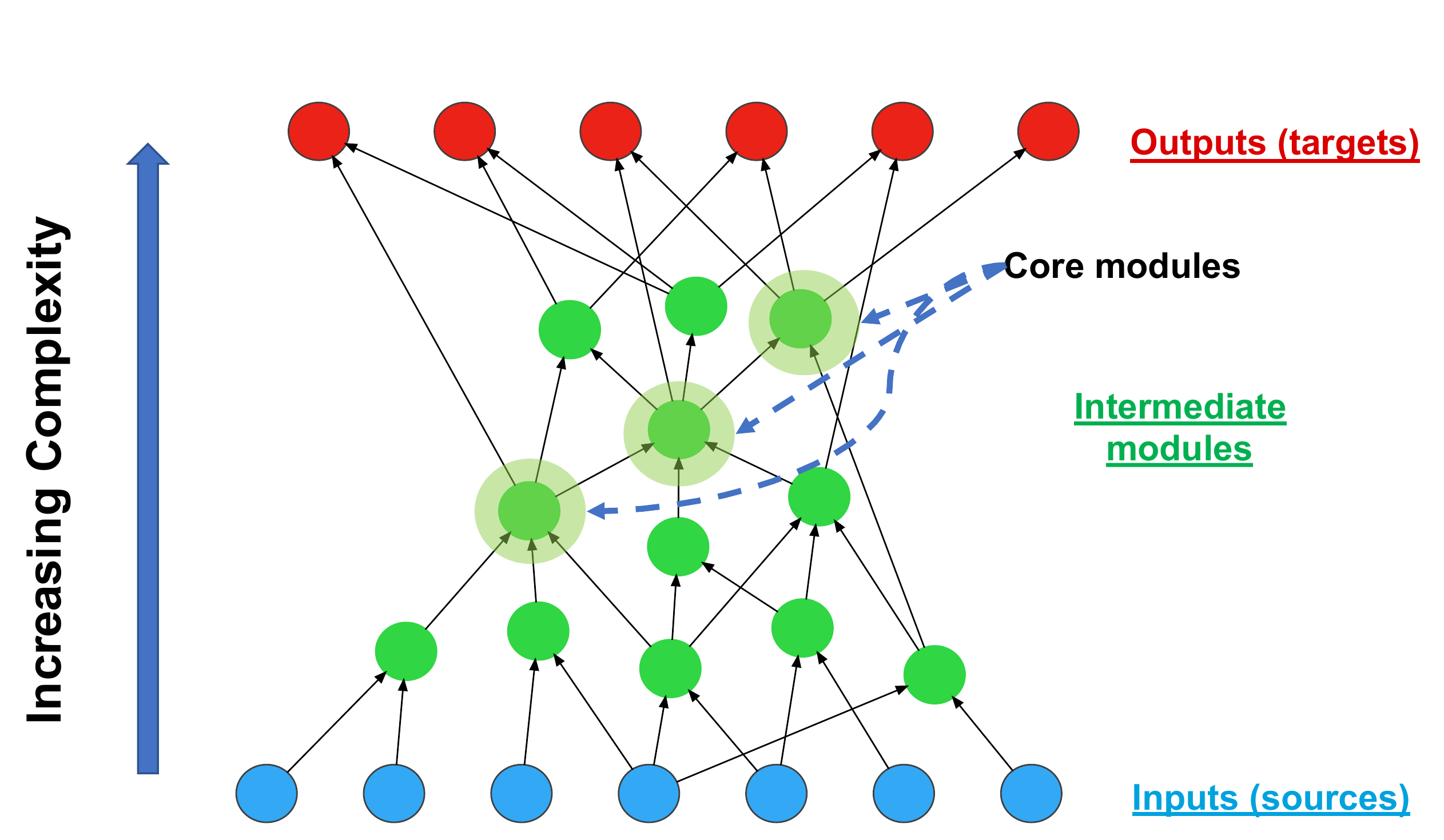}
\caption{A hierarchical system is represented as a directed-acyclic graph in which each module is shown as a node, and the dependencies from more elementary modules to more complex modules are shown as upward edges. The hourglass effect occurs when the system of interest produces many outputs from many inputs through a relatively small number of intermediate core modules (here, highlighted nodes with transparent surroundings) \cite{sabrin2016}.
}
\label{fig:hourglass}
\end{figure}

An additional focus of our work is the hourglass effect in hierarchical systems. Across many fields, such as 
 in computer networking \cite{akhshabi2011}, deep neural networks \cite{hourglass-dnn}, embryogenesis \cite{hourglass-embryo}, metabolism \cite{hourglass-metabolism},  and many others \cite{sabrin2016}, it has been observed that hierarchically modular systems often exhibit the architecture of an  hourglass. Informally, an hourglass architecture means that the system of interest produces many outputs from many inputs through a relatively small number of highly central intermediate modules, referred to as the ``waist'' of the hourglass (Fig. \ref{fig:hourglass}). The waist of the hourglass (also referred to as ``core'' in \cite{sabrin2016} as well as in this paper) includes critical modules of the system that are also sometimes more conserved during the evolution of the system compared to other modules \cite{akhshabi2011,sabrin2016}. Despite recent research on the hourglass effect in different types of hierarchical systems \cite{akhshabi2011,akhshabi2014,alon2015,sabrin2016}, one of the questions that is still open is to identify the conditions under which the hourglass effect emerges in hierarchies that are produced when the objective is to minimize the cost of interconnections.

In this paper, we present \emph{Evo-Lexis}, a modeling framework for the emergence and evolution of hierarchical structure in complex systems. To develop \emph{Evo-Lexis}, we extend a previously proposed optimization framework, called \emph{Lexis} \cite{siyari2016a}, that was designed for structure discovery in sequential data. Lexis models the most elementary modules of the system as symbols (``sources'') and the modules at the highest level of the hierarchy as sequences of those symbols (``targets''). \emph{Evo-Lexis} is a dynamic or evolving version of Lexis, in the sense that the set of targets changes over time through additions (births) and removals (deaths) of targets. \emph{Evo-Lexis} computes an (approximate) minimum-cost adjustment of a given hierarchy when the set of targets changes over time (a process we refer to as ``incremental design''). For comparison purposes, \emph{Evo-Lexis} also computes the (approximate) minimum-cost hierarchy that generates a given set of targets from a set of sources in a static (non-evolving) setting (referred to as ``clean-slate design''). The premise behind the incremental design approach is that in practice systems are rarely designed from scratch -- instead,  they are incrementally modified over time to accommodate the changes (e.g. provide new outputs and potentially to support new inputs every time there is a change).

In general, a system interacts with its environment in a bidirectional manner: the environment imposes various constraints on the system and the system also affects its environment. To capture this co-evolutionary setting in \emph{Evo-Lexis}, we study how changes in the set of targets affect the resulting hierarchy but also how the current hierarchy affects the selection of new targets (i.e. whether a new candidate target is selected or not depends on its fitness or cost -- and that depends on how easily that target can be supported by the given hierarchy). By incorporating well-known evolutionary mechanisms, such as tinkering (mutation), recombination, and selection, \emph{Evo-Lexis} can capture such co-evolutionary dynamics between the generation of new targets and the hierarchy that supports them.

The questions we focus on are:
\begin{enumerate}
\item How do key properties of the emergent hierarchies, e.g. depth of the network, reuse or centrality of each module, complexity (or sequence length) of intermediate modules, etc., depend on the evolutionary process that generates the new targets of the system?
\item
Under what conditions do the emergent hierarchies exhibit the so called ``hourglass effect''? Why are few intermediate modules reused much more than others? 
\item
Do intermediate modules persist during the evolution of hierarchies? Or are there ``punctuated equilibria'' where the  highly reused modules change significantly?
\item
Which are the differences in terms of cost and structure between the incrementally designed and the corresponding clean-slate designed hierarchies?
\end{enumerate}

The structure of the paper is as follows: In Section \ref{evolexis-background}, we present an overview of Lexis, the static optimization framework that serves as the main building block in Evo-Lexis.\footnote{The static (i.e., non-evolving) version of the proposed modeling framework is referred to as "Lexis" and it has been published at the ACM KDD 2016 conference \cite{siyari2016a}.} In Section \ref{evolexis-evo}, we present the components of the Evo-Lexis framework, along with the metrics that we use for the analysis of evolving hierarchies. In Section \ref{evolexis-models}, we evaluate the evolution of hierarchies under different target generation models. Sections \ref{evolexis-evolvability} and \ref{evolexis-transitions} present further analysis regarding the evolvability and major transitions in hierarchies produced using the most full-fledged (MRS) target generation model. Finally, Section \ref{evolexis-overhead} focuses on the comparison between clean-slate and incremental design in terms of cost and structure. In Section \ref{evolexis-related}, we review related work in the context of Evo-Lexis. Section \ref{evolexis-conclusion} discusses the results and presents some future research possibilities.

\begin{figure}[h]
\centering
\includegraphics[trim = 0cm 21.35cm 0cm 0cm,clip,width=\textwidth]{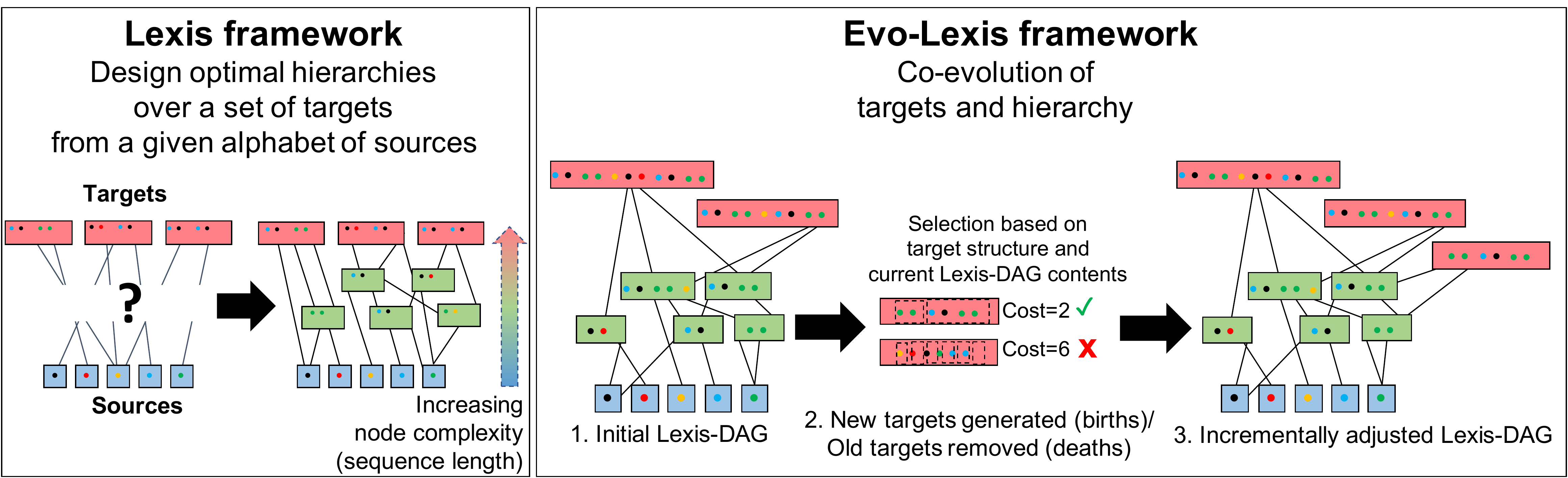}
\includegraphics[trim = 7.25cm 0cm 7.25cm 14.5cm,clip,width=\textwidth]{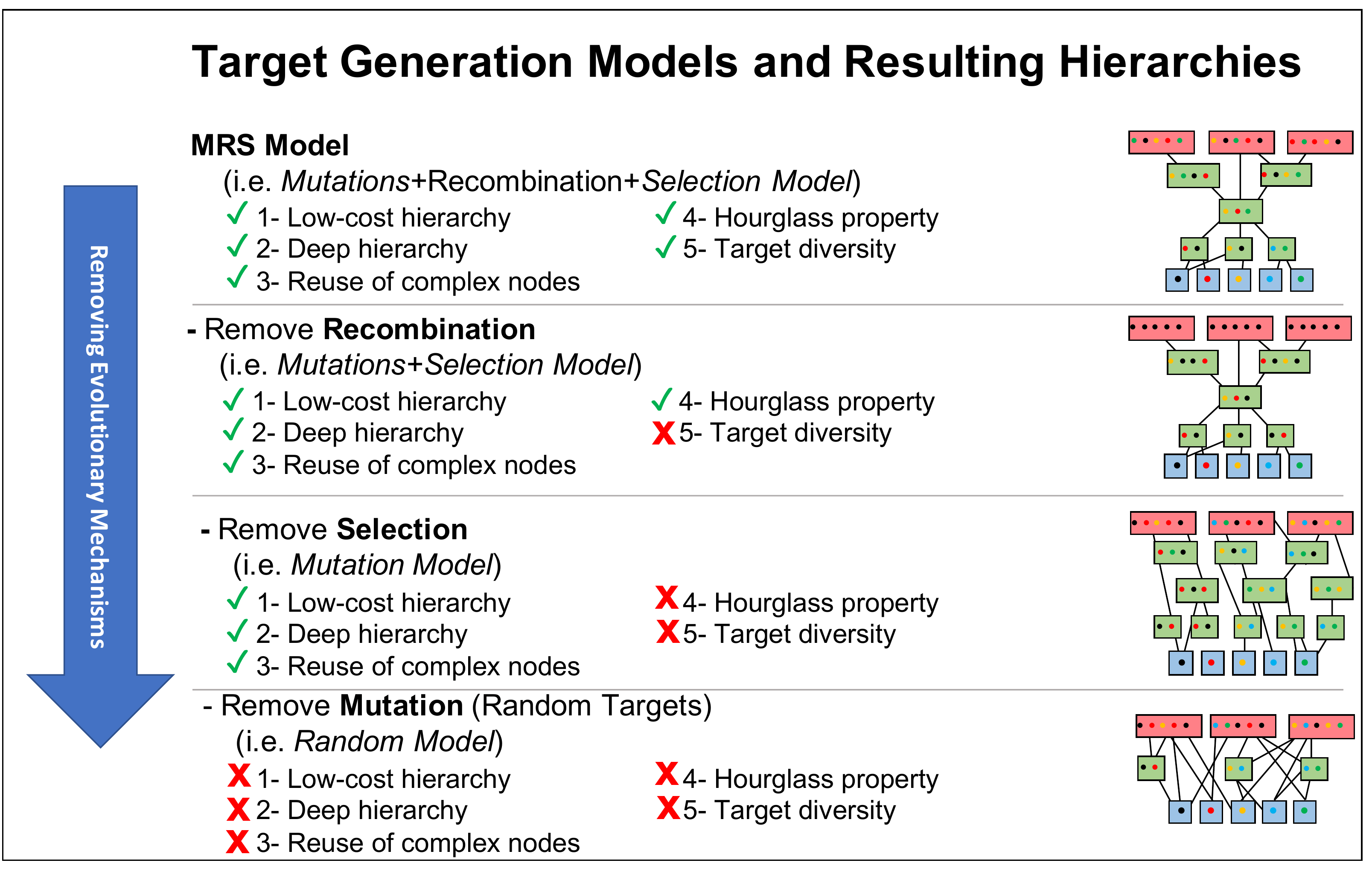}
\caption{{\bf Overview of this study.} The \emph{Evo-Lexis} modeling framework  captures the process of incrementally designing optimized hierarchies for a time-varying set of targets. Hierarchies are modeled as \emph{Lexis-DAGs}. We focus on key properties of the resulting hierarchies (e.g. cost, depth, reuse of intermediate components) and on how these properties depend on the evolutionary mechanisms that generate new targets. By focusing on well-known evolutionary mechanisms such as mutations, recombination and selection, we analyze how each of them affects the structure and evolution of the resulting hierarchies.
\\
Blue, green and red nodes show source, intermediate and target nodes, respectively. Colored dots represent an instance of a source node and are used to show the extent of diversity among target nodes.}
\label{fig:intro}
\end{figure}

%% file: 03-lexis-background.tex
\section{Lexis Background}
\label{evolexis-background}

In this section, we present an overview of Lexis \cite{siyari2016a}, the optimization framework that we use as the main building block of the \emph{Evo-Lexis} framework.
\subsection{Lexis-DAG}
Given an alphabet $S$ and a set of ``target'' strings $T$ over the alphabet $S$, we need to construct a Lexis-DAG.
A Lexis-DAG $D$ is a directed acyclic graph $D(V,E)$, where $V$ is the set of nodes and $E$ the set of edges, that satisfies the following three constraints:\footnote{To simplify the notation, even though $D$ is 
a function of $S$ and $T$, we do not denote it as such.} 

First, each node $v \in V$ in a Lexis-DAG represents a string $\mathcal{S}(v)$
of characters from the alphabet $S$.
The nodes $V_S$ that represent characters of $S$ are referred to as \emph{sources}, and they have zero in-degree.
The nodes $V_T$ that represent target strings $T=\{t_1, t_2, \dots, t_m\}$ are referred to as {\em targets}, and they
have zero out-degree.
$V$ also includes a set of {\em intermediate nodes} $V_M$, which represent substrings that appear in the targets $T$.  
So, $V=V_S\cup V_M\cup V_T$.

Second,
each node in $V_M\cup V_T$ of a Lexis-DAG represents a string that is the concatenation 
of two or more substrings, specified by the incoming edges from other nodes to that node. 
Specifically, 
an edge $e \in E$ from node $u$ to node $v$ is a triplet $(u,v,i)$ such that
the string $\mathcal{S}(u)$ appears as substring of $\mathcal{S}(v)$ at index $i$ 
(the first character of a string has index 1).
Note that there may be more than one edges from node $u$ to node $v$. 
The number of incoming and outgoing edges for a node $v$ is denoted by $d_{in}(v)$ and $d_{out}(v)$, respectively. 

\begin{figure}
\center
\subfloat[]{
\includegraphics[trim = 1.3cm 1.35cm 1.40cm 1.39cm,clip,scale=0.37]{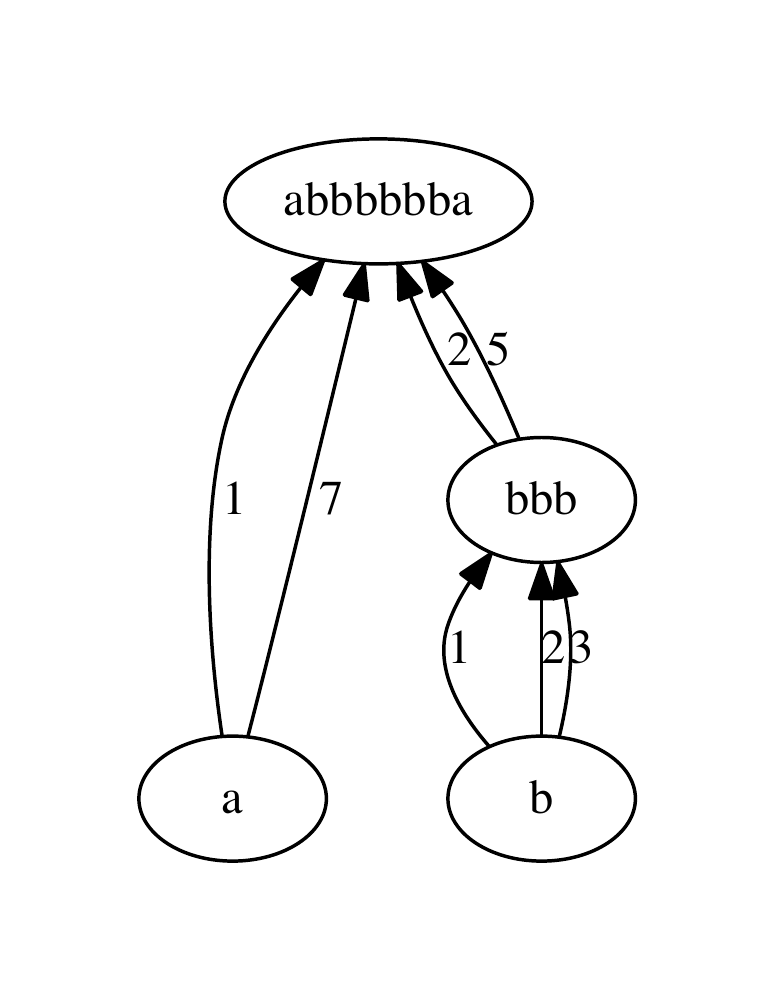}
}
\hspace{2cm}
  \subfloat[]{
  \includegraphics[trim = 1.4cm 1.35cm 1.35cm 1.39cm,clip,scale=0.37]{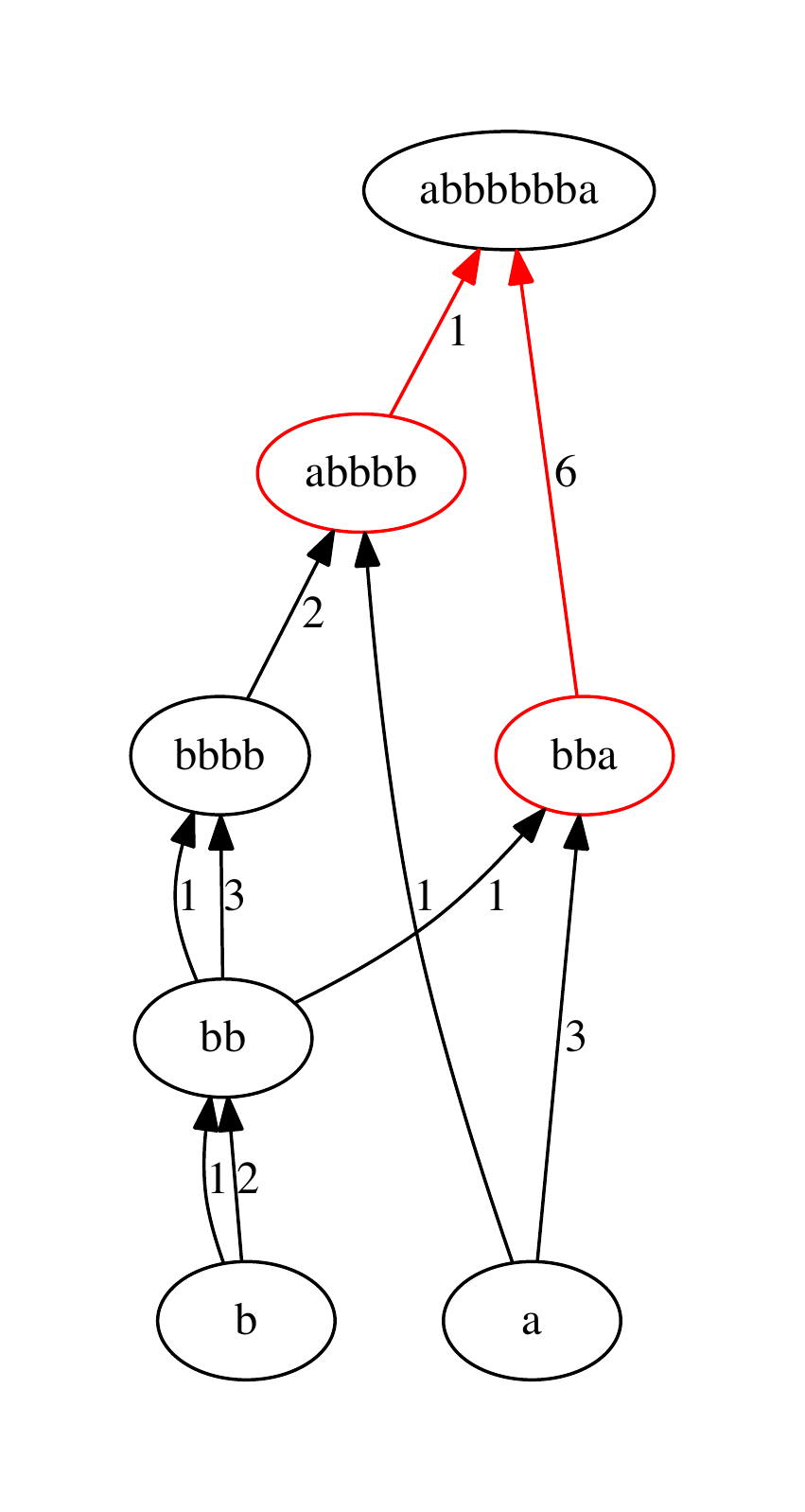}
  }
  \hspace{2cm}
    \subfloat[]{
    \includegraphics[trim = 1.3cm 1.35cm 1.35cm 1.39cm,clip,scale=0.37]{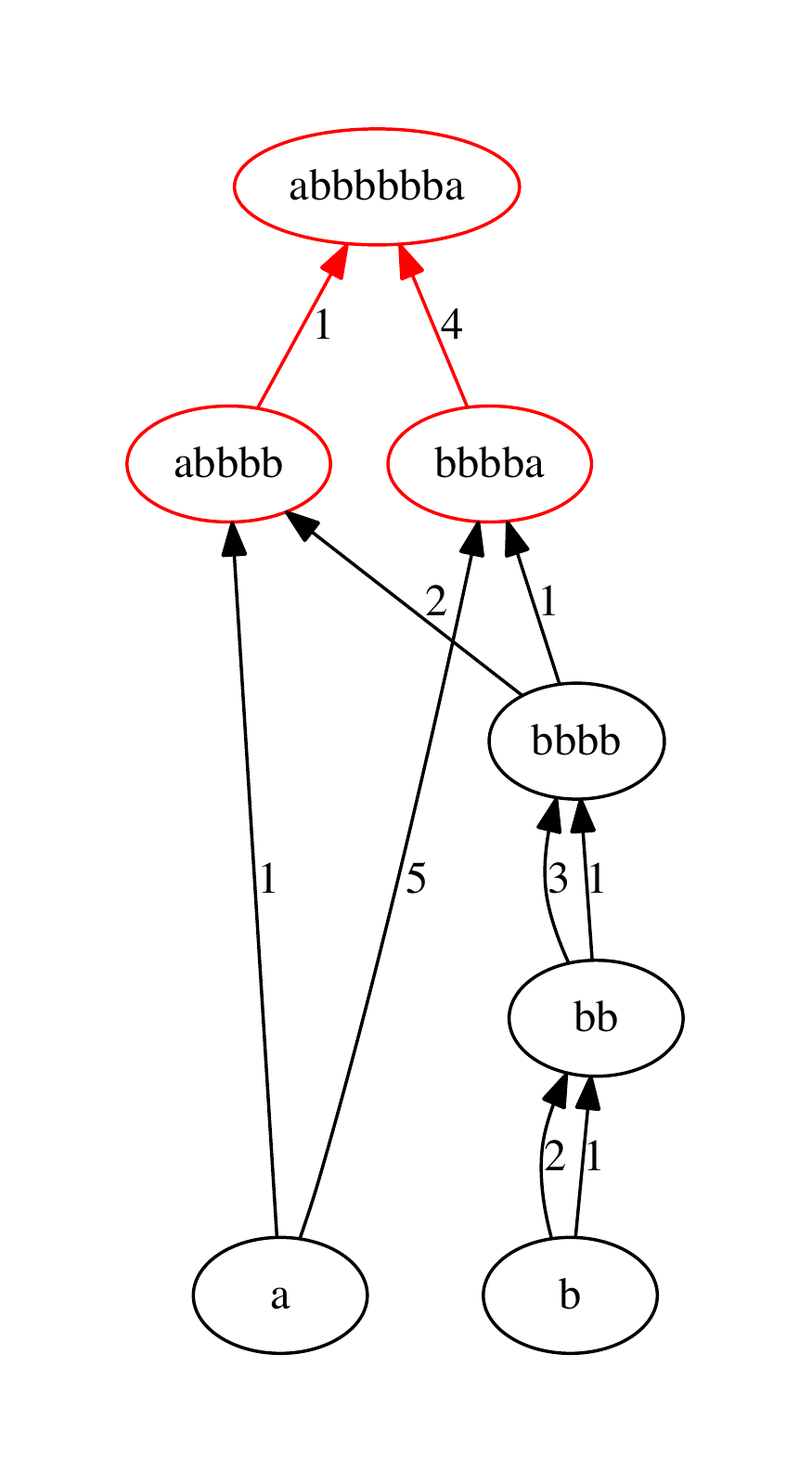}
    }
\caption{\small Illustration of the Lexis-DAG  for a single target $T=\{abbbbbba\}$ and sources $S=\{a,b\}$. Edge-labels indicate the occurrence indices: {\bf (a)} A valid Lexis-DAG having both minimum number of concatenations and edges. {\bf (b)} An invalid Lexis-DAG: two intermediate nodes are re-used only once. {\bf (c)} An invalid Lexis-DAG: the top-layer string is not equal to the concatenation of its two in-neighbors (best viewed in color).}
\label{fig:dagDef}
\end{figure}

Third, a Lexis-DAG should only include intermediate nodes that have an out-degree of at least two,
$\forall v\in V_M, d_{out}(v) \geq 2$. 
In other words, every intermediate node $v \in V_M$ in a Lexis-DAG should be such that 
the string $\mathcal{S}(v)$ is re-used in at least two concatenation operations. 
Otherwise, $\mathcal{S}(v)$ is either not used in any concatenation operation, 
or it is used only once and so the outgoing 
edge from $v$ can be replaced 
by re-wiring the incoming edges of $v$ straight to the single occurrence of $\mathcal{S}(v)$. 
In both cases node $v$ can be removed from the Lexis-DAG, resulting in a more parsimonious
hierarchical representation of the targets.  
Fig.~\ref{fig:dagDef} illustrates the concepts introduced in this subsection.
\subsection{The Lexis Optimization Problem}
The {\em Lexis} optimization problem is to construct a minimum-cost Lexis-DAG for the 
given alphabet $S$ and target strings $T$.
In other words, the problem is to determine the set of intermediate nodes $V_M$ and all
required edges $E$ so that the corresponding Lexis-DAG $D$ is optimal in terms of a given cost
function $C(D)$. This problem can be formulated as follows: 

\begin{equation}
\label{eq:Lexis}
\begin{aligned}
\text{~}&min_{(E,V_{M})}\text{~}C(D)\\
\text{~}&s.t.\text{~} D=(V,E)\text{ is a Lexis-DAG for $S$ and $T$}
\end{aligned}
\end{equation}

The selection of an appropriate cost function is somewhat application-specific. 
A natural cost function, as investigated in previous work \cite{siyari2016a}, is the number of edges in the Lexis-DAG. More general cost formulations, such as a variable edge cost or a weighted average 
of a node cost and an edge cost, are interesting but they are not pursued in this paper.  The {\em edge cost} to construct a node $v \in V$ is defined as the number of incoming edges required
to construct $\mathcal{S}(v)$ from its in-neighbors, which is equal to $d_{in}(v)$. 
The edge cost of source nodes is obviously zero. 
The edge cost $\mathcal{E}(D)$ of Lexis-DAG $D$ is defined as the edge cost of all nodes, which is
equal to the number of edges in $D$, 
\begin{equation}
\mathcal{E}(D) = \sum_{v\in V} d_{in}(v) = |E|
\label{eq:firstDAGCost}
\end{equation}
With edge cost, the problem in Eq. \eqref{eq:Lexis} is NP-Hard \cite{siyari2016a}. This problem is similar to the \emph{Smallest Grammar Problem} (SGP) \cite{sgp} and in fact its NP-Hardness is shown by a reduction from SGP \cite{siyari2016a}.

\begin{figure}[t]
\center
\subfloat[]{
  \includegraphics[trim = 1.3cm 1.4cm 1.44cm 1.4cm,clip,scale=0.35]{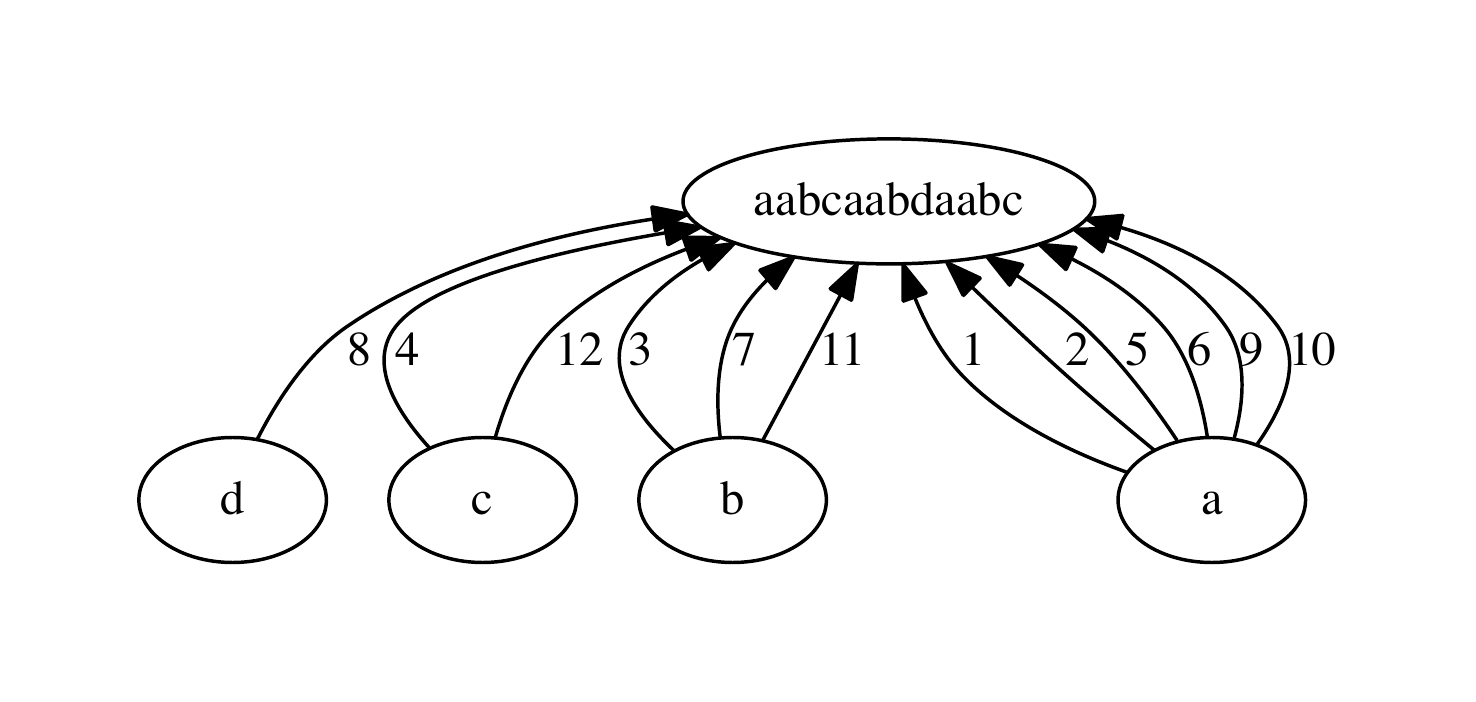}
}
\subfloat[]{
  \includegraphics[trim = 1.4cm 1.35cm 1.35cm 1.39cm,clip,scale=0.35]{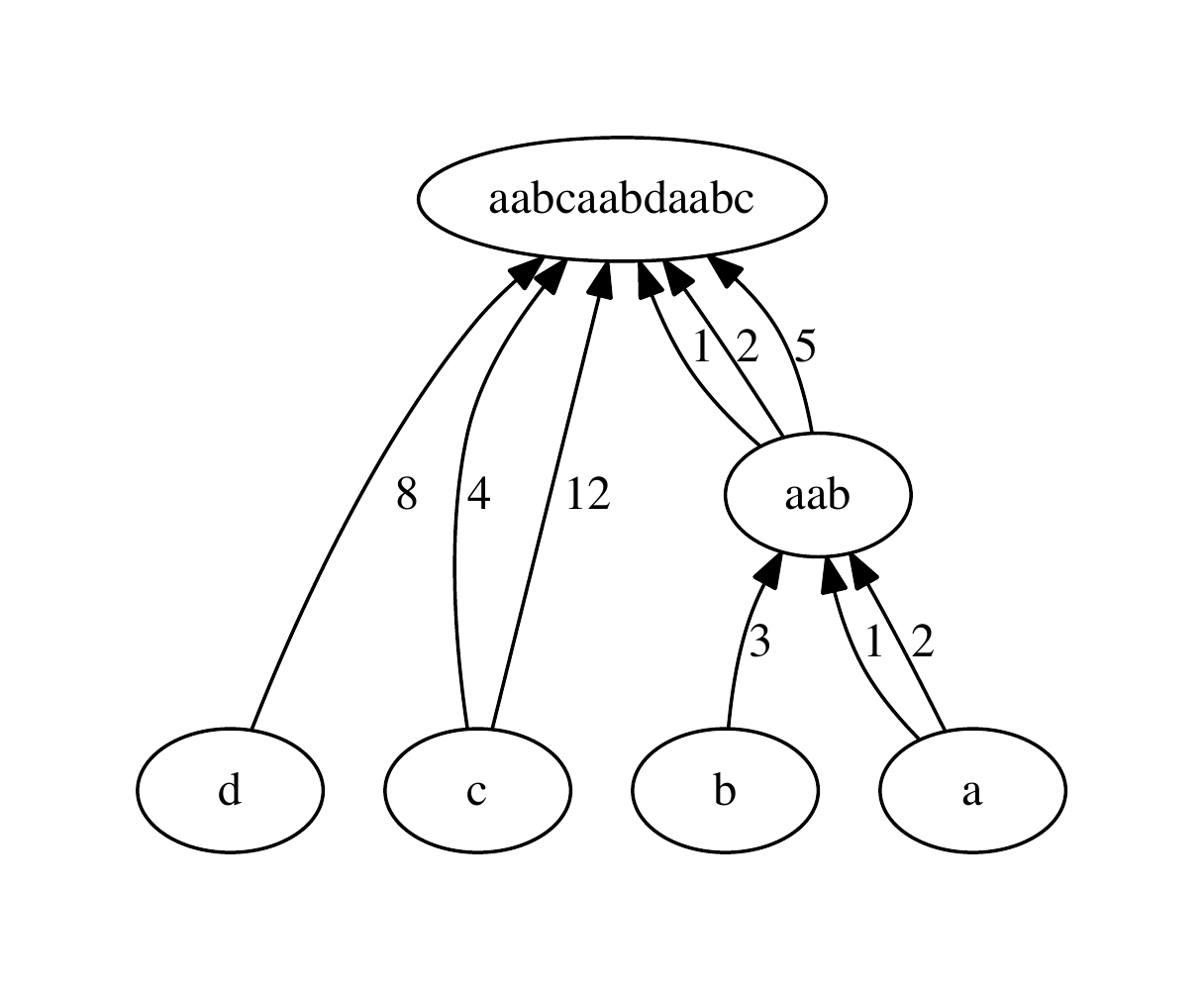}
}
\subfloat[]{
  \includegraphics[trim = 1.37cm 1.4cm 1.4cm 1.39cm,clip,scale=0.35]{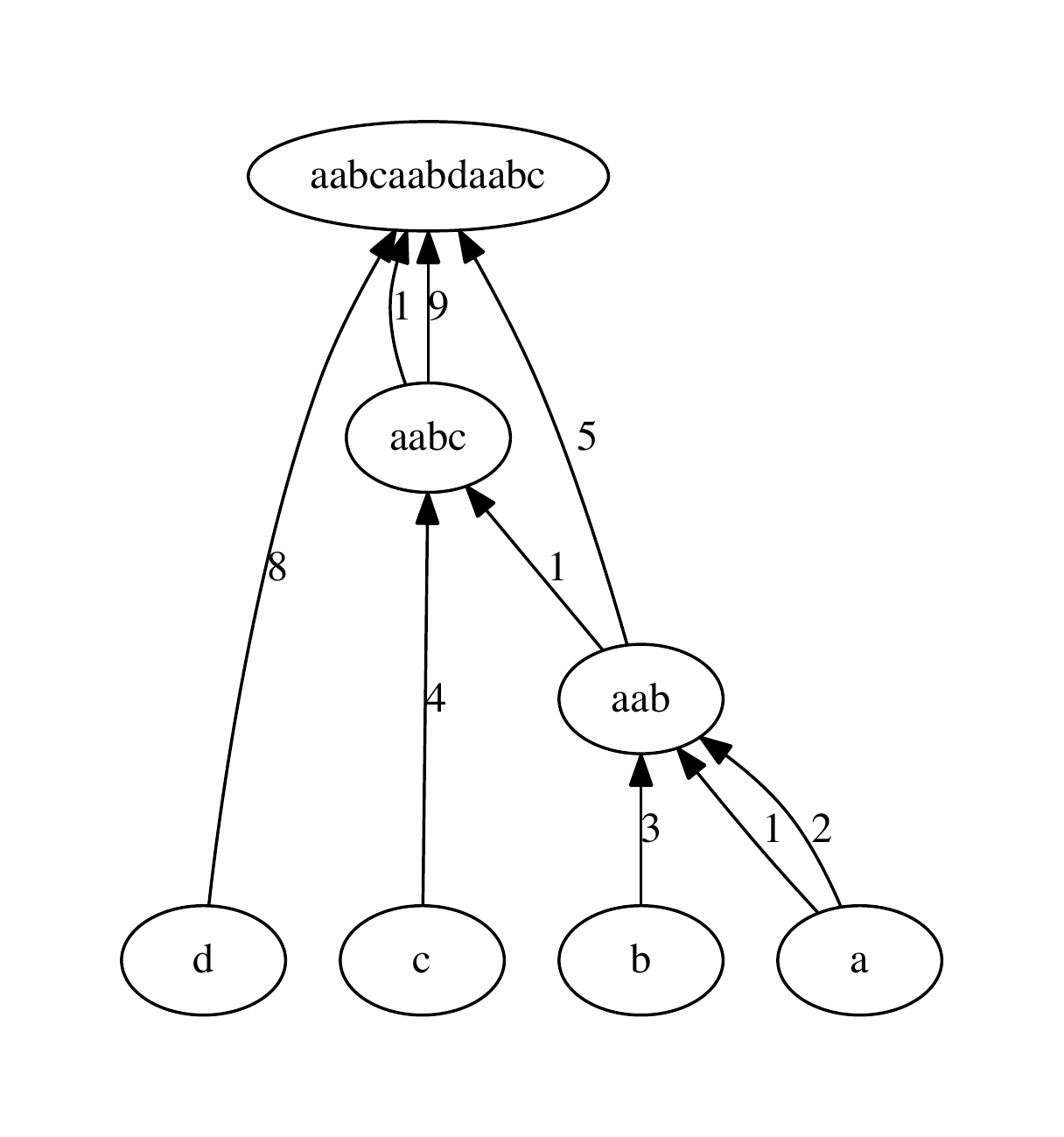}
}
\caption{Illustration of \textsc{G-Lexis} algorithm given target $T=\{aabcaabdaabc\}$ and sources $S=\{a,b,c,d\}$. - {\bf (a)} Initial Lexis-DAG. {\bf (b)} Substring \emph{aab} has maximum cost reduction by reducing the number of edges in the Lexis-DAG from 12 to 9. {\bf (c)} The substring $aabc$ has maximum cost reduction. Note how $aabc$ is partially made from the previously added substring $aab$. In this example, this would be the last iteration of \textsc{G-Lexis}.}
\label{fig:glexisExample}
\end{figure}

We solve the Lexis optimization problem in Eq. \eqref{eq:Lexis} with a greedy heuristic, called \textsc{G-Lexis}. 
\textsc{G-Lexis} starts with the trivial flat Lexis-DAG, and at each iteration it chooses the substring $\xi$  that maximally reduces the edge cost, when it is added as a new intermediate node to the Lexis-DAG and the corresponding edges are rewired by its addition.
The algorithm terminates when there are no more substrings that reduce the cost of the Lexis-DAG. An example of application of the \textsc{G-Lexis} algorithm is shown in Fig. \ref{fig:glexisExample}. More details regarding the efficient implementation and complexity of the algorithm can be found in \cite{siyari2016a}.

\subsection{Path-Centrality and the Core of a Lexis-DAG}
After constructing a Lexis-DAG, an important question is to rank the constructed intermediate
nodes in terms of significance or {\em centrality.} In a Lexis-DAG, a path that starts from a source and terminates at a target represents a dependency chain in which each node depends on all previous nodes in that path.
Thus, the higher the number of such source-to-target paths traversing an intermediate node $v$ is,
the more important $v$ is in terms of the number of dependency chains it participates in.
More formally, let $P_D(v)$ be the number of source-to-target paths that traverse node $v \in V_M$;
we refer to $P_D(v)$ as the {\em path centrality} of intermediate node $v$.
Path centrality can be computed as:
\begin{equation}
\vspace{-2mm}
P(v) = P_S(v) \, P_T(v)
\end{equation}
where $P_S(v)$  is the number of paths from any source to $v$, 
and $P_T(v)$ is the number of paths from $v$ to any target. \footnote{A similar metric, called \emph{stress centrality} of a vertex, is studied in \cite{sdm-cent}.} It is easy to see that $P_T(v)$ is equal to the number of times the string that corresponds to $v$ is used in the set of targets $T$ . Similarly, $P_S(v)$ is equal to the number of times any source node is used in the string of $v$, which is simply the length of that string. Hence, the path centrality of a node $v$ is simply the product of the length of the string of $v$ (proxy for complexity) and its number of appearances (proxy for generality).

An important follow-up question is to identify the {\em core} of a Lexis-DAG, i.e., a set 
of intermediate nodes that represent, as a whole, the most important substrings in that Lexis-DAG. The core set is the representative set of nodes that summarizes the structure of the targets. 
Intuitively, we expect that the core should include nodes of high path centrality, and that almost 
all source-to-target dependency chains of the Lexis-DAG should traverse at least one of these core nodes. 

More formally, suppose $K$ is a set of intermediate nodes 
and $\mathscr{P}^-(K)$ is the set of source-to-target paths after we remove the nodes in $K$ from $D$. 
The core of $D$ is defined as the minimum-cardinality set of intermediate nodes $Core(\tau)=\hat{K}$ such that
the fraction of remaining source-to-target paths after the removal of $\hat{K}$ is at most $\tau$:\footnote{To simplify notation, we do not denote the core set as function of $D$.}
\begin{equation}
\label{eq:coreid}
\begin{aligned}
\hat{K}=arg min_{\text{~}K \subseteq V_M}\text{~~~}&|K|\\
s.t.\text{~}&|\mathscr{P}^-(K)| \leq \tau \, |\mathscr{P}^-(\varnothing)| 
\end{aligned}
\end{equation}
where $|\mathscr{P}^-(\varnothing)|$ is the number of source-to-target paths in the original Lexis-DAG,
without removing any nodes.\footnote{It is easy to see that $|\mathscr{P}^-(\varnothing)|$ is equal to the cumulative 
length $L$ of all target strings.} Fig. \ref{fig:hscore} shows an example defining the concepts regarding the core of a Lexis-DAG.

Note that if $\tau=0$ the core identification problem in Eq. \eqref{eq:coreid} becomes equivalent to finding the min-vertex-cut of 
the given Lexis-DAG. 
In practice, a Lexis-DAG often includes some {\em tendril-like} source-to-target paths traversing
a small number of intermediate nodes that very few other paths traverse. These paths can cause a large increase in
the size of the core. For this reason, we prefer to consider the case of a positive, but potentially small, 
value of the threshold $\tau$.   

We solve the core identification problem with a greedy algorithm referred to as \textsc{G-Core}. 
This algorithm adds in each iteration the node with the highest path-centrality value to the core set, 
updates the Lexis-DAG by removing that node and its edges, and recomputes the path centralities of the remaining nodes before
the next iteration. The algorithm terminates when the desired fraction of source-to-target paths is achieved.

\subsection{Hourglass score}
Intuitively, a Lexis-DAG exhibits the hourglass effect if it has a small core. To make this intuition more precise, we compare the size of the core of a Lexis-DAG with the core size of a derived Lexis-DAG which maintains the source-target paths of the original Lexis-DAG but that is not presenting the hourglass structure by construction.

We use a metric, named as Hourglass
Score, or \emph{H-Score}, in our study for measuring the ``hourglass-ness'' of a network. This metric was originally presented in \cite{sabrin2016}.

To calculate the H-score, we create a flat Lexis-DAG $D_f$ containing the same targets as the original Lexis-DAG $D$. Note that $D_f$ preserves the source-target dependencies of $D$: each target in $D_f$ is constructed based on the same set of sources as in $D$. However, the dependency paths in $D_f$ are direct, without forming any intermediate modules that could be reused across different targets. So, by construction, the flat Lexis-DAG $D_f$ cannot have a non-trivial core since it does not have any intermediate nodes. 

We define the H-score as follows:
\begin{equation}
H_D(\tau)=1- \frac{|Core(\tau)|}{|Core_f(\tau)|}
\end{equation}

Where $Core(\tau)$ and $Core_f(\tau)$ are the core sets of $D$ and $D_f$ for a given threshold $\tau$, respectively. Note that $Core_f$ can include a combination of sources and targets, and it would never be larger than either the set of sources or targets, i.e.,

\begin{equation}
|Core_f(\tau)| \leq min\{|S|,|T|\}
\end{equation}

\begin{figure}[t]
\center
\subfloat[]{
  \includegraphics[trim = 1.4cm 0cm 1.35cm 1.39cm,clip,scale=0.2]{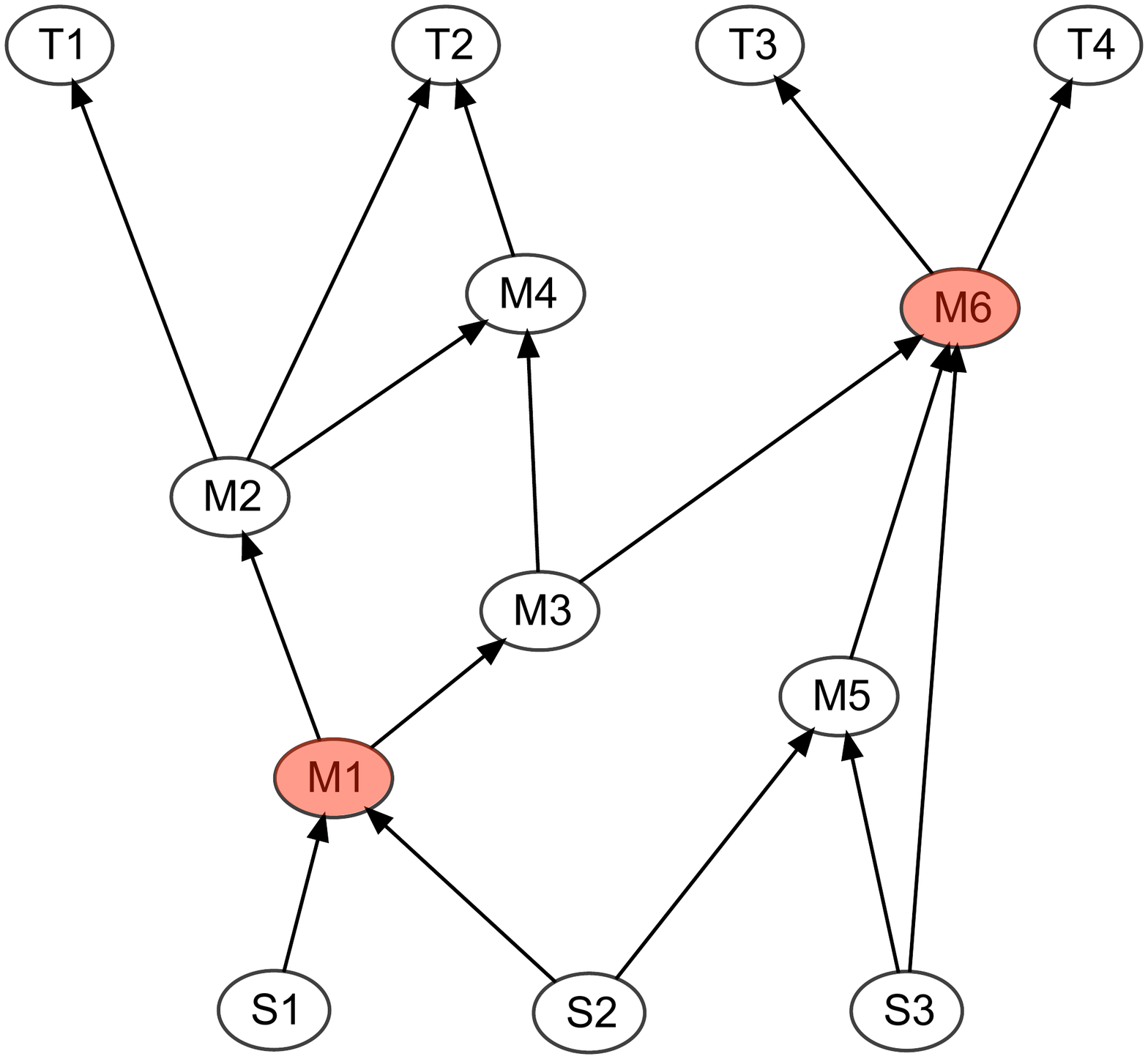}
}
\subfloat[]{
  \includegraphics[trim = 1.37cm 0cm 1.4cm 1.39cm,clip,scale=0.2]{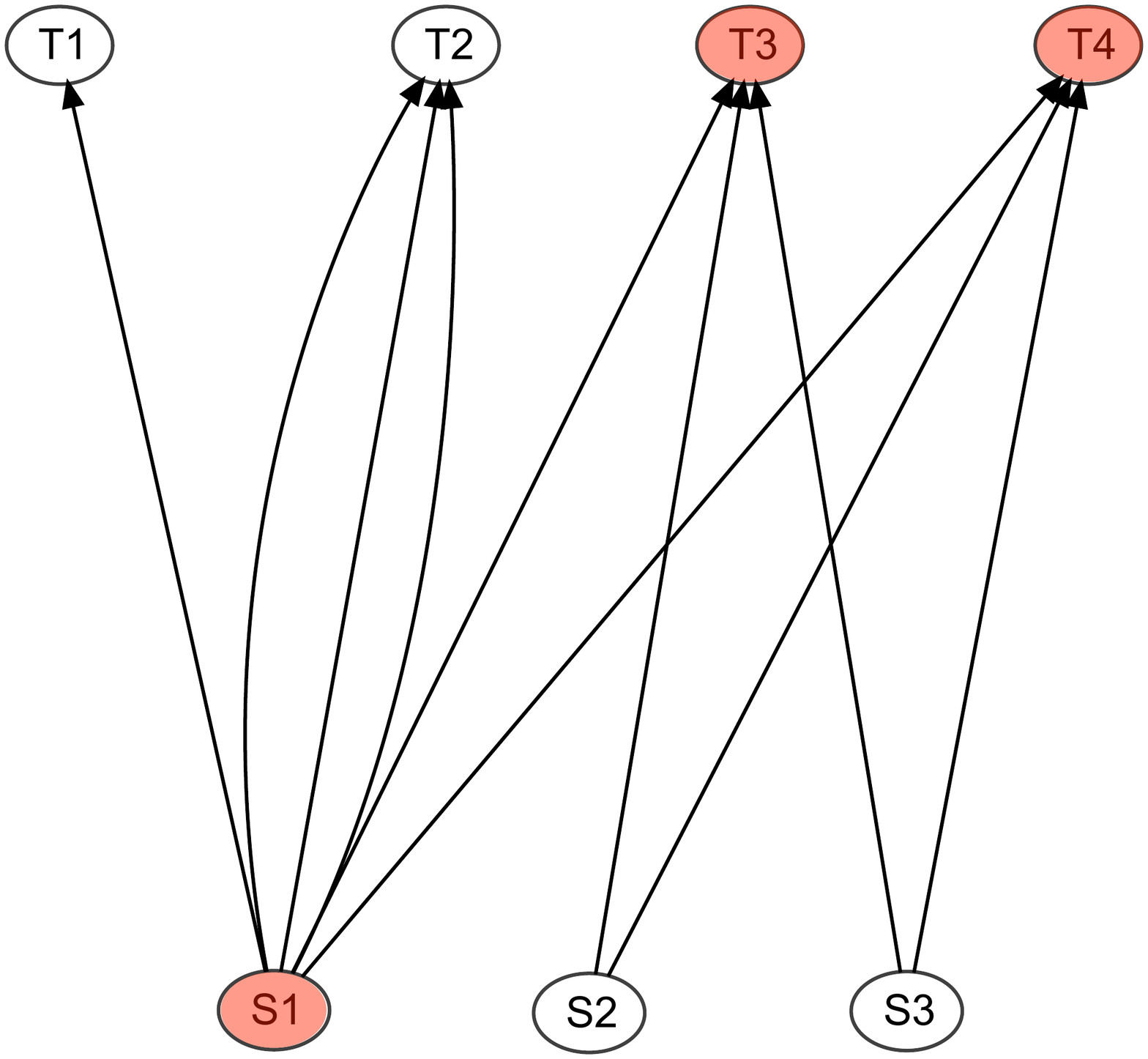}
}
\caption{{\bf (a)} Original Lexis-DAG $D$ and its core nodes highlighted (for clarity, the string of each node is not shown and the nodes are referred to with labels). For $\tau=0.9$, we have $Core(\tau) = \{M1,M6\}$. {\bf (b)} $D_f$, flat version of $D$. For same $\tau=0.9$, we have $Core_f(\tau)=\{T3,T4,S1\}$. Hence, the H-score is $H_D(\tau)=1-\frac{2}{3}=0.33$.
}
\label{fig:hscore}
\end{figure}

Clearly, $0 \leq H(\tau) \leq 1$. The H-score of $D$ is approximately one if the core size of the original Lexis-DAG is negligible compared to the the core size of the corresponding flat Lexis-DAG. Fig. \ref{fig:hscore} illustrates the definition of this metric. An ideal hourglass-like Lexis-DAG would have a single intermediate node that is traversed by every single source-to-target path (i.e., $Core(1)=1$), and a large number of sources and targets none of which originates or terminates, respectively, a large fraction of source-to-target paths (i.e., a large value of $Core_f(1)$). The H-score of this Lexis-DAG would be approximately equal to one.


%% file: 04-evo-modeling.tex
\section{Evo-Lexis Framework and Metrics}
\label{evolexis-evo}

\begin{figure}[h]
\center
\includegraphics[trim = 0cm 0cm 0cm 0cm,clip,width=\textwidth]{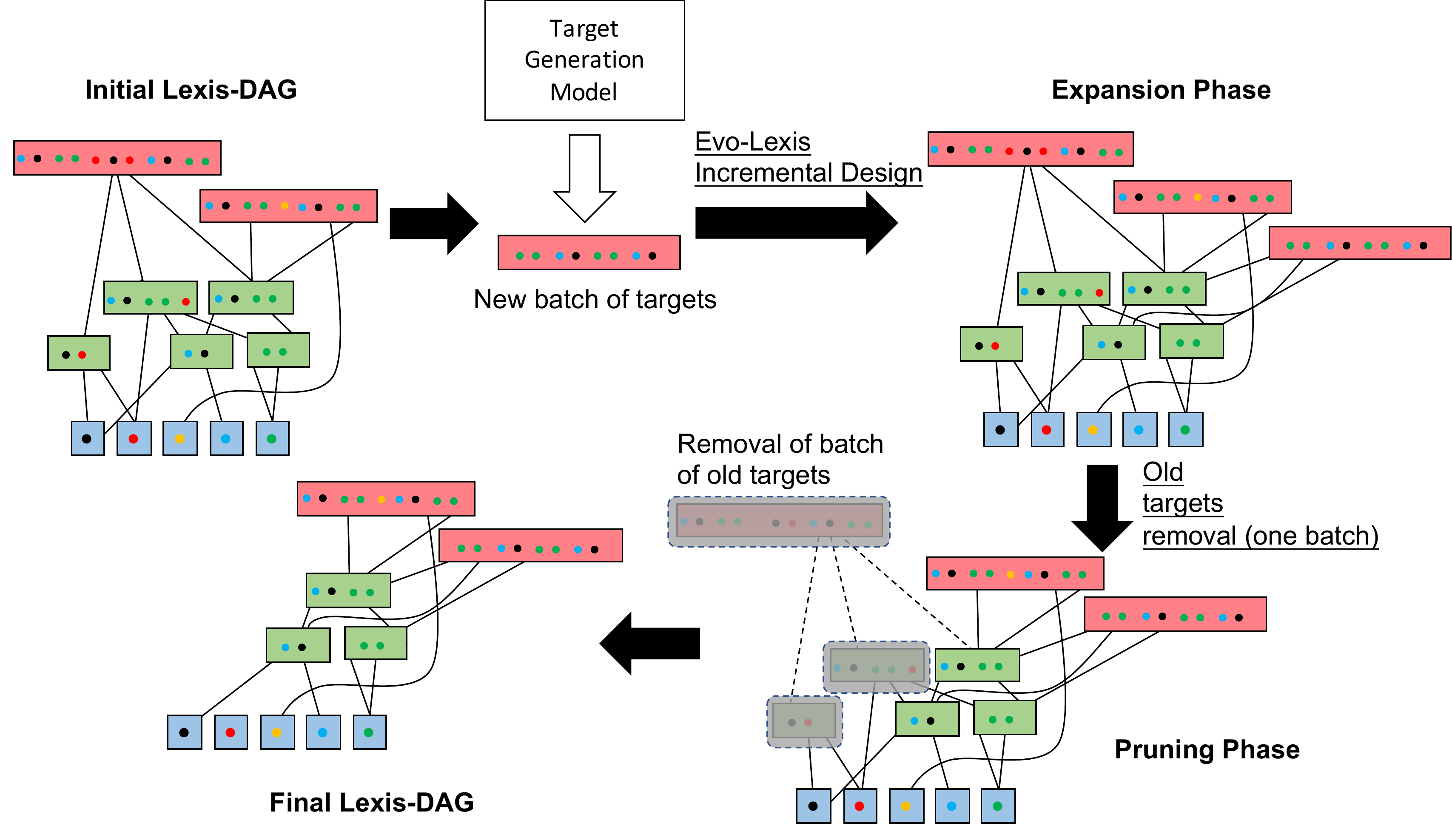}
\caption{A diagram of the Evo-Lexis framework. In every iteration, the following steps are performed: {\bf (1)} A batch of new targets is generated via a target generation model. {\bf (2)} In the ``expansion phase'', the new targets are added incrementally to the current Lexis-DAG by minimizing the marginal cost of adding every new target to the existing hierarchy. {\bf (3)} If the number of targets that are present in the system has reached a steady-state threshold, we also remove the batch of oldest targets from the Lexis-DAG. During this ``pruning phase'', some intermediate nodes may also be removed because every intermediate node in a valid Lexis-DAG should have an out-degree of at least two. 
}
\label{fig:evo-model}
\end{figure}

The Evo-Lexis framework includes a number of components that are described below. A general illustration of the framework is shown in Fig. \ref{fig:evo-model}.
\begin{itemize}
\item
{\bf Lexis-DAG:} The network that encodes the system's architecture at a given point in time. The inputs of the system are the sources of the DAG and the outputs are the targets.
\item 
{\bf Target Generation Model:} This model specifies the evolutionary process that creates new targets. For simplicity, we consider the addition of only new targets, not new sources. 
The generation of new targets can be either independent of the current hierarchy ({\em exogenous target generation}) or it can depend on that hierarchy ({\em endogenous target generation}).
\item
{\bf Target Removal Model:} Models the removal of older targets. The total number of targets remains constant during the evolution of the network. 
\item
{\bf Hierarchy Design Algorithm:} This is how the Lexis-DAG is adjusted whenever we introduce new targets. This procedure can be as simple as building a Lexis-DAG from scratch (by running the \textsc{G-Lexis} algorithm) on the set of existing targets. We refer to this approach as \emph{Clean-Slate design}. On the contrary, the algorithm can be incremental, starting with the previously constructed hierarchy and incorporating new targets in a way that minimizes the adjustment cost. We refer to this algorithm an \emph{Incremental design}, and it is described next. \end{itemize}

\subsection{Incremental Design Algorithm}
The Evo-Lexis algorithm generates an optimized hierarchy for the given set of targets in every evolutionary iteration. As mentioned previously, the Clean-Slate design approach is to discard the existing hierarchy and redesign from scratch a new Lexis-DAG  for the given set of targets using the \textsc{G-Lexis} algorithm.
Such a design methodology is not realistic however in either technological or natural evolution. A more realistic approach is to adjust the existing Lexis-DAG incrementally, as described below.

In incremental design, given a Lexis-DAG $D_0$ with a set of targets $T_0$, a set of new targets $T_+$ to be added, and a set of old targets $T_-$ to be removed, the problem is to construct a Lexis-DAG $D^{\text{\textsc{Inc}}}$ that supports the set of targets $\{T_0 \cup T_+ - T_-\}$, and that minimizes the cost difference with respect to $D_0$: 

\begin{equation}
\label{incv1}
\begin{aligned}
\text{~}&min_{D^{\text{\textsc{Inc}}}}~\{\mathcal{E}\left(D^{\text{\textsc{Inc}}}\right) - \mathcal{E}\left(D_0\right)\}\\
\text{~}&\text{s.t. } D^{\text{\textsc{Inc}}} \text{ is a Lexis-DAG for } \{T_0 \cup T_+ - T_-\}
\end{aligned}
\end{equation}

If $D_0 = \phi$ (i.e., there is no initial Lexis-DAG), $T_-=\phi$, and $T_+$ is the  entire target set, the incremental design problem becomes equivalent to the original Lexis Optimization Problem in Eq. \eqref{eq:Lexis}.

The incremental design problem is NP-Hard (as the original Lexis design problem
in which $D_0 = \phi$ and $T_-=\phi$), and so we rely on a heuristic that we refer to as \textsc{Inc-Lexis}. The algorithm 
proceeds in two phases: first, in the ``expansion phase'', it adds the set
of new targets $T_+$ attempting to reuse as much as possible existing
intermediate nodes. Second, in the ``pruning phase'', the algorithm removes
the set of old targets $T_-$, and it also removes any intermediate nodes
that are left with zero or one outgoing edges. 

\begin{figure}
\center
\subfloat[\label{fig:inc-lexis:1}]{
  \includegraphics[trim = .2cm 3cm -4cm 5cm,clip,width=\textwidth]{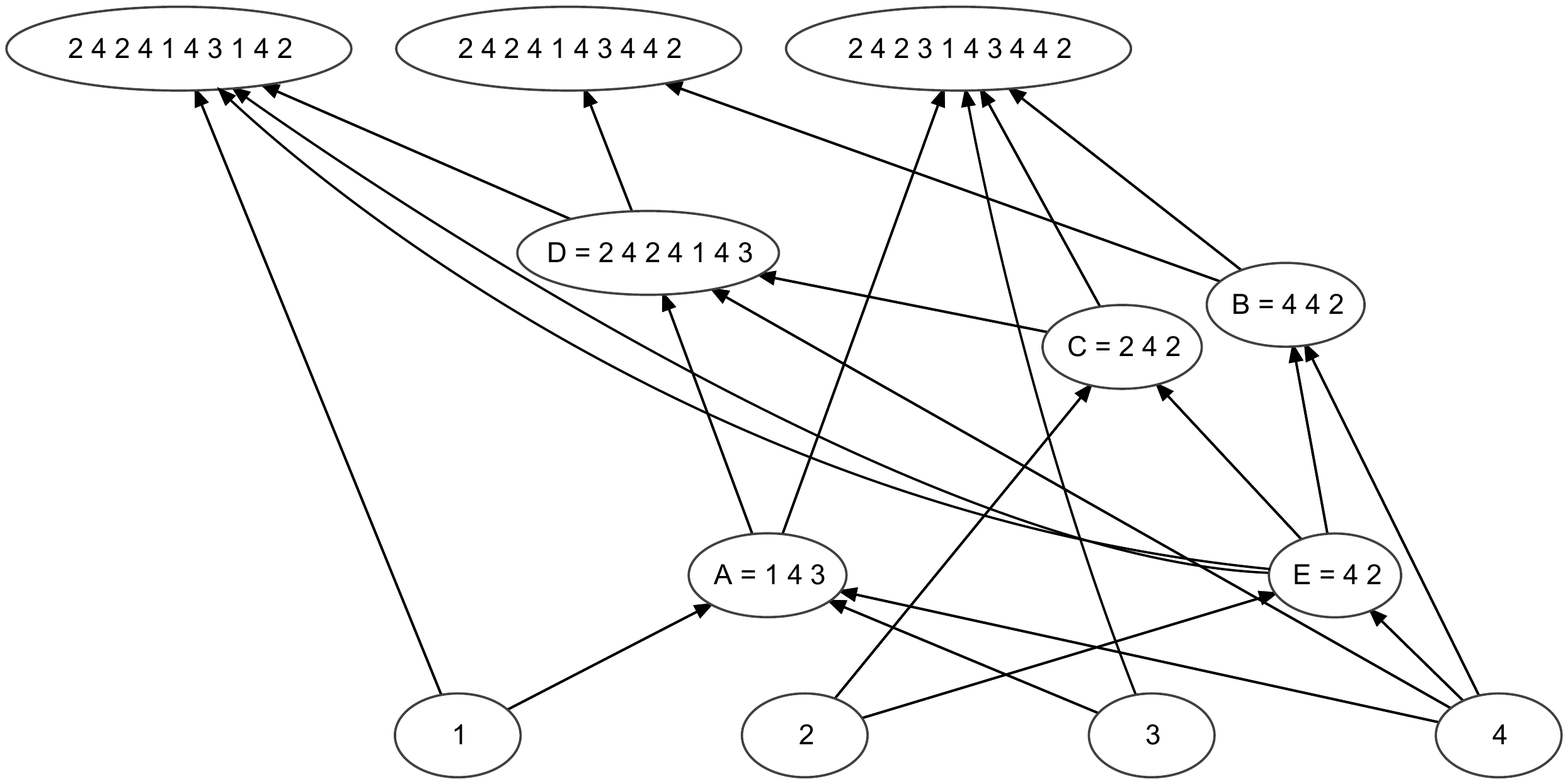}
}
\\
\subfloat[\label{fig:inc-lexis:2}]{
  \includegraphics[trim = 0cm 4.5cm 2.5cm 7cm,clip,width=\textwidth]{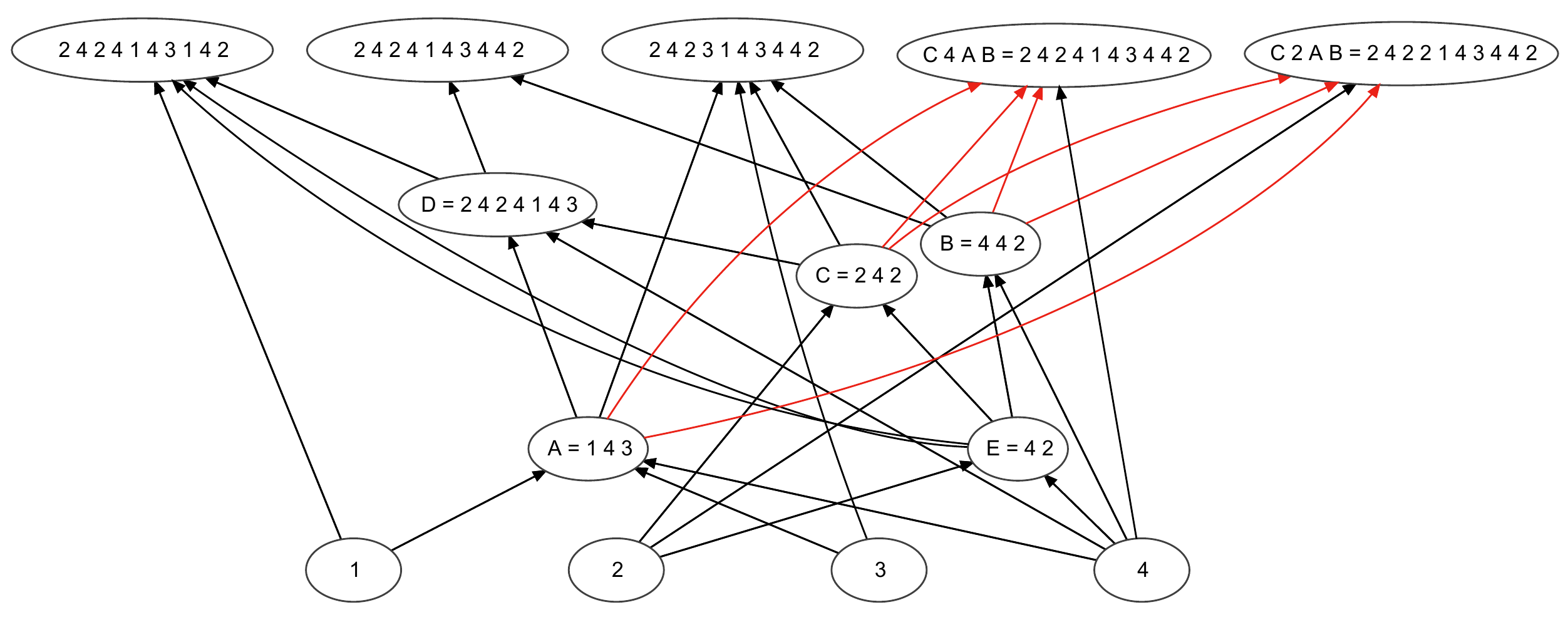}
}
\\
\subfloat[\label{fig:inc-lexis:3}]{
  \includegraphics[trim = 0cm 4.5cm 2.5cm 7cm,clip,width=\textwidth]{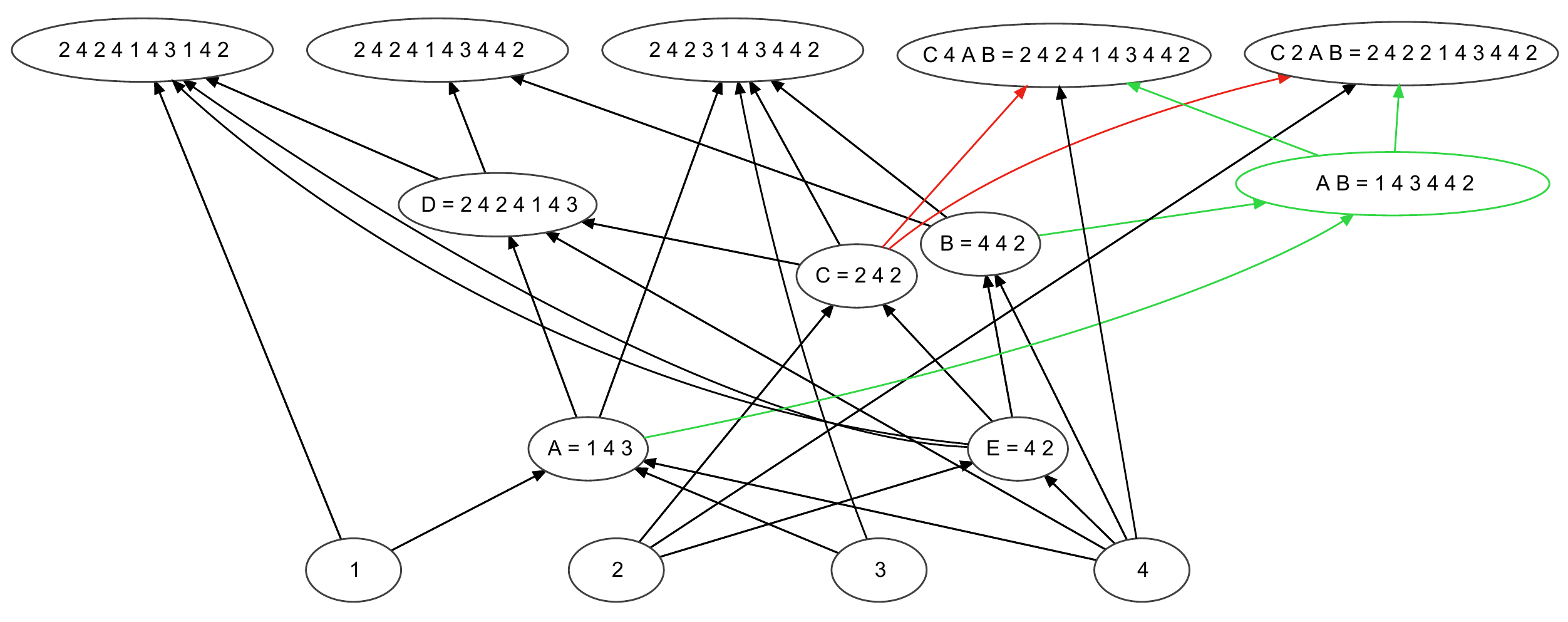}
}
\caption{Illustration of \textsc{Inc-Lexis}. - {\bf (a)} Initial Lexis-DAG $D_0$ with $T=\{2424143142,2424143442,2423143442\}$, $S=\{1,2,3,4\}$ and $M=\{2424143,442,242,143,42\}$. {\bf (b)} The new targets are $T_+=\{0424143442,2424143242,2422143442\}$. In the first stage of \textsc{Inc-Lexis}, the substrings in $M \cup S$ are reused to construct $T_+$. Red edges show the reuse of substrings in $T_+$. Node labels show the representation of each node  using the extended alphabet formed by intermediate nodes. This representation is used in the second stage of the expansion phase to run \textsc{G-Lexis} on $T_+$. {\bf (c)} The Lexis-DAG after running \textsc{G-Lexis} on the set $T_+$ in its extended alphabet form. The green nodes and edges are the results of this stage. (Continued in Fig. \ref{fig:inc-lexis2})}
\label{fig:inc-lexis}
\end{figure}
\begin{figure}[t]
\center
\subfloat[\label{fig:inc-lexis:4}]{
  \includegraphics[trim = 0cm 4.5cm 2.5cm 7cm,clip,width=\textwidth]{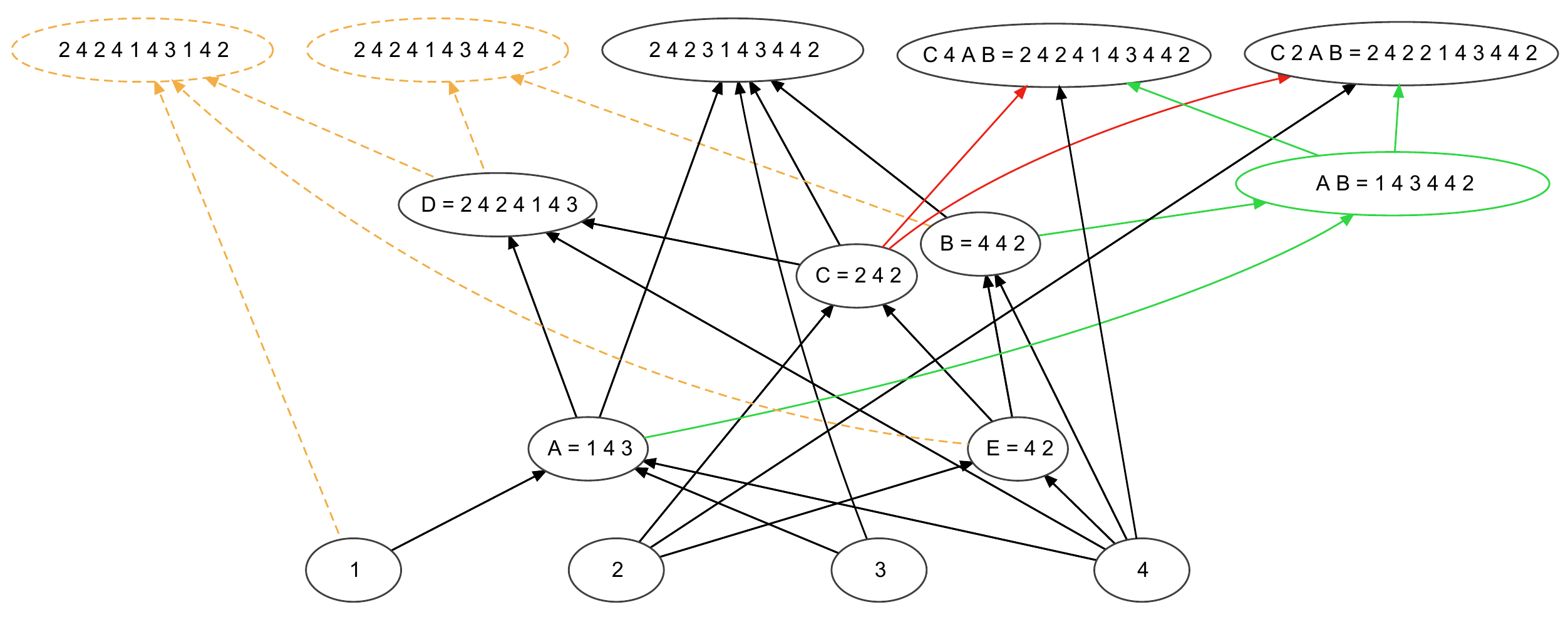}
}
\\
\subfloat[\label{fig:inc-lexis:5}]{
  \includegraphics[trim = -5cm 3cm .2cm 5cm,clip,width=\textwidth]{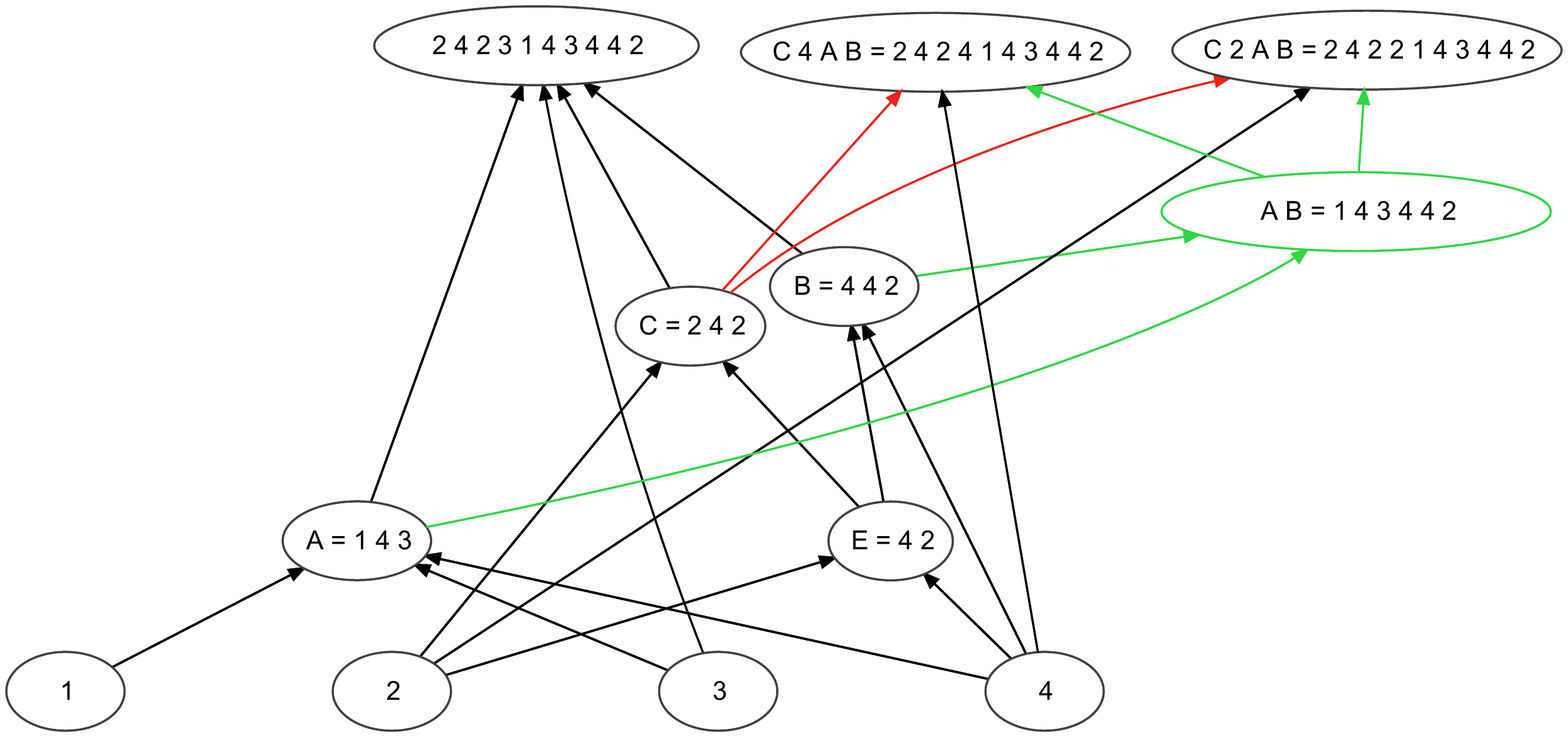}
}
\caption{(Continued from Fig. \ref{fig:inc-lexis}) Illustration of \textsc{Inc-Lexis}. {\bf (d)} The target nodes $2424143142$ and $2424143442$ are  removed during the pruning phase. All incoming edges (dashed and shown in yellow) will also be removed, which leaves the node $D=2424143$ with zero out-degree. {\bf {(e)}} The final Lexis-DAG after removal of targets and intermediate nodes with zero and one out-degree.}
\label{fig:inc-lexis2}
\end{figure} 

In more detail, the expansion phase of \textsc{Inc-Lexis} consists of two stages: in stage-1, we reuse intermediate nodes present in $D_0$ to cover $T_+$ with minimum cost. 
In stage-2 of the expansion phase, we further optimize the hierarchy that supports the targets in $T_+$ by building an optimized Lexis-DAG for them using \textsc{G-Lexis}.  The resulting new intermediate nodes and edges are added in the existing DAG. 

Note that stage-1 relates to the well-known {\em Optimal Parsing} problem, which is:
given a set of target strings $T$, a set of substrings $M$ and the corresponding
alphabet $S$, what is the minimum number of substrings and letters that can construct $T$ from the elements of $M \cup S$? The optimal parsing problem can be formulated as a shortest-path problem in directed graphs \cite{optimal-parsing}. If the length of the targets is $N$, it can be optimally solved in $O(N + |M \cup S|)$ as the corresponding directed acyclic graph has $N$ nodes  and $O(N+|M \cup S|)$ unweighted edges.

In the pruning phase,  we remove the oldest batch of targets. We also ensure that there is no redundant node in the Lexis-DAG, as implied by the constraint: $\forall v\in V_M, d_{out}(v) \geq 2$. This ensures that the Lexis-DAG does not include two types of redundancies: nodes with zero out-degree and nodes that are only reused once. 

Figures \ref{fig:inc-lexis} and \ref{fig:inc-lexis2}
give an example of how \textsc{Inc-Lexis} adjusts a hierarchy, given a set
of targets to be added and a set of targets to be removed.

\subsection{Target Generation Models}
The targets are generated through well-known evolutionary mechanisms, such as tinkering/mutation, recombination and selection:
\begin{itemize}
\item 
The generation of new targets from minor changes in earlier targets is similar to \emph{Tinkering}/\emph{Mutation}. 
Tinkering is common in technological evolution: 
small ``upgrades'' in a software or hardware artifacts are the most common example of this process. In biological systems, it is well-known that mutation is basically ``the engine of evolution'' \cite{mutation-evo-engine}. In Evo-Lexis, tinkering/mutation is performed by replacing one character of a given target with a randomly chosen character.
\item
In the technological world, \emph{Recombination} is known to be one of the central mechanisms for the creation of new technologies \cite{Arthur2009}. Technological design is often considered to be a search over a space of combinatorial possibilities \cite{youn2015}. In fact, many breakthroughs in the history of technology were in fact just a new combination of existing modules. A recent example is the first version of the iPhone in 2007, which was introduced to be ``a phone, an internet communicator and an iPod''. In biology, it is well known that recombination and crossover is essential as it produces highly diverse genotypes, compared to mutations.
\item
\emph{Selection} is an essential mechanism in evolution. In natural systems, selection determines whether a new genotype can survive the competition with existing genotypes (i.e., the incumbents) by evaluating the phenotypic fitness 
of the former relative to the latter. In the technological world, selection is the process of evaluating the functionality and cost of a new product, perhaps during an R\&D cycle \cite{mut-tech}. 
In the Evo-Lexis framework, selection is performed to decide whether a candidate target can be accepted, by evaluating the cost of adding that target in the current hierarchy. In other words, selection creates an \emph{endogenous} target generation process in which the existing hierarchy determines the cost of the potential new targets and thus, whether each new target is cost-competitive compared to the targets it evolved from.
\end{itemize}

\subsubsection{MRS Model}

\begin{figure}[h]
\centering
\includegraphics[trim = 0cm 0cm 0cm 0cm, clip,width=\textwidth]{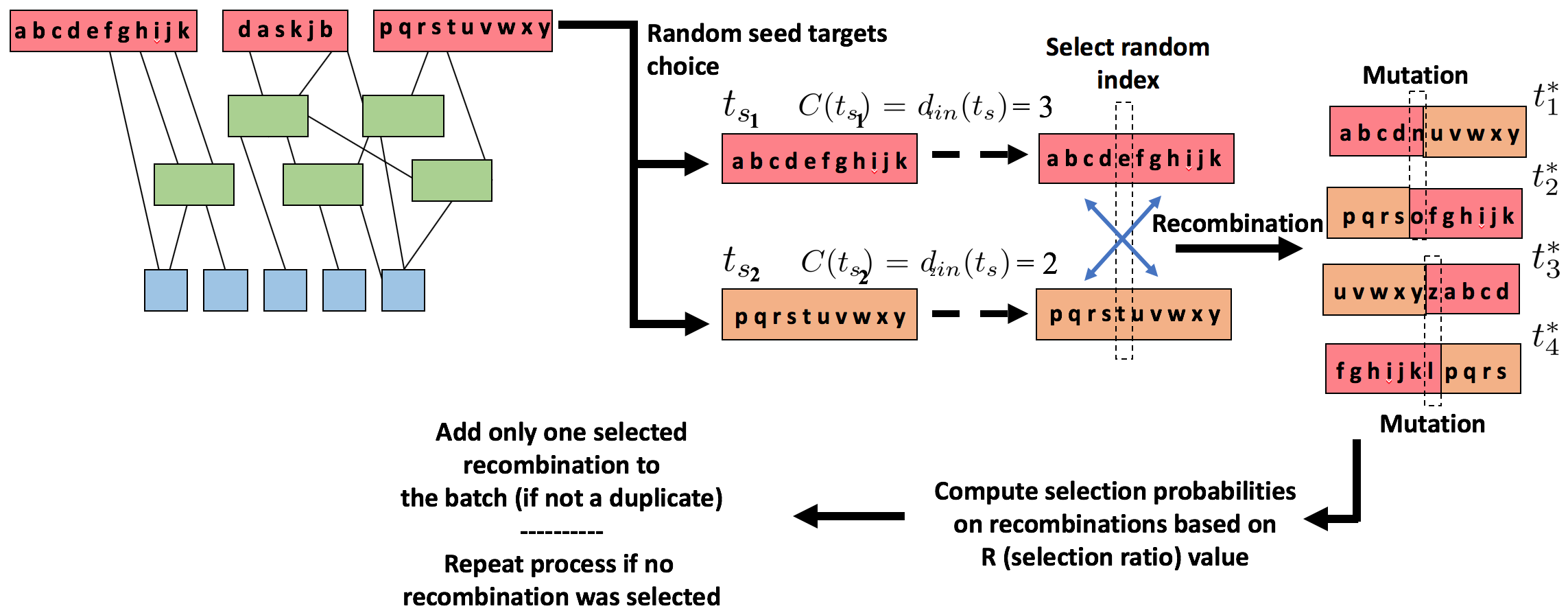}
\caption{Illustration of MRS Model}
\label{fig:TGM:MRS}
\end{figure}
\begin{figure}
\centering
\includegraphics[trim = 0cm 7cm .2cm 4cm, clip, width=.6\textwidth]{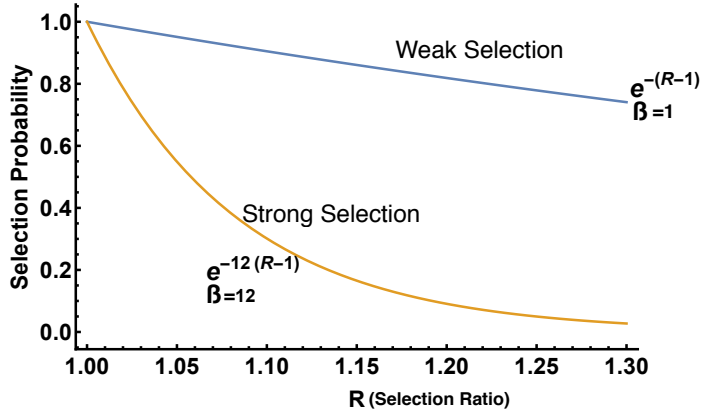}
\caption{The difference of the new target acceptance probability for weak ($\beta=1$) and strong ($\beta=12$) selection. $R$ is the ratio between the cost of the new candidate target and the cost of the targets it evolved from. In MRS-weak, the probability of accepting the new target is high. However, this probability quickly drops in the MRS-strong model.
}
\label{fig:exp_plot}
\end{figure}

\begin{figure}[h]
\center
\subfloat[Normalized Cost\label{fig:results-2:cost}]{
  \includegraphics[trim = 0.1cm 0.1cm 0.1cm 0.1cm, clip, width=0.4\textwidth]{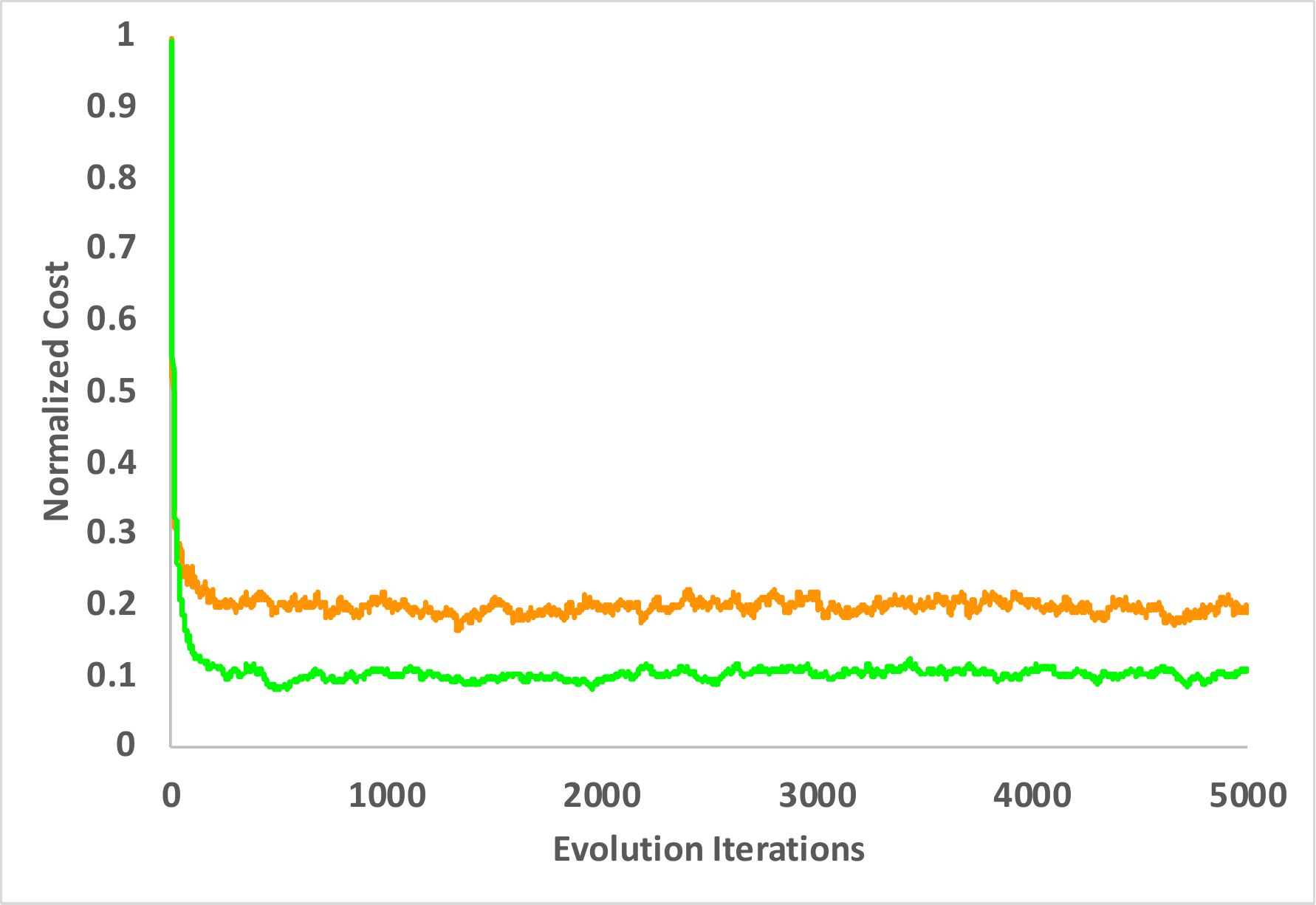}
}
\hspace{1cm}
\subfloat[Normalized Cost\label{fig:results-1:cost}]{
  \includegraphics[trim = 0.1cm 0.1cm 0.1cm 0.1cm, clip, width=0.4\textwidth]{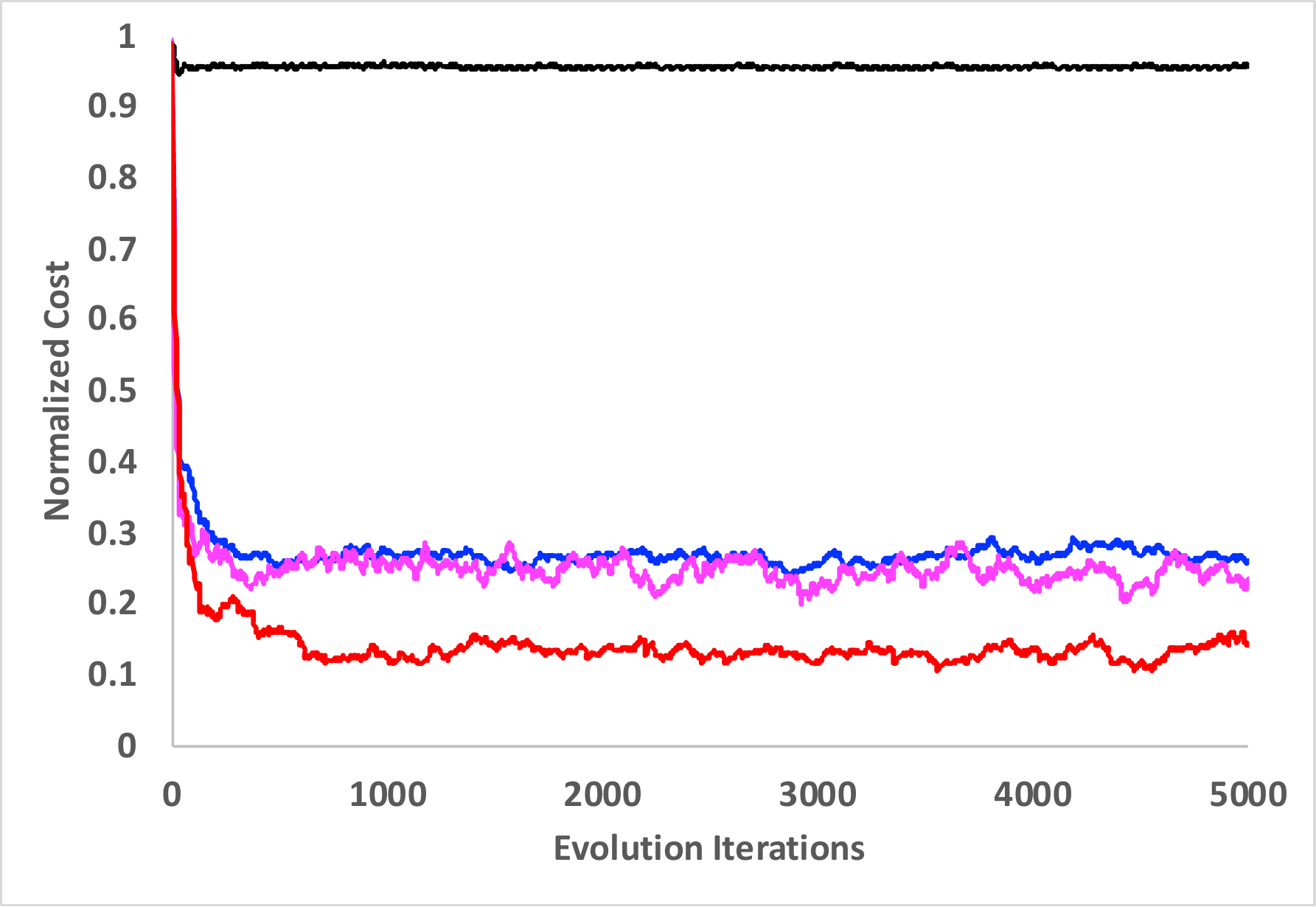}
}
\\\subfloat[Average Depth\label{fig:results-2:depth}]{
  \includegraphics[trim = 0.1cm 0.1cm 0.1cm 0.1cm, clip, width=0.4\textwidth]{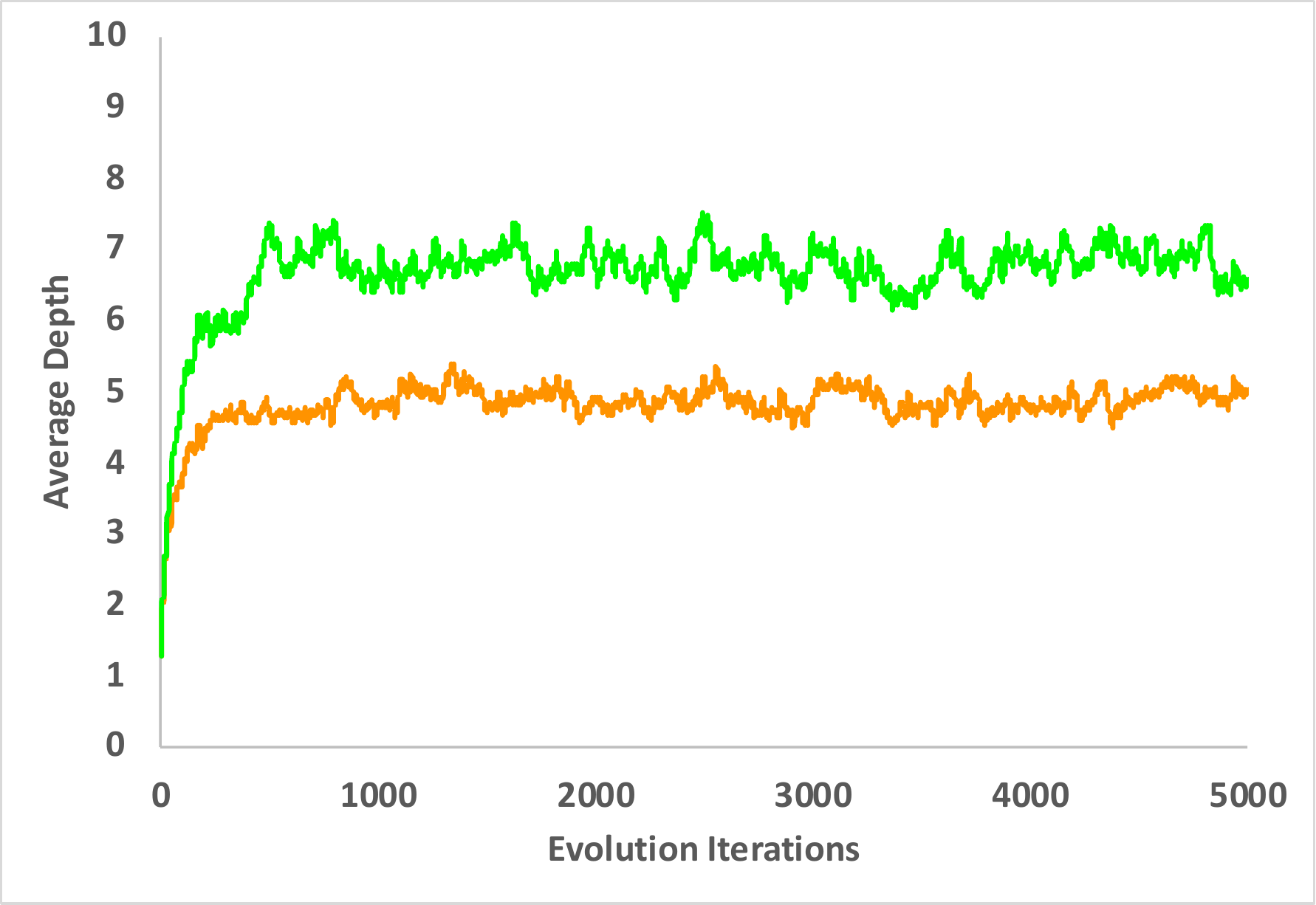}
}
\hspace{1cm}\subfloat[Average Depth\label{fig:results-1:depth}]{
  \includegraphics[trim = 0.1cm 0.1cm 0.1cm 0.1cm, clip, width=0.4\textwidth]{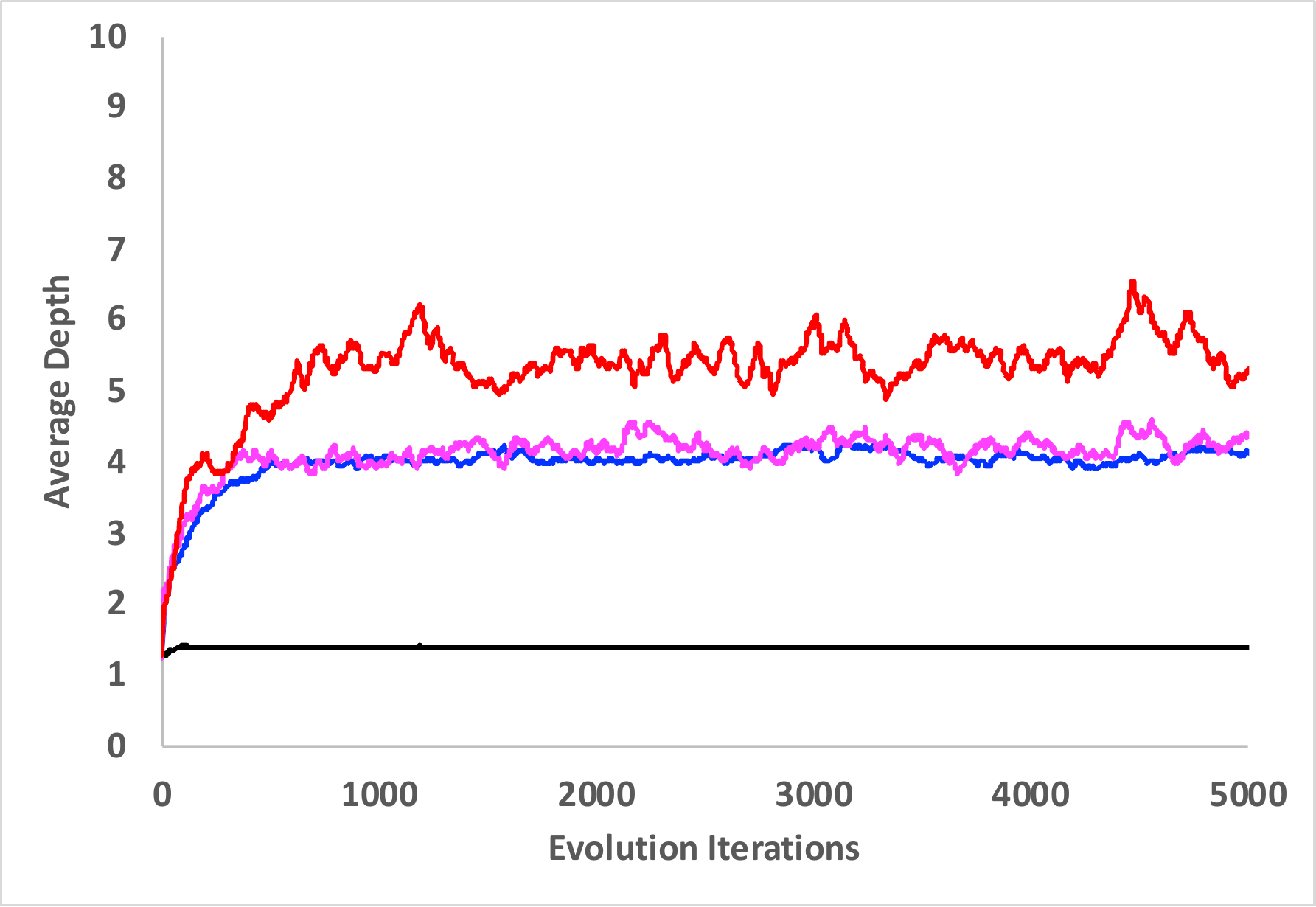}
}
\\
\subfloat[Node Length\label{fig:results-2:length}]{
  \includegraphics[trim = 0.1cm 0.1cm 0.1cm 0.1cm, clip, width=0.4\textwidth]{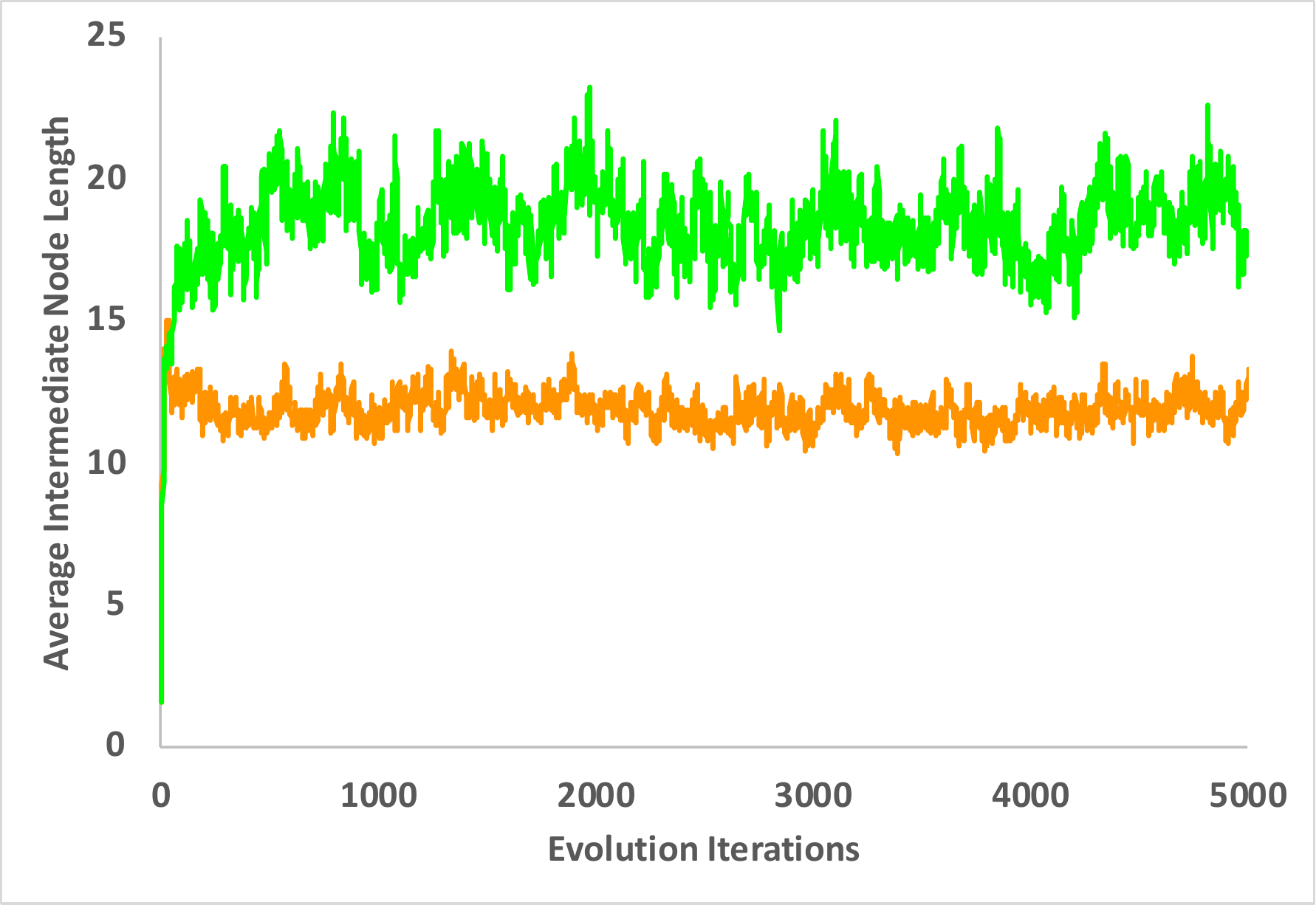}
}
\hspace{1cm}
\subfloat[Node Length\label{fig:results-1:length}]{
  \includegraphics[trim = 0.1cm 0.1cm 0.1cm 0.1cm, clip, width=0.4\textwidth]{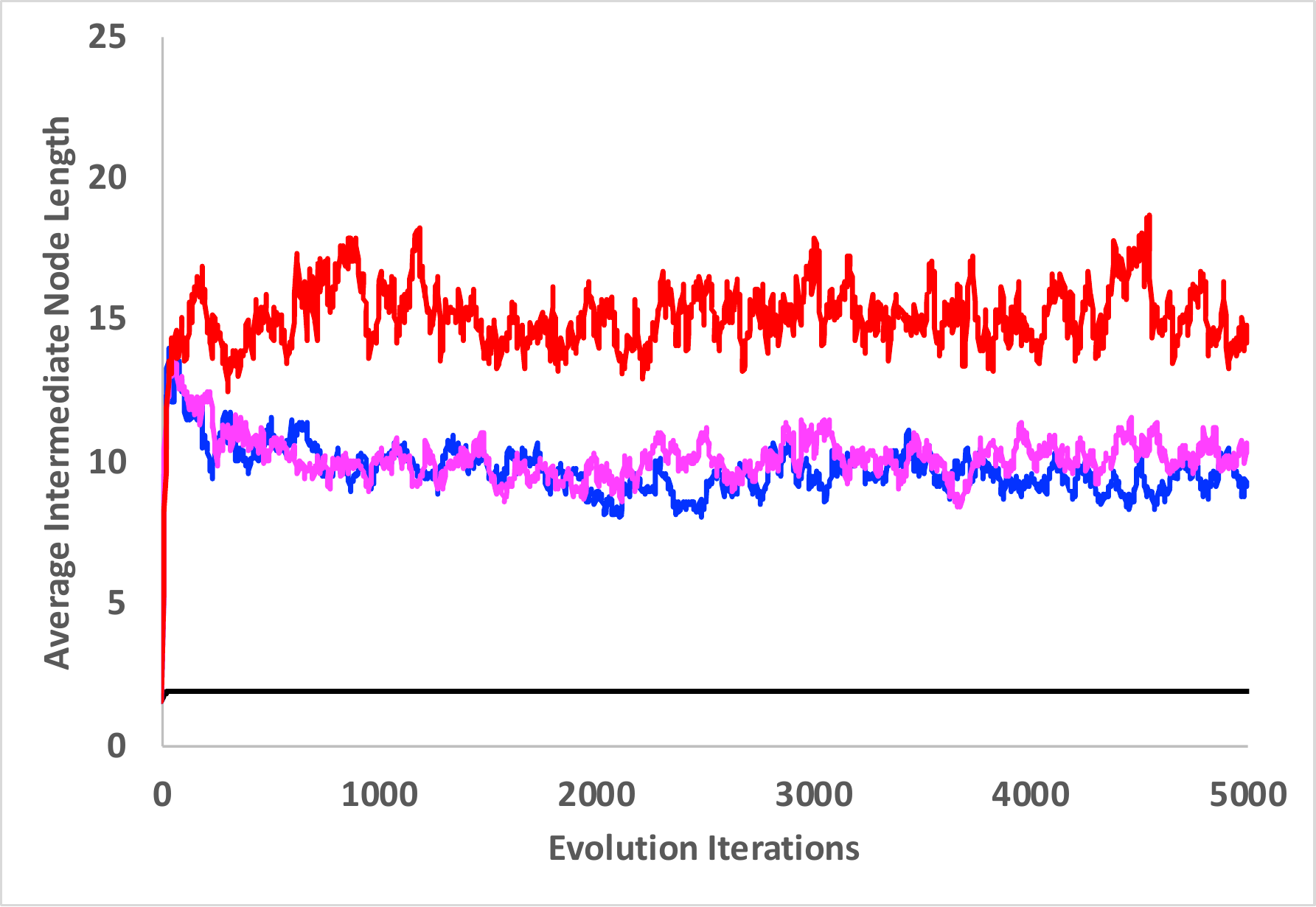}
}
\\
\subfloat[Target Diversity\label{fig:results-2:diversity}]{
  \includegraphics[trim = 0.1cm 0.1cm 0.1cm 0.1cm, clip, width=0.4\textwidth]{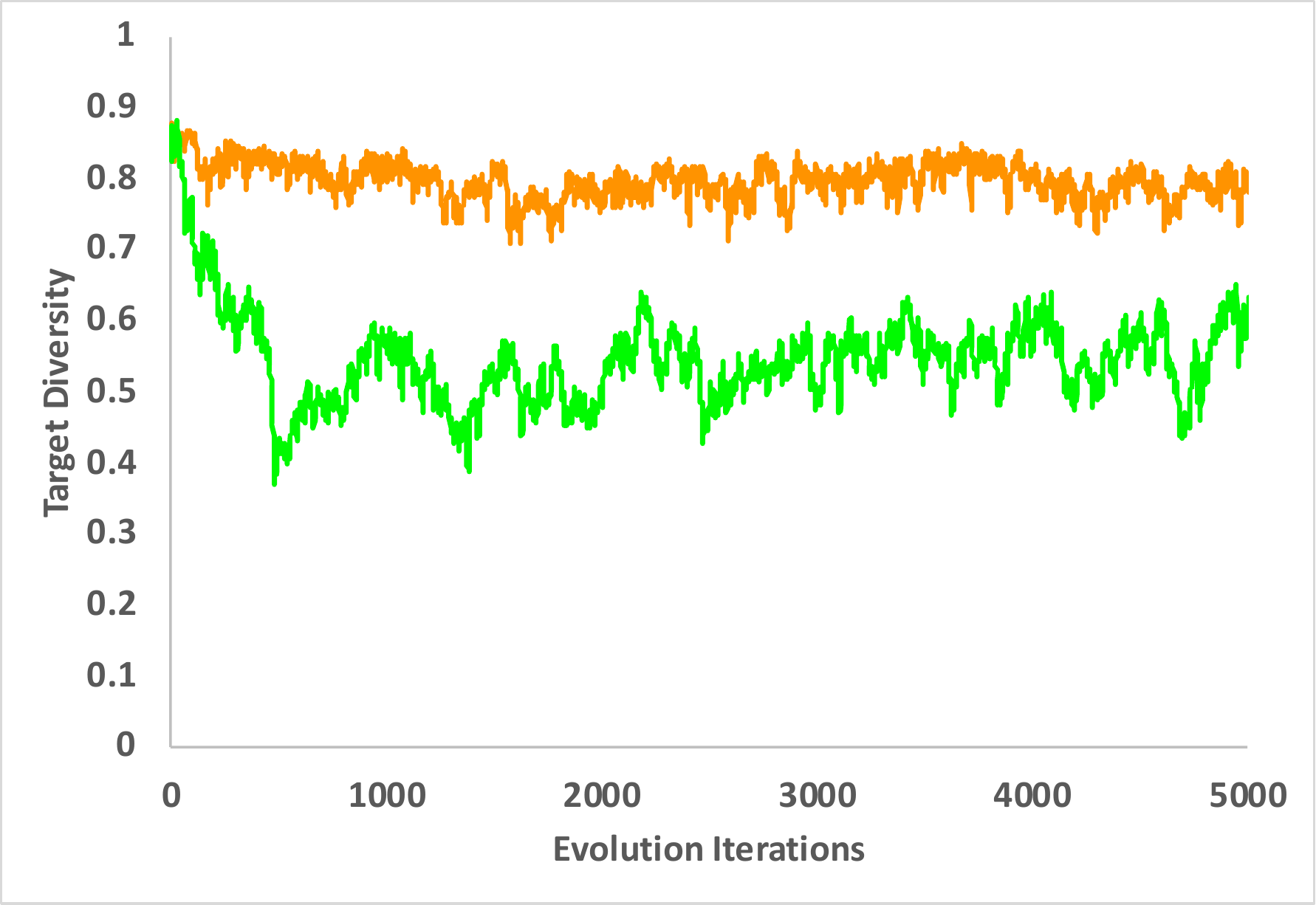}
}
\hspace{1cm}\subfloat[Target Diversity\label{fig:results-1:diversity}]{
  \includegraphics[trim = 0.1cm 0.1cm 0.1cm 0.1cm, clip, width=0.4\textwidth]{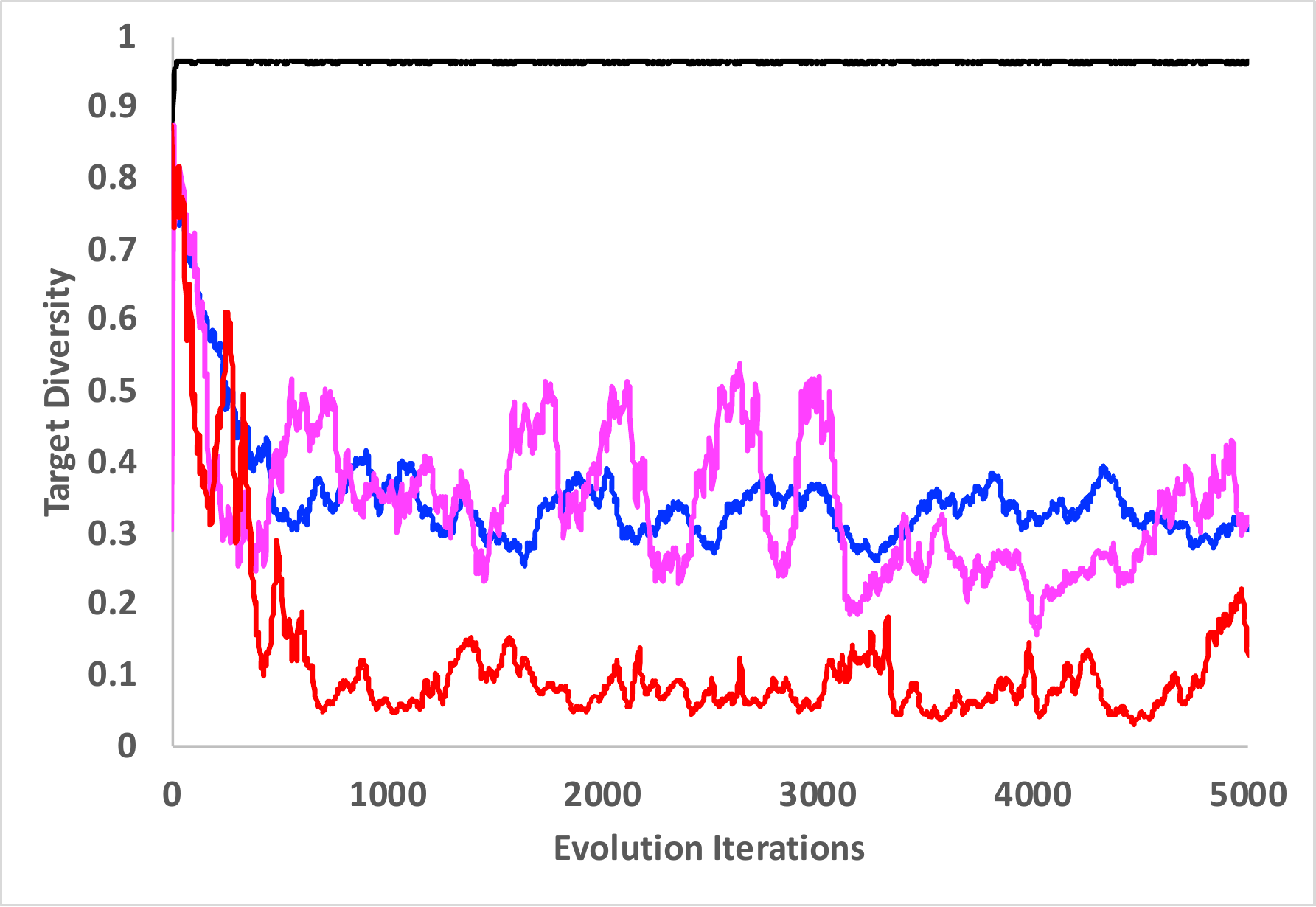}
}
\\
\includegraphics[trim = 1cm 13.5cm .5cm 13cm, clip, width=.8\textwidth]{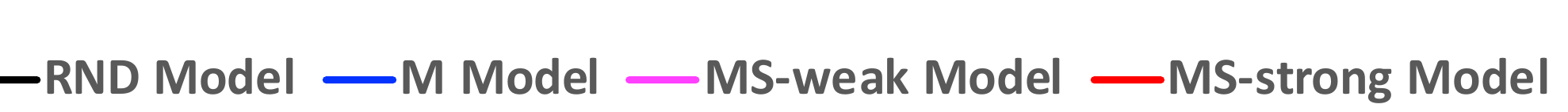}
\\
\includegraphics[trim = 1cm 13cm .5cm 13cm, clip, width=.55\textwidth]{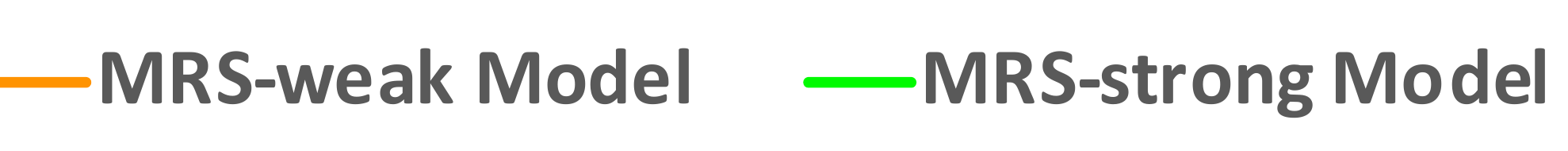}
\caption{Normalized Cost, (average) Hierachical Depth, (average) Intermediate Node Length, and Target diversity of Lexis-DAGs produced by various target generation models (weak selection models: $\beta=1$, strong selection models: $\beta=12$). (Continued in Fig. \ref{fig:results-1})
}
\label{fig:results-2}
\end{figure}

\begin{figure}[h]
\centering
\subfloat[Core Size\label{fig:results-2:core}]{
  \includegraphics[trim = 0.1cm 0.1cm 0.1cm 0.1cm, clip, width=0.4\textwidth]{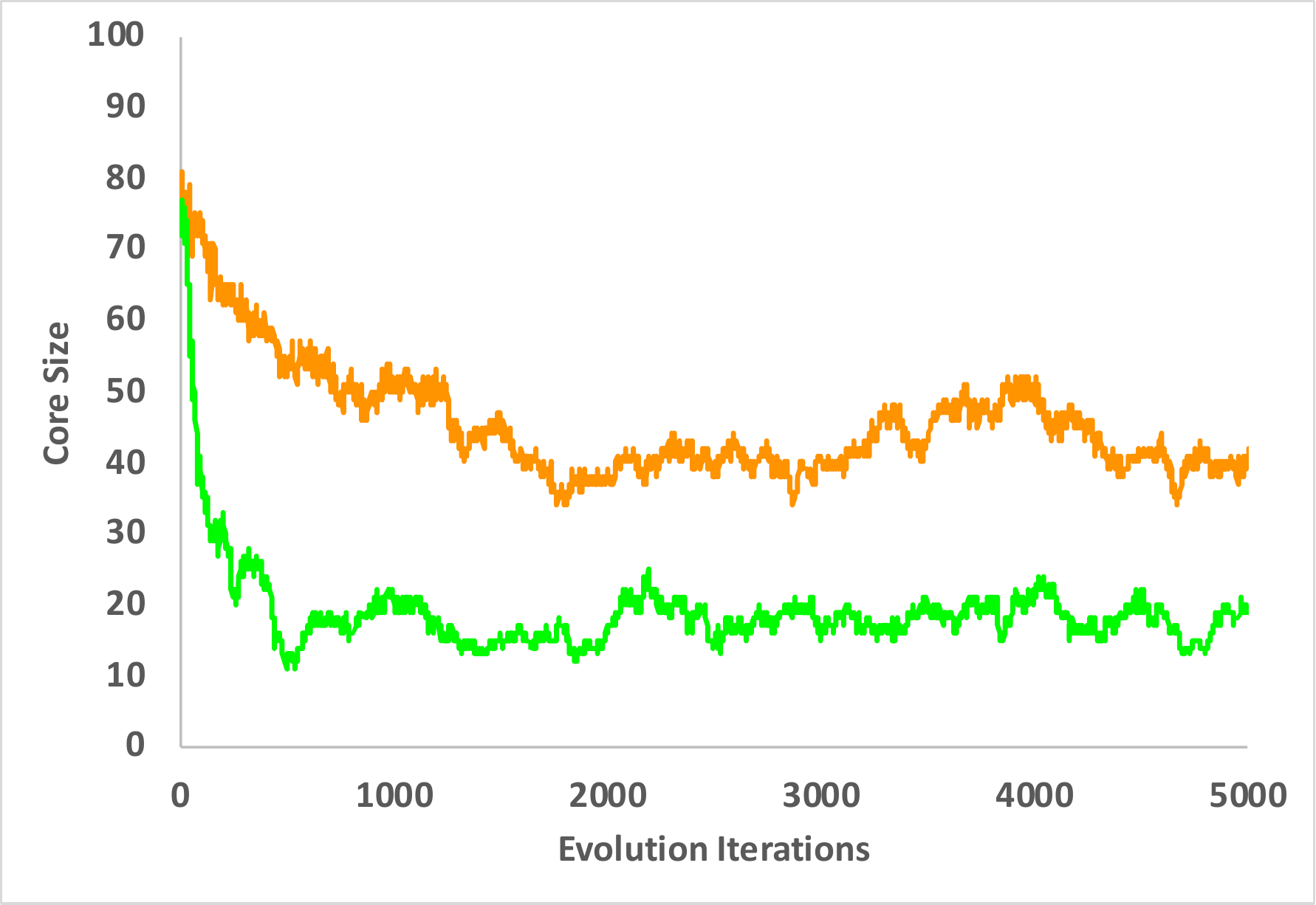}
}
\hspace{1cm}\subfloat[Core Size\label{fig:results-1:core}]{
  \includegraphics[trim = 0.1cm 0.1cm 0.1cm 0.1cm, clip, width=0.45\textwidth]{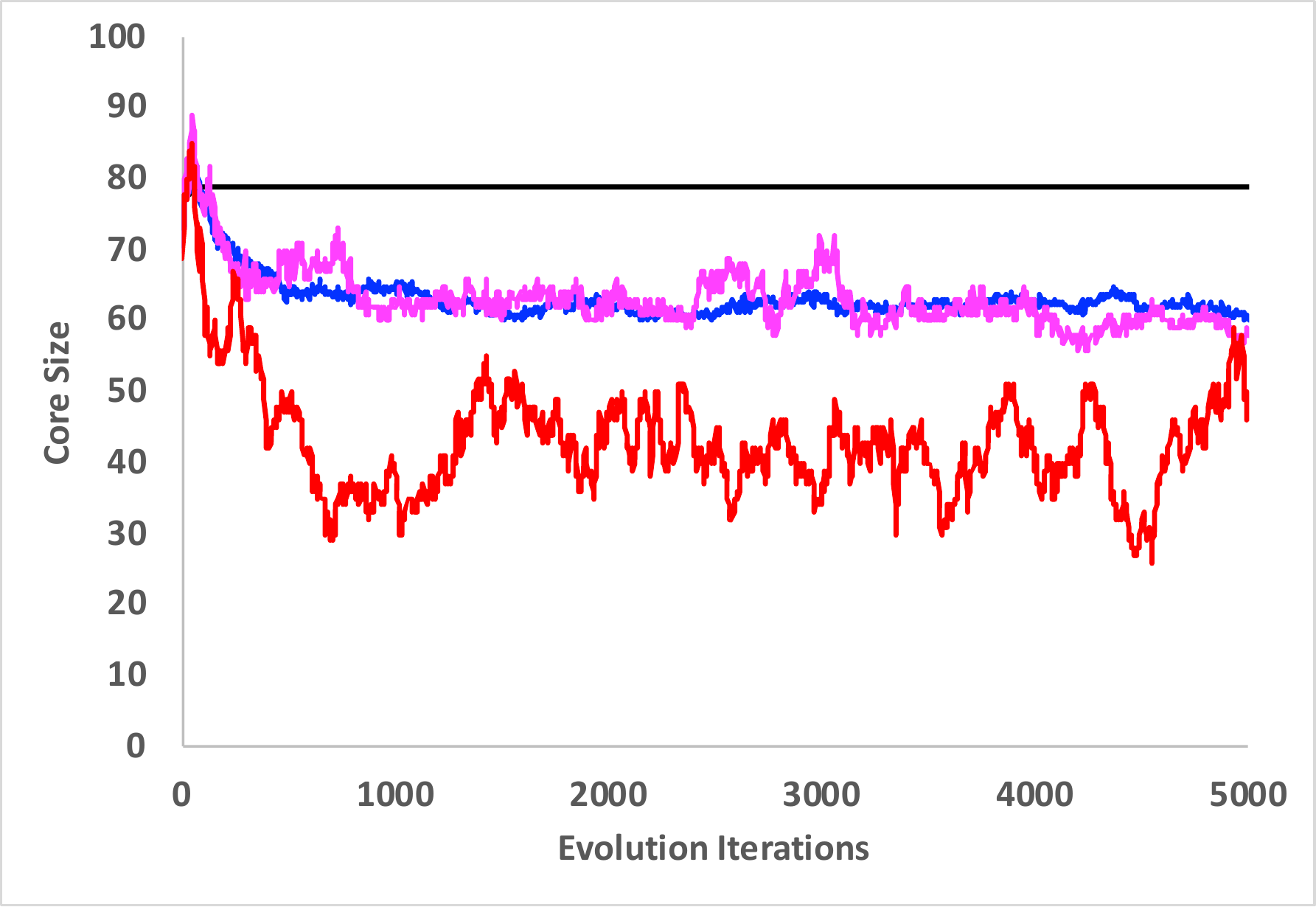}
}
\\
\subfloat[H-Score\label{fig:results-2:hscore}]{
  \includegraphics[trim = 0.1cm 0.1cm 0.1cm 0.1cm, clip, width=0.4\textwidth]{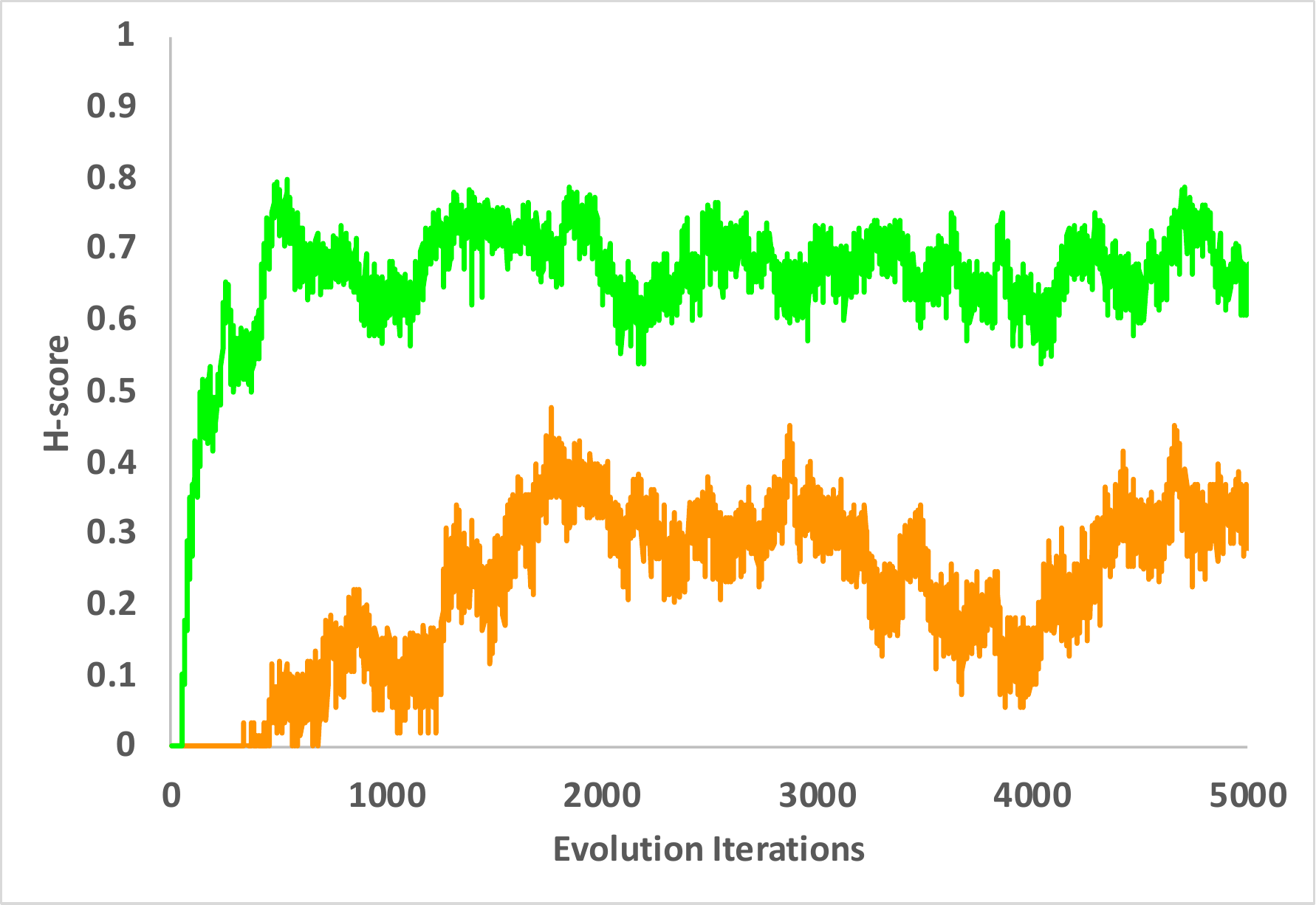}
}
\hspace{1cm}\subfloat[H-Score\label{fig:results-1:hscore}]{
  \includegraphics[trim = 0.1cm 0.1cm 0.1cm 0.1cm, clip, width=0.4\textwidth]{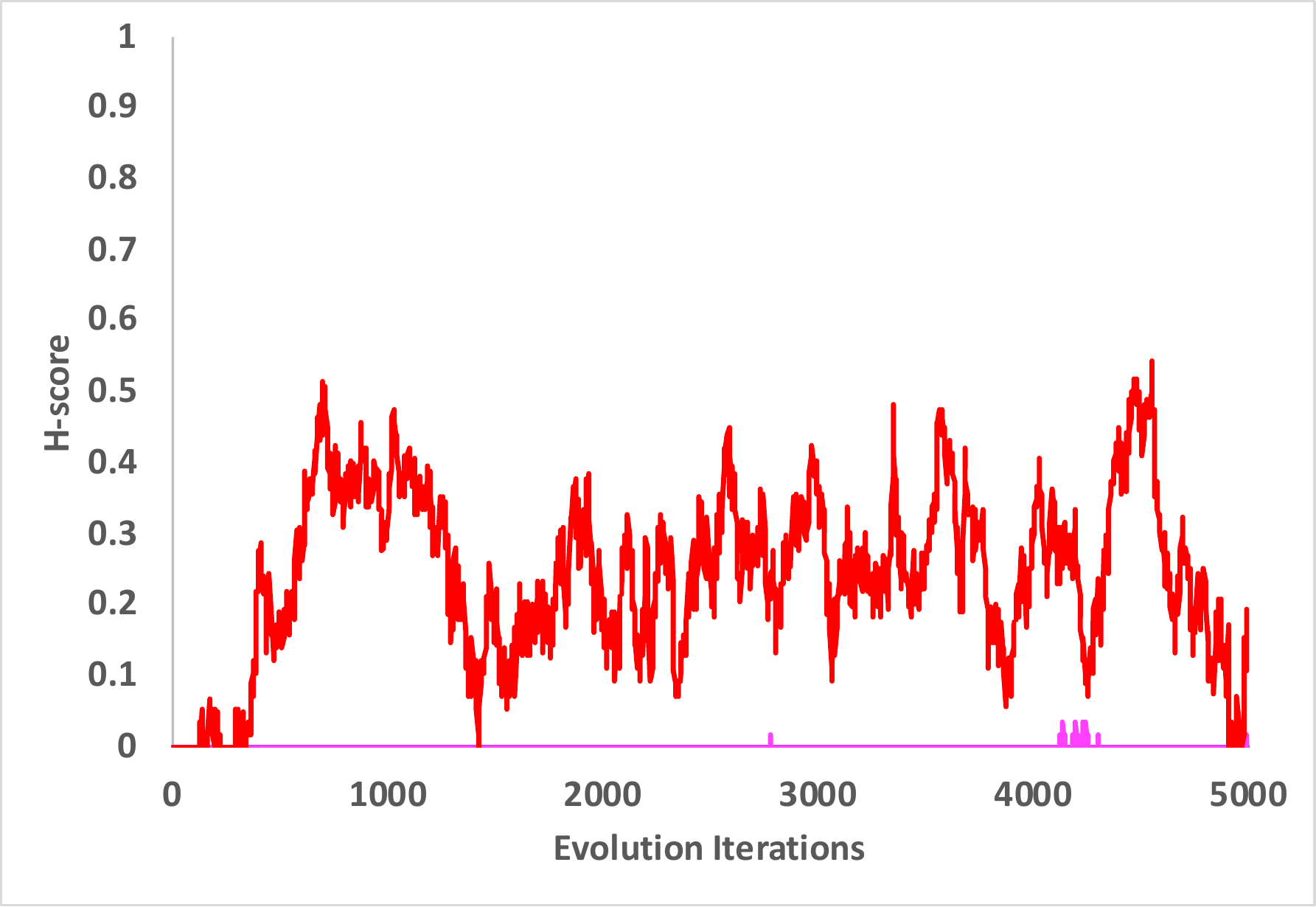}
}
\\
\subfloat[Robustness Analysis\label{fig:results-2:robustness}]{
  \includegraphics[trim = 0.1cm 0.1cm 0.1cm 0.1cm, clip, width=0.4\textwidth]{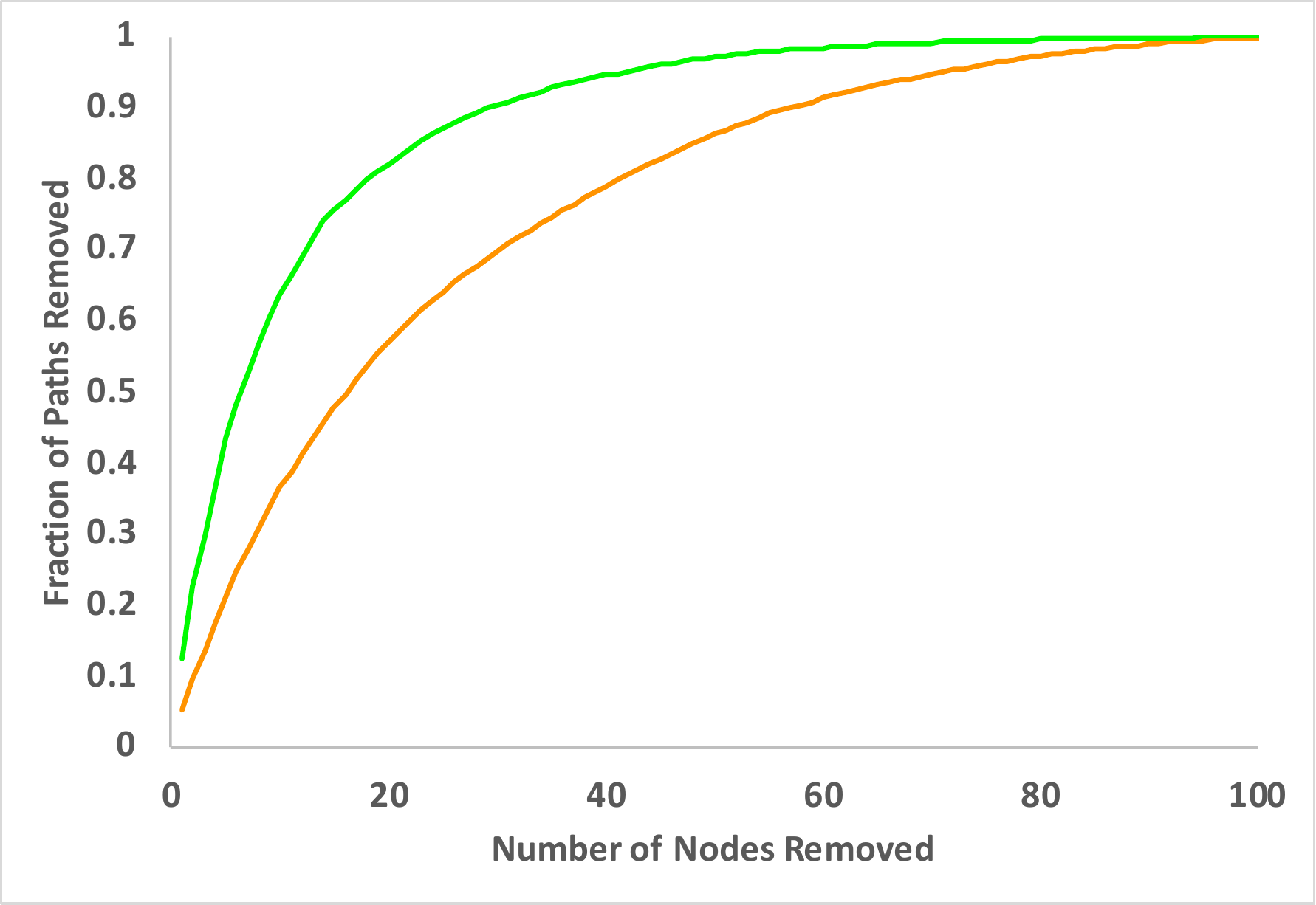}
}
\hspace{1cm}
\subfloat[Robustness Analysis\label{fig:results-1:robustness}]{
  \includegraphics[trim = 0.1cm 0.1cm 0.1cm 0.1cm, clip, width=0.4\textwidth]{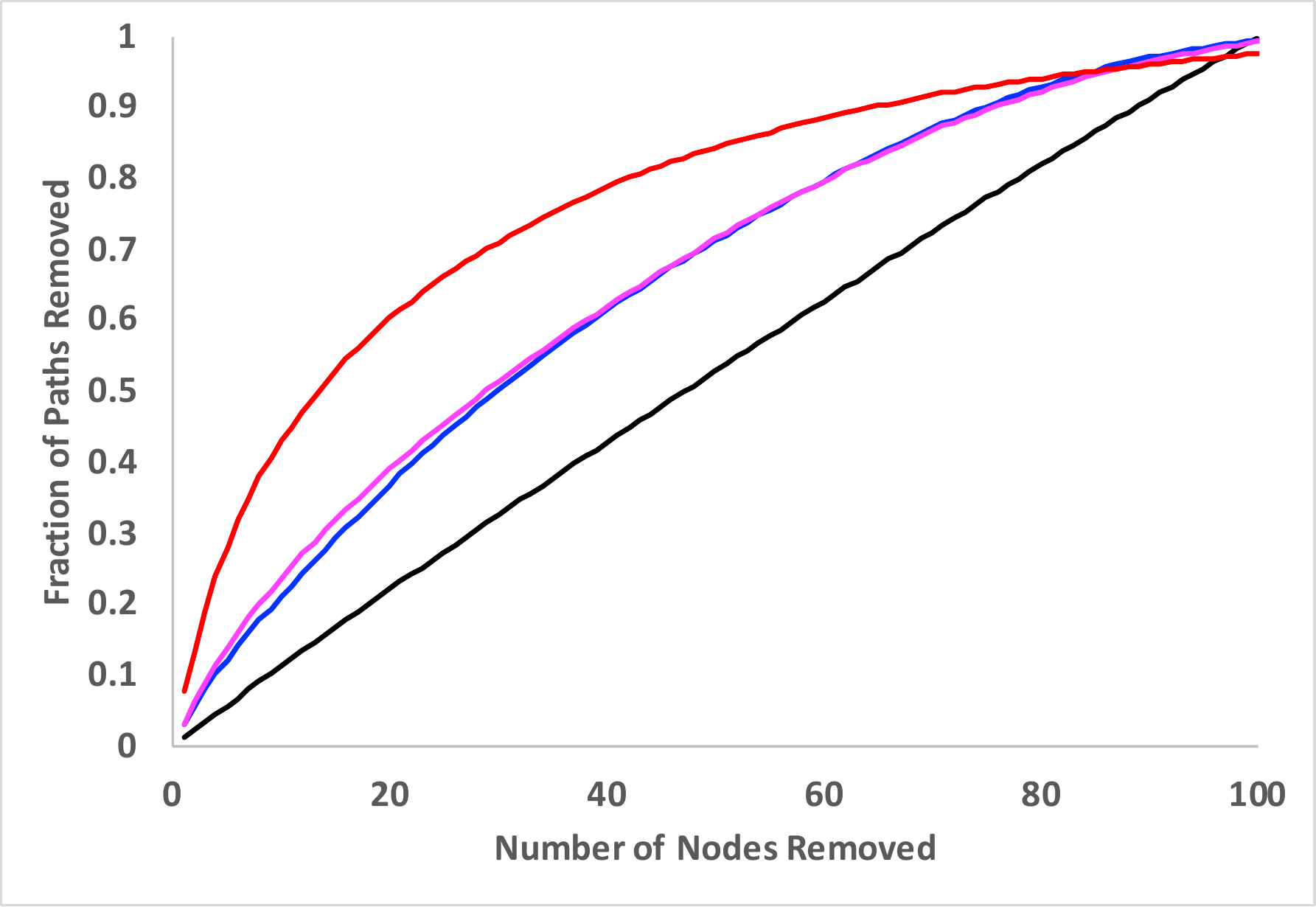}
}
\\
\subfloat[Core Stability\label{fig:results-2:stability}]{
  \includegraphics[trim = 0.1cm 0.1cm 0.1cm 0.1cm, clip, width=0.4\textwidth]{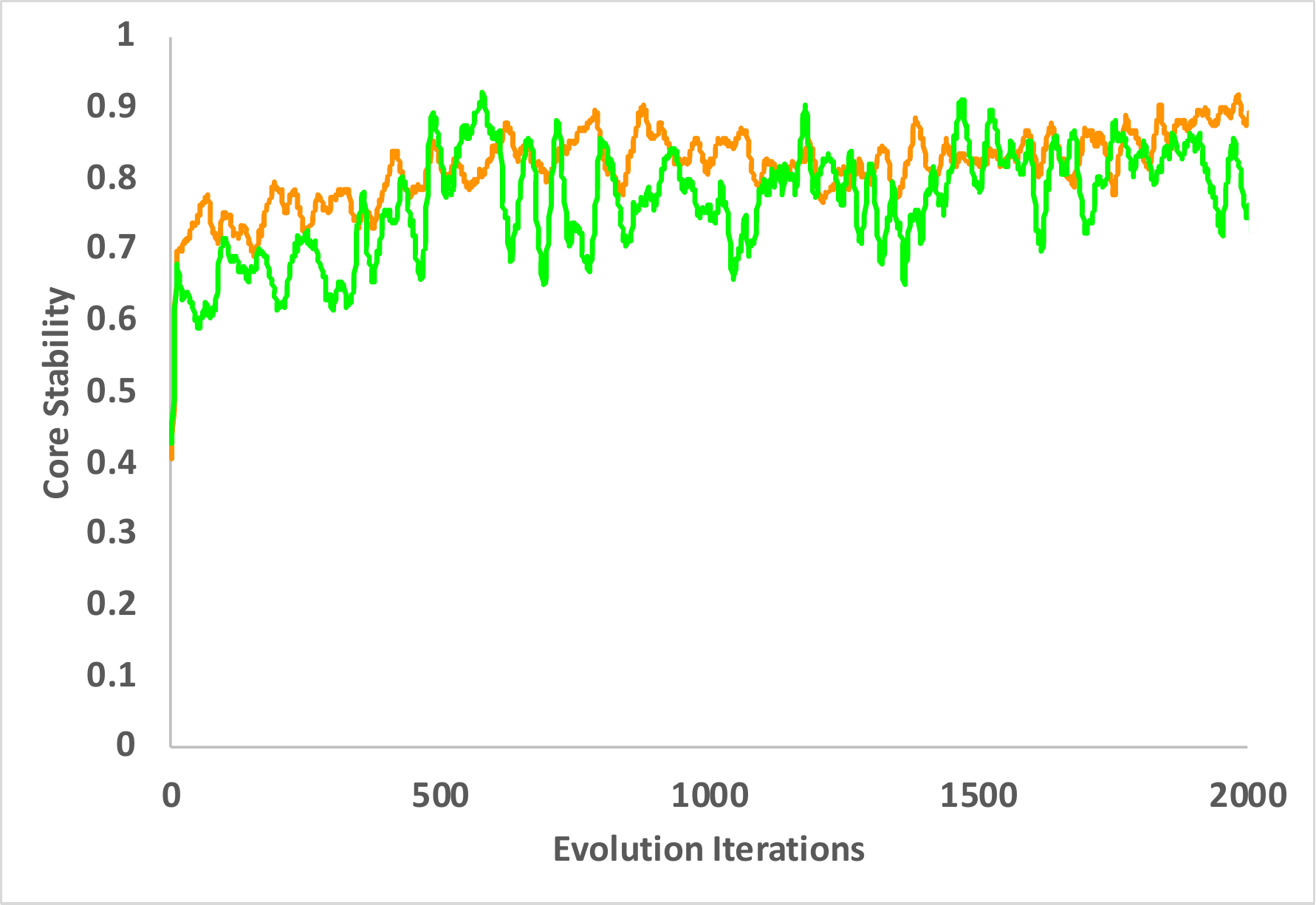}
}
\hspace{1cm}\subfloat[Core Stability\label{fig:results-1:stability}]{
  \includegraphics[trim = 0.1cm 0.1cm 0.1cm 0.1cm, clip, width=0.4\textwidth]{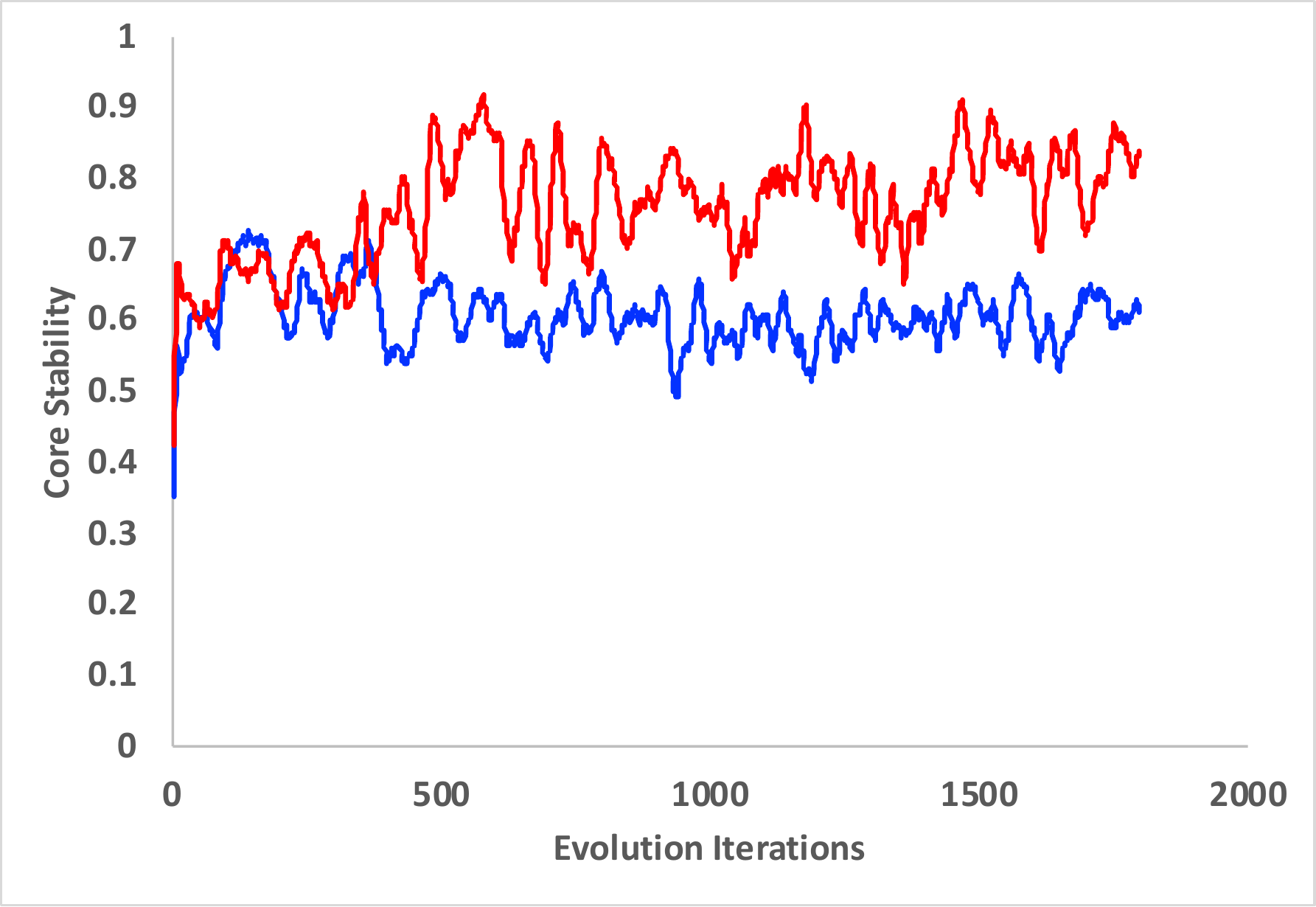}
}
\\
\includegraphics[trim = 1cm 13.5cm .5cm 13cm, clip, width=.8\textwidth]{images/legend-1.pdf}
\\
\includegraphics[trim = 1cm 13cm .5cm 13cm, clip, width=.55\textwidth]{images/legend-5.pdf}
\caption{(Continued from Fig. \ref{fig:results-2})
Core size, H-score, Robustness to core node removals, and Core stability of Lexis-DAGs produced by various target generation models (weak selection models: $\beta=1$,  strong selection models: $\beta=12$). For core selection, we set $\tau=0.85$. For core stability, a sliding window equal to the size of 10 batches is used to track changes in the core set.)
}
\label{fig:results-1}
\end{figure}

The main target generation model we consider is based on {\bf M}utation, {\bf R}ecombination and {\bf S}election, thus called {\em MRS model}. The mechanism for this model is illustrated in Fig. \ref{fig:TGM:MRS}. In detail: 
\begin{enumerate}
\item 
Two distinct targets $t_{s_1}$ and $t_{s_2}$ (referred to as ``seeds'') are chosen randomly from the existing set of targets. Their cost is denoted by $C(t_{s_1}) = d_{in}(t_{s_1})$ and $C(t_{s_2}) = d_{in}(t_{s_2})$, respectively,
and it is equal to the number of incoming edges that form $t_s$ from the intermediate nodes in the current Lexis-DAG.
\item
A randomly chosen ``crossover index'' $1\leq i \leq k-1$ is chosen (recall that $k$ is the length of the targets) and the following recombinations are generated:
\begin{itemize}
\item
$t^*_1 = t_{s_1}[1:i-1] + \widetilde{c} + t_{s_2}[i+1:k]$
\item
$t^*_2 = t_{s_2}[1:i-1] + \widetilde{c} + t_{s_1}[i+1:k]$
\item
$t^*_3 = t_{s_2}[i+1:k] + \widetilde{c} + t_{s_1}[1:i-1]$
\item
$t^*_4 = t_{s_1}[i+1:k] + \widetilde{c} + t_{s_2}[1:i-1]$
\end{itemize}
where the numbers in braces show string indices, and $\widetilde{c}$ is a randomly chosen character that represents the mutated element. In other words, 
each recombination also includes a single-character mutation. 
\item
For each of the four recombinations, we calculate its cost when it is added as a new target to the current Lexis-DAG. This cost can be seen as the marginal overhead that $t_x^*$ introduces when added to the current hierarchy $D_0$:
\begin{equation}
C(t_x^*) = \mathcal{E}\left(D^{\text{\textsc{Inc}}}(D_0,\{t_x^*\})\right) - \mathcal{E}(D_0)
\end{equation}
where $D^{\text{\textsc{Inc}}}(D_0,\{t_x^*\})$ is the new hierarchy 
after adding $t_x^*$ to $D_0$ using the INC-Lexis algorithm.
\item
The model selects a newly generated recombination $t_x^*$ if it satisfies the following selection constraint:
\begin{itemize}
\item 
Suppose $t_x^*$ is formed by recombining  the fragments $t_{x_1}$ (from $t_{s_1}$) and $t_{x_2}$ (from $t_{s_2}$), where the length of these target fragments are $|t_{x_1}|$ and $|t_{x_2}|$. \\
The \emph{selection ratio} is defined as:
\begin{equation}
R = \frac{C(t^*_x)}{|t_{x_1}| \times C(t_{s_1}) + |t_{x_2}| \times C(t_{s_2})}
\end{equation}
\item
If $R\leq 1$, we definitely accept $t_x^*$.
\item
If $R>1$, we accept $t_x^*$ probabilistically with \emph{selection probability} $p=e^{-\beta (R-1)}$.
\end{itemize}
\item
If none of the recombinations passes the previous selection constraint, the target generation process is repeated. However, if one or more recombinations pass the selection constraint, the model chooses one of them randomly and adds it as an accepted target in the batch of new targets.
\end{enumerate}

$\beta$ determines how strongly the current hierarchy influences the selection of new targets. The larger the parameter $\beta$ is, the less likely it becomes that a new target that is more costly than its seeds (i.e. $R > 1$) will be selected. For large $\beta$, we get \emph{Strong Selection} and  refer to the model as {\em MRS-strong}. A small $\beta$ implies \emph{Weak Selection}, and the model is referred to as {\em MRS-weak}. 
We use $\beta=1$ and $\beta=12$ for weak and strong selection, respectively. Fig. \ref{fig:exp_plot} shows the difference of the two $\beta$ values for typical values of $R$ (when $R>1$).

To analyze the effect of each evolutionary mechanism, we also consider target generation models by removing certain elements from the MRS model -- hence the name ``ablation study''. 

\subsubsection{MS Model}
The MS model is derived from MRS by removing recombination (hence the name {\bf M}utations+{\bf S}election Model or MS Model). 
The model generates new targets as follows:
\begin{enumerate}
\item 
A target seed $t_s$ is chosen from the existing set of targets. Suppose the cost of $t_s$ is $C(t_s) = d_{in}(t_s)$ in the current Lexis-DAG $D_0$. 
\item
The seed is mutated (single character mutation), as in MRS model, to $t_s^*$.
\item
We calculate the cost of adding $t_s^*$ to the current Lexis-DAG. This cost can be seen as the marginal overhead that $t_s^*$ introduces when it is added to the current Lexis-DAG:
\begin{equation}
C(t_s^*) = \mathcal{E}\left(D^{\text{\textsc{Inc}}}(D_0,\{t_s^*\})\right) - \mathcal{E}(D_0)
\end{equation}
\item
The model will select the newly generated target $t_s^*$ if it satisfies the following constraint:
\begin{itemize}
\item 
$R = \frac{C(t_s^*)}{C(t_s)}$
\item
If $R\leq 1$, accept $t_s^*$.
\item
if $R>1$, accept $t_s^*$ probabilistically where selection probability $p=e^{-\beta (R-1)}$.
\end{itemize}
Otherwise, the newly generated target is rejected and the target generation repeats.
\end{enumerate}

\subsubsection{M Model}
This is derived from the MS model by removing the Selection constraint. Note that with this change the target generation process is not influenced by the current Lexis-DAG and it operates ``exogenously'' to the hierarchy. This model is referred to as {\bf M}utation Model (or M Model) and it generates targets as follows:
\begin{enumerate}
\item 
Among the targets that exist in the current Lexis-DAG, a seed target $t_s$ is chosen randomly.
\item
The seed target $t_s$ is mutated to $t_s^*$ through a random single character mutation.
\item
If the newly generated target $t_s^*$ is a duplicate of one of the existing targets, the new target is rejected and the target generation repeats. If not, the  generated target is added to the batch of new targets.
\end{enumerate}

\begin{figure}[h]
\center
\subfloat[RND Model\label{fig:scatters:random}]{
  \includegraphics[scale=0.25]{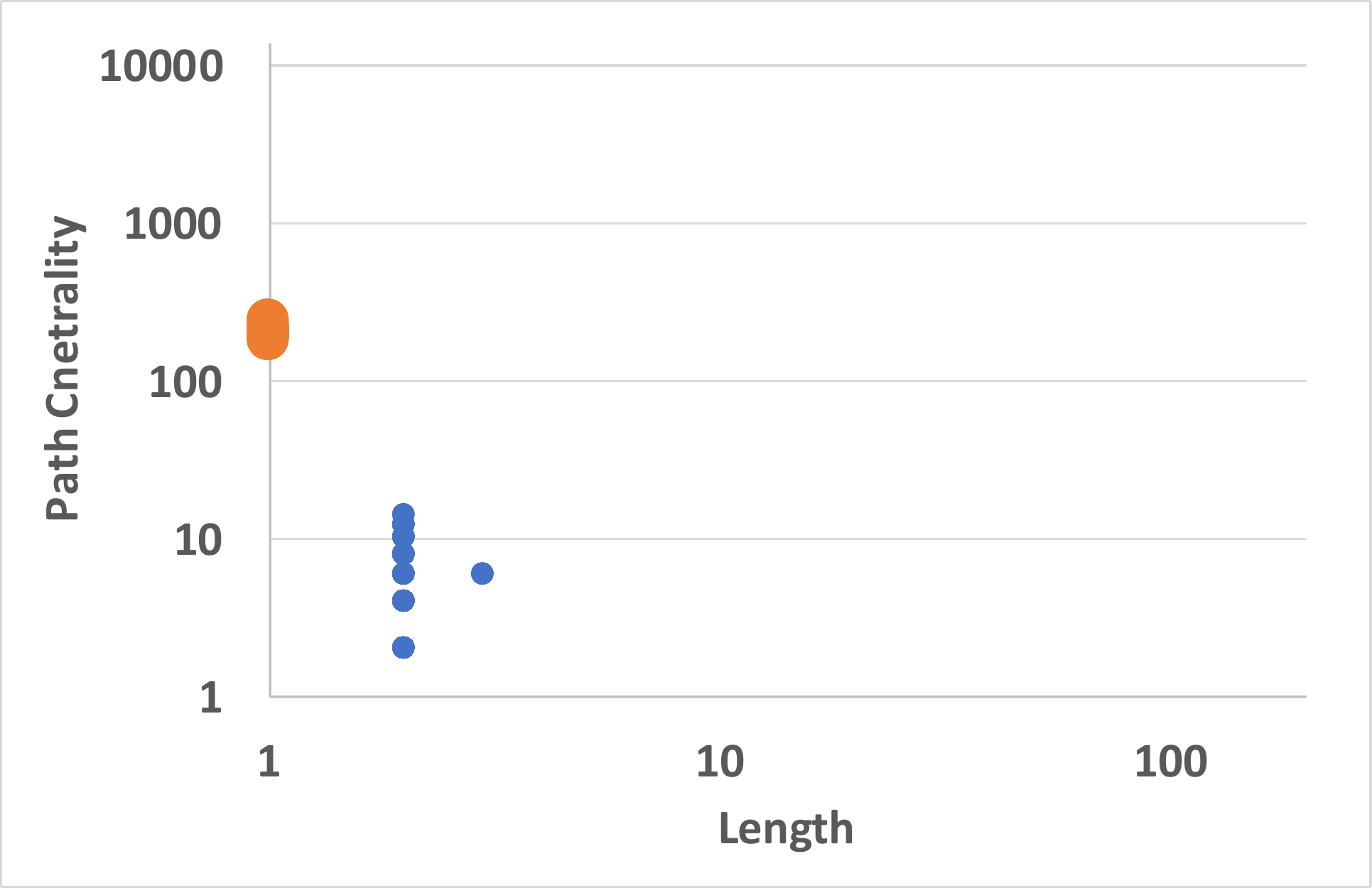}
}
\hspace{1cm}\subfloat[M Model\label{fig:scatters:mutation}]{
  \includegraphics[scale=0.25]{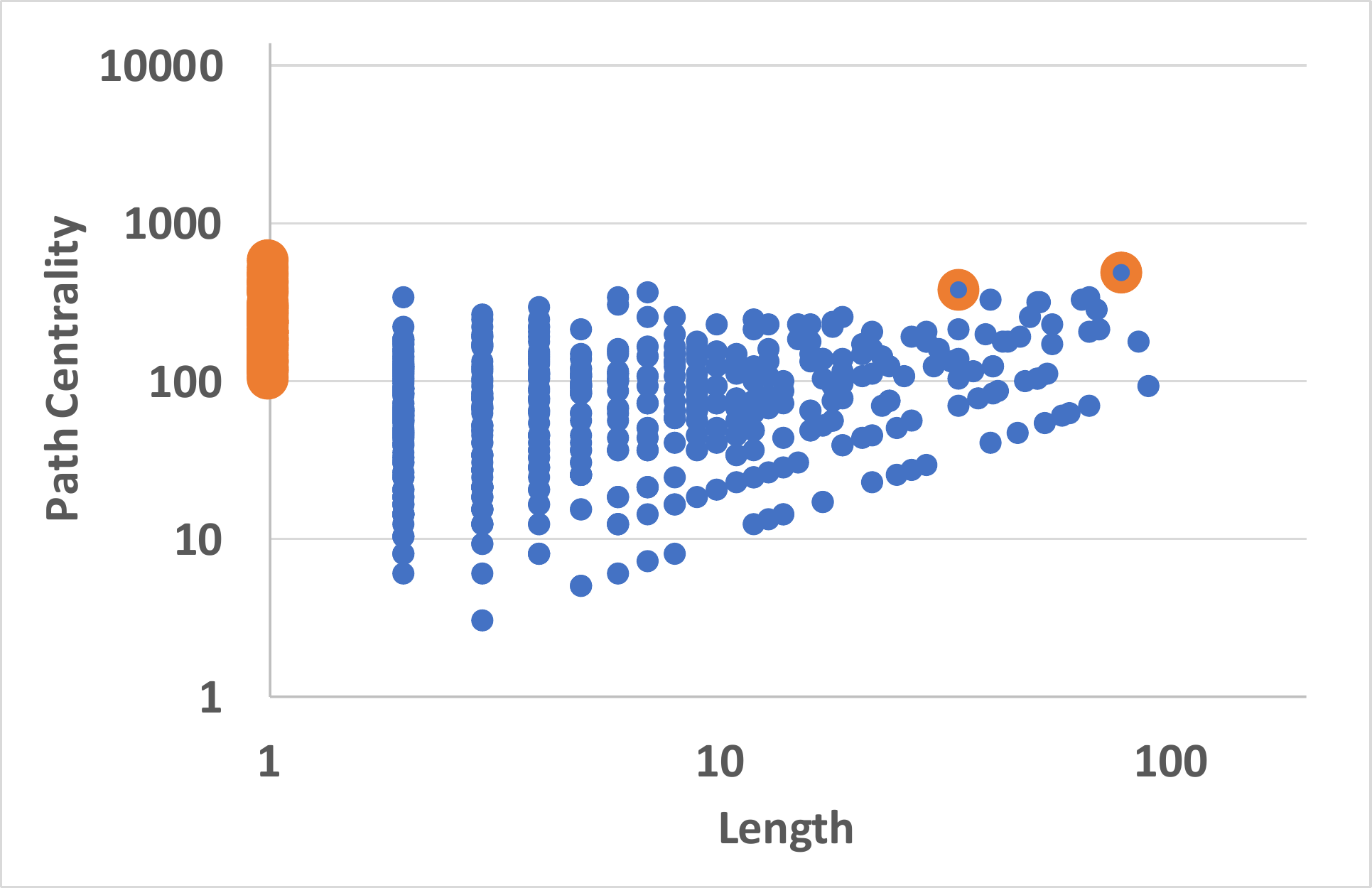}
}
\\
\subfloat[MS-weak\label{fig:scatters:mut+sel-weak}]{
  \includegraphics[scale=0.25]{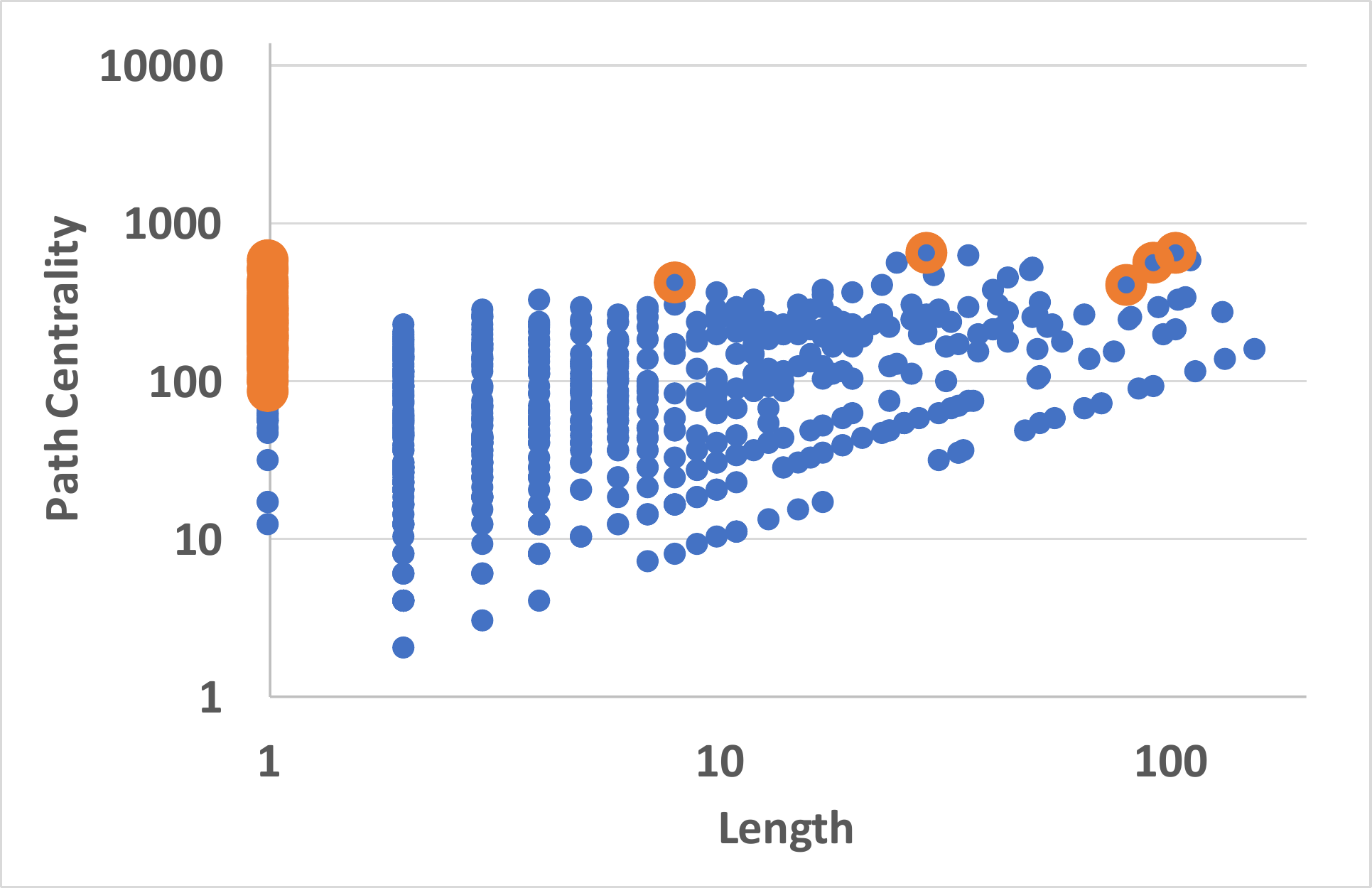}
}
\hspace{1cm}\subfloat[MS-strong\label{fig:scatters:mut+sel-strong}]{
  \includegraphics[scale=0.25]{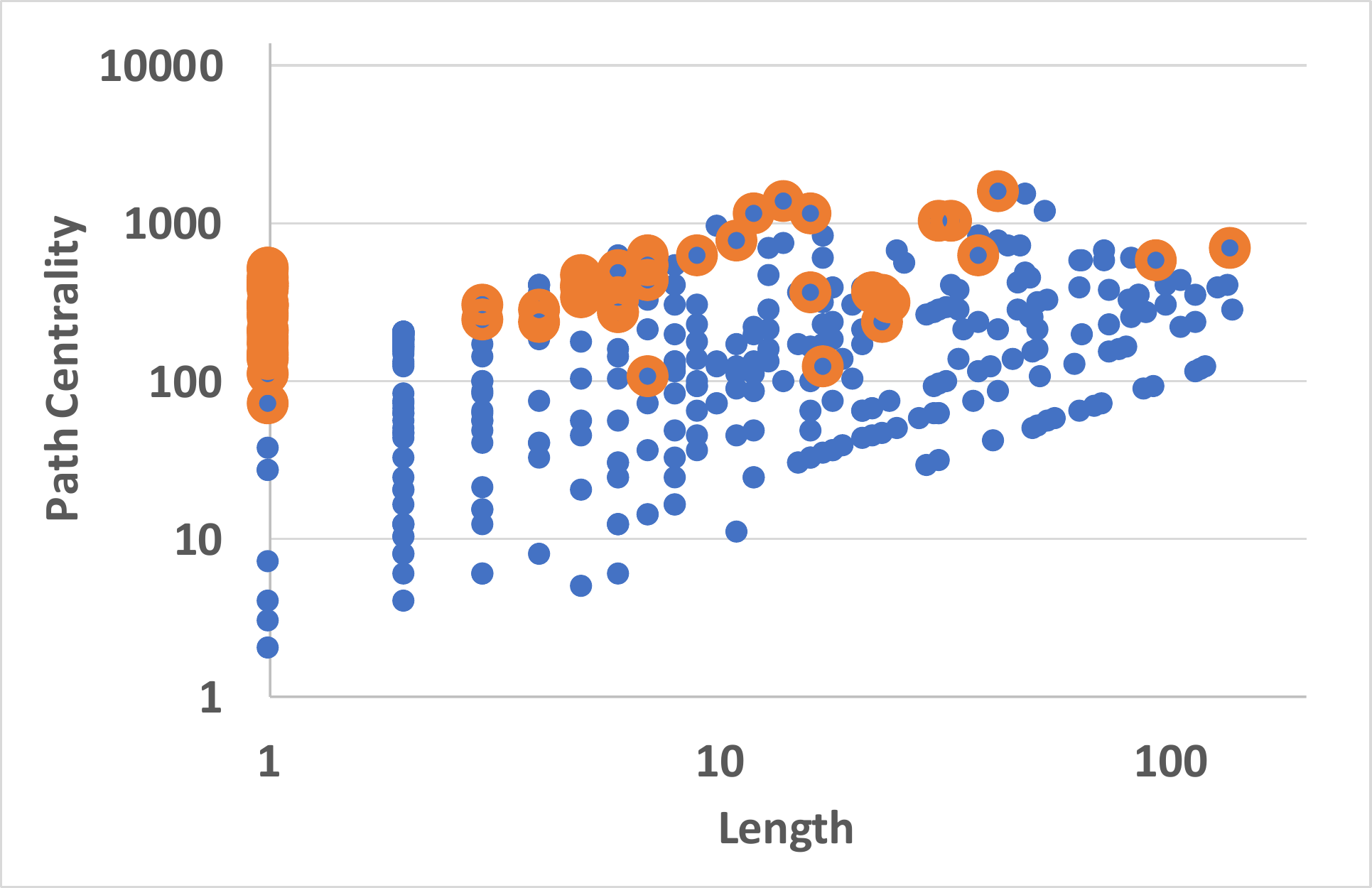}
}
\\
\subfloat[MRS-weak Model \label{fig:scatters:rec+sel-weak}]{
  \includegraphics[scale=0.25]{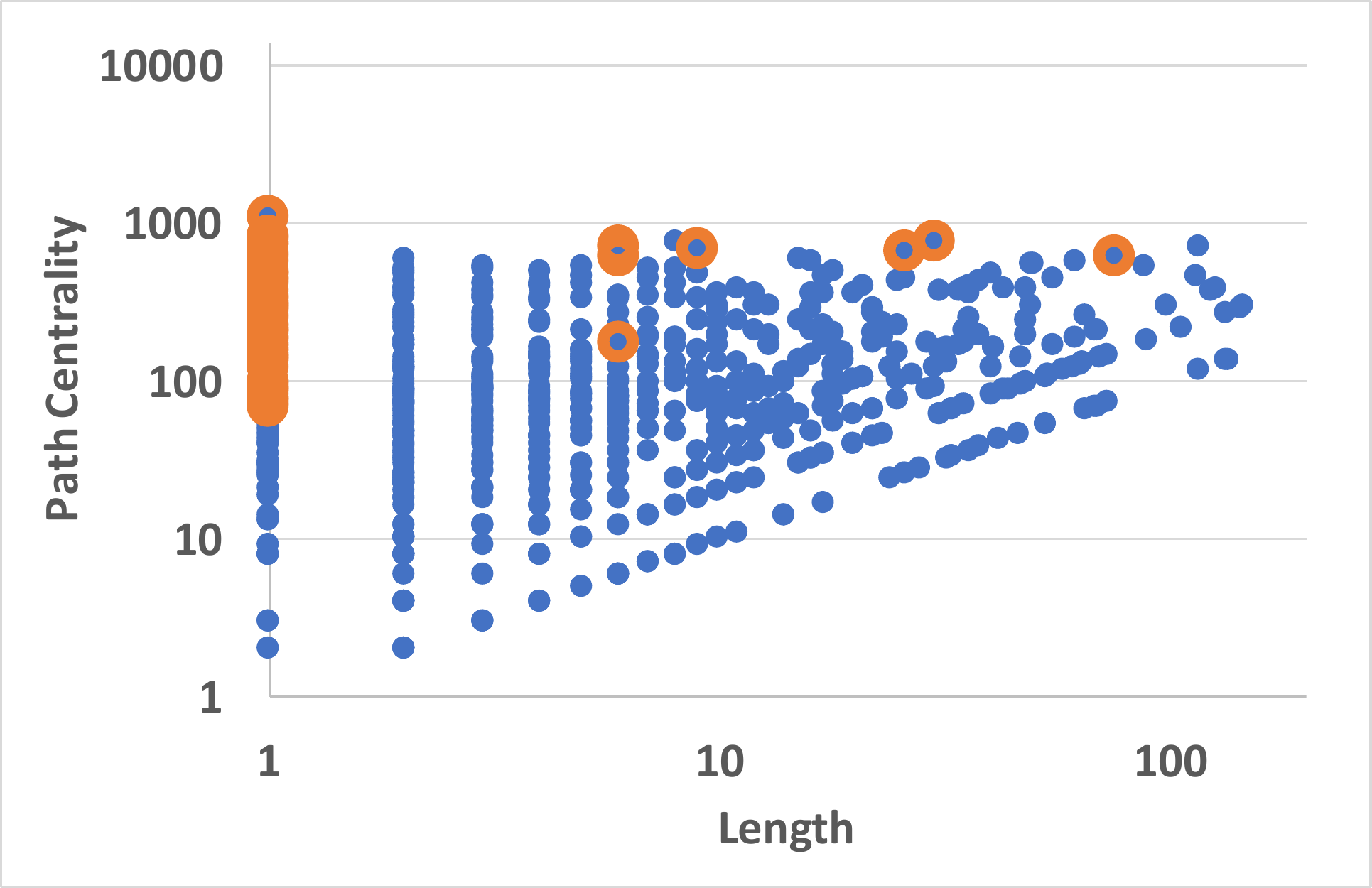}
}
\hspace{1cm}\subfloat[MRS-strong Model\label{fig:scatters:rec+sel-strong}]{
  \includegraphics[scale=0.25]{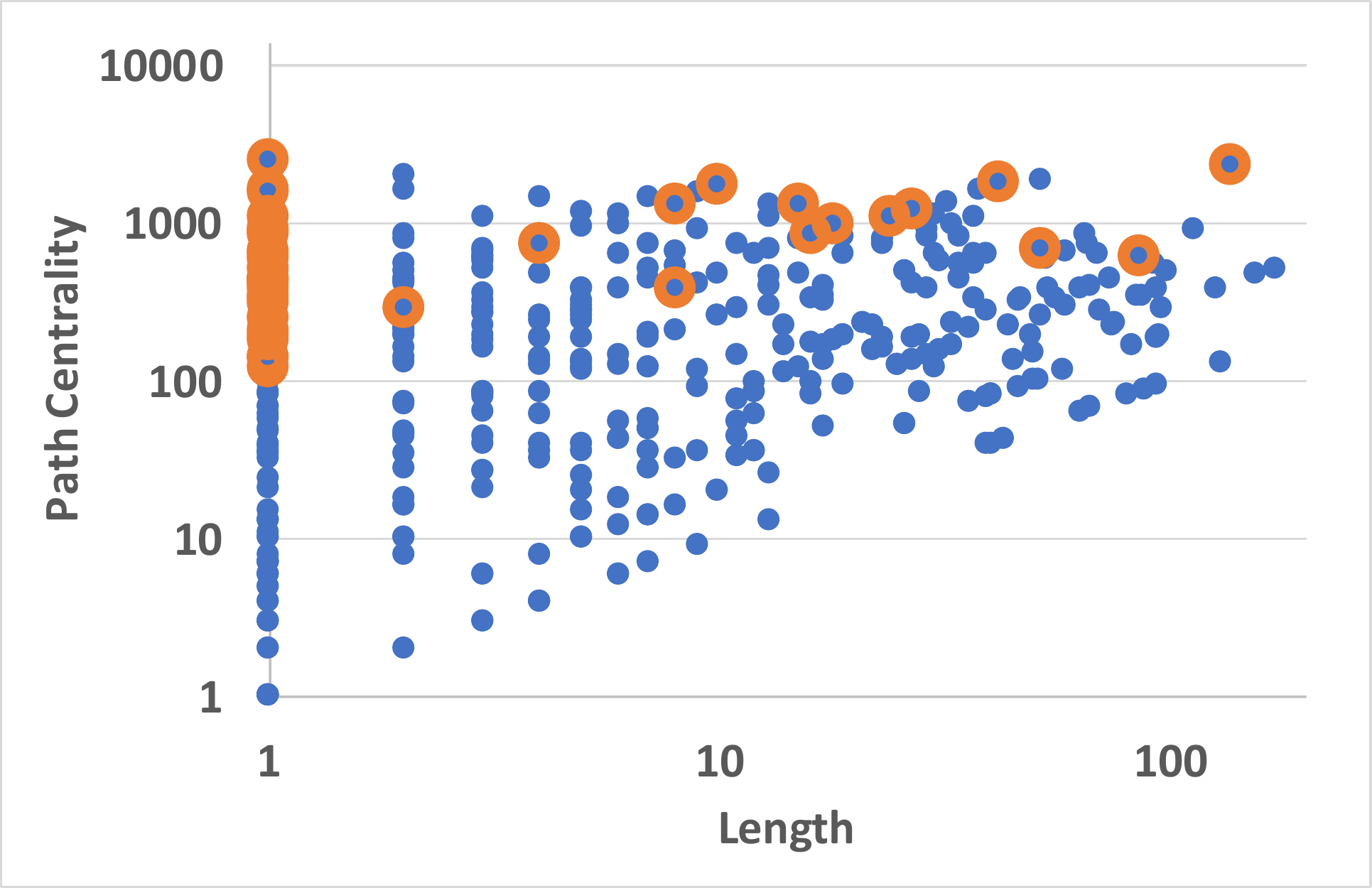}
}
\\
\includegraphics[trim = 0cm 12cm 0cm 12cm, clip, width=0.45\textwidth]{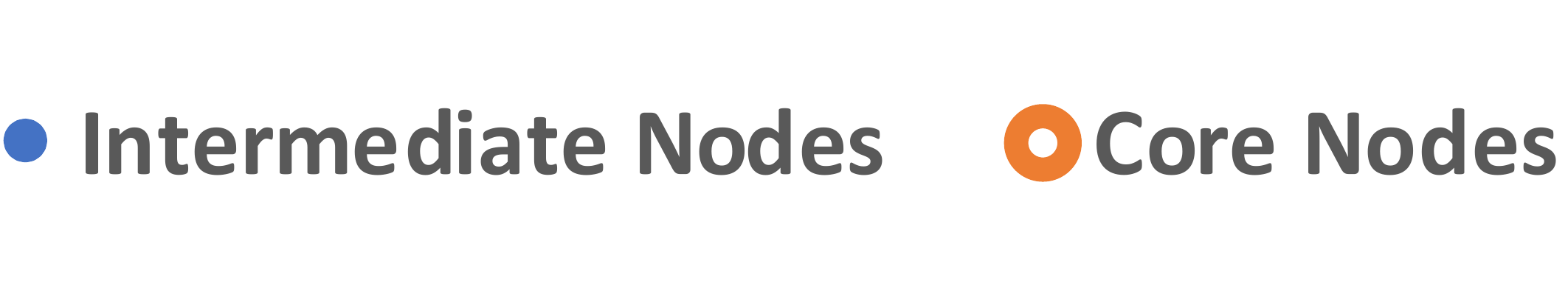}
\caption{Comparison of node length and path-centrality in Lexis-DAGs at the 5,000th iteration (for weak selection model $\beta=1$ and for strong selection model $\beta=12$). For core selection, we set $\tau=0.85$.}
\label{fig:scatters}
\end{figure}

\subsubsection{RND Model}
We also consider a random target generation process, referred to as {\em RND}, where tinkering/mutation are removed from Mutation model. In this model, a new target is randomly generated using $k$ random and independent choices among the sources.

\subsection{Key Metrics}

\subsubsection{Cost Metrics}
\indent{\bf Normalized Cost:} This is the cost of the Lexis-DAG $D_T$ (the Lexis-DAG for the target set $T$) normalized by the total length of the targets, $\mathcal{L}_T$. We denote the normalized cost by $\mathcal{C_N}(D_T)$:
\begin{equation}
0 \leq \mathcal{C_N}(D_T) =\frac{\mathcal{E}(D_T)}{\mathcal{L}_T} \leq 1
\end{equation}

\indent{\bf Penalty of Incremental Design (PID):} This measure evaluates the cost overhead of incremental design relative to a clean-slate design:
\begin{equation}
PID_{T} = \frac{\mathcal{E}(D^{\text{INC}}_T)}{\mathcal{E}(D^{\text{CS}}_T)}
\end{equation}
where $D^{\text{INC}}_T$ is the incremental design for the target set $T$, and $D^{\text{CS}}_T$ is the  clean-slate design for the same set of targets.
The value of PID is bounded as follows:
\begin{equation}
1 \leq PID_T \leq \frac{\mathcal{L}_T}{\mathcal{E}(D^{\text{CS}}_T)}
\end{equation}
because 
 an incremental design cannot be more efficient than a clean-slate design (at least when the two design problems are optimally solved), and 
the maximum cost of incremental
design is $\mathcal{L}_T$).

\subsubsection{Topological Metrics}
\indent{\bf Average Depth:} This metric is an indicator of how deep a Lexis-DAG hierarchy is. For each target $t$, we calculate the average length of all source-target paths ending on that target: $\overline{d}(t)$. The average across all $t$ is defined as the average depth of the hierarchy:
\begin{equation}
\overline{\mathcal{D}}(D_T) = \frac{\sum_{t \in T}\overline{d}(t)}{|T|}
\end{equation}

\indent{\bf Core Stability:}
We have already defined the core size and the H-score (Section \ref{evolexis-background}). Here we define an additional metric, related to the stability
of the core across time.

We track the stability of the core set
by comparing two core sets at two different times. A direct comparison of the core sets via the Jaccard index leads to poor results. The reason is that often 
the strings of the two sets 
are similar to each other but not completely identical.

Thus, we define a generalized version of Jaccard similarity that we call \emph{Levenshtein-Jaccard Similarity}:
\begin{itemize}
\item 
The Levenshtein distance $LD(s,t)$ between two strings $s$ and $t$ is the number of deletions, insertions, or substitutions required to transform one string to another. The higher the number of required operations, the more distant two strings are from each other \cite{clrs}.
\item
Suppose we aim to compute the similarity of two sets A and B of strings. We define the mapping $A\rightarrow B$ where every element $a \in A$ is mapped to the most similar element $b \in B$. We also define the mapping $B \rightarrow A$ from every element $b \in B$ to the most similar element $a \in A$:
\begin{equation}
\begin{cases}
A\rightarrow B = \{(a,b)~s.t.~a \in A~\&~b \in B~\&~b = arg~max_{x \in B} Sim(a,x)\}
\\
B\rightarrow A = \{(b,a)~s.t.~a \in A~\&~b \in B~\&~a = arg~max_{x \in A} Sim(b,x)\}
\end{cases}
\end{equation}
where $Sim(a,b)$ is the similarity of $a$ to $b$ and is calculated as: 
\begin{equation}
Sim(a,b) = 1-\frac{LD(a,b)}{max(|a|,|b|)}
\end{equation}
Notice that $max(|a|,|b|)$ is the maximum value of Levenshtein distance between $a$ and $b$. This ensures that if $a=b$ then $Sim(a,b) = 1$, and if $a$ and $b$ have the maximum distance then $Sim(a,b)=0$.
\item
Considering both $A \rightarrow B$ and $B \rightarrow A$, we get the union of the two mappings and define the Levenshtein-Jaccard similarity as follows:
\begin{equation}
LevJac(A,B) = \frac{\sum_{(a,b) \in A \rightarrow B} Sim(a,b) + \sum_{(b,a) \in B \rightarrow A} Sim(b,a)}{(|A|+|B|)}
\end{equation}
We can see that if $A=B$ (all weights are equal to one) then $LevJac(A,B) = 1$.
Also if none of the elements in $A$ are similar to $B$ (all the element pairs take zero similarity value), then $LevJac(A,B) = 0$.
\end{itemize}

For example, suppose that $A=\{abc, cdef, fgh\}$ and $B = \{abcd, cgef, xyh\}$. The similarity of the most similar pairings is shown next:
\begin{equation}
\begin{cases}
A \rightarrow B = \{(abc,abcd), (cdef,cgef),(fgh,xyh)\}
\\
~~~~~~~~\text{where: } Sim(abc,abcd) = \frac{3}{4}, Sim(cdef,cgef)=\frac{3}{4}, Sim(fgh,xyh)=\frac{1}{3}\\
~~~~~~~~\Rightarrow \sum_{(a,b) \in A \rightarrow B} Sim(a,b) = 1.83\\
B \rightarrow A = \{(abcd,abc), (cgef,cdef),(xyh,fgh)\}
\\
~~~~~~~~\text{where: } Sim(abcd,abc) = \frac{3}{4}, Sim(cgef,cdef)=\frac{3}{4}, Sim(xyh,fgh)=\frac{1}{3}\\
~~~~~~~~\Rightarrow \sum_{(b,a) \in B \rightarrow A} Sim(b,a) = 1.83
\end{cases}
\end{equation}
Hence, we have:
\begin{equation}
LevJac(A,B)=\frac{\sum (A\rightarrow B)+\sum (B\rightarrow A)}{|A|+|B|} = \frac{1.83+1.83}{3+3} = 0.61
\end{equation}

\subsubsection{Target Diversity Metric}
Suppose we have a set of strings $T=\{t_1, t_2, ..., t_n\}$. The goal is to provide a single number that quantifies how dissimilar these elements are to each other.
\begin{itemize}
\item 
We first identify the \emph{medoid} $\mathcal{M}_T$ within the set $T$, i.e., the element that has the lowest average distance from all other elements. We use Levenshtein distance:
\begin{equation}
\mathcal{M}_T=arg~min_{m \in T} \sum_{t \in T}LD(t,m)
\end{equation}
\item
To compute how diverse the elements are with respect to each other, we average  the distance of all elements from the medoid. We call this measure $\sigma_T$, the \emph{Diversity} of set $T$. The bigger the diversity metric, the more diverse the set of strings is (because the distance of each target from the medoid is the number of single-character operations needed to convert any element within the set to the medoid):
\begin{equation}
\sigma_T=\frac{\sum_{t \in T}LD\left[t,\mathcal{M}_T\right]}{|T|}
\end{equation}
\end{itemize}

%% file: 05-target-models.tex
\section{Computational Results}
\label{evolexis-models}

\subsection{Parameter Values and Evolutionary Iteration}
We can summarize an evolutionary iteration of the Evo-Lexis framework as follows:
\begin{enumerate}
\item
Initially, we start with a small number $s$ of randomly constructed targets. Each target has the same length $k$, and the number of possible sources is $n$. An initial Lexis-DAG is constructed using the \textsc{G-Lexis} algorithm.
\item In every evolutionary iteration, the following steps are performed:
\begin{enumerate}
\item
A new batch of $b$ targets is generated via a target generation model.
\item
In the Incremental Design approach, the Evo-Lexis algorithm adjusts the existing hierarchy minimizing the marginal cost of adding each new target in the existing hierarchy. 
\item
If the total number of targets that are present in the system have reached a steady-state (the number of targets is $T_s$), we also remove the oldest batch of $b$ targets from the Lexis-DAG. This target removal process may also trigger the removal of intermediate nodes that are not reused by at least two other nodes in the hierarchy. The total number of targets remains constant ($T_s$) because the number of target additions is equal to the number of removals ($b$). 
\item 
The evolutionary process is repeated for a user-specified number of iterations. The parameters $n$, $k$ and $b$ do not change during this process.
We run each model ten times for a total of 5,000 iterations. 
We take the mean value of each metric.
\end{enumerate}
\end{enumerate}

The parameters used in the following experiments are 
presented in Table \ref{params}.
\begin{table}[h]
\centering
\caption{\label{params} Definition and  parameter values of Evo-Lexis in following experiments}
\begin{tabular}{|c|c|c|}
\hline
{\bf ~~Parameter~~} & {\bf Definition} & {\bf ~~Value~~}
\\\hline
$s$ & Number of initial targets & 10
\\\hline
$n$ & Number of sources & 100
\\\hline
$k$ & Target length (characters) & 200
\\\hline
$b$ & Batch size for new targets birth/old targets death & 10
\\\hline
$T_s$ & ~~Steady-state number of targets present in Lexis-DAG~~ & 100
\\\hline
\end{tabular}
\end{table}

\subsection{Results}

\paragraph{\textit{Emergence of low-cost hierarchies due to tinkering/mutation and selection}}
In Fig. \ref{fig:results-2:cost} and \ref{fig:results-1:cost}, we observe a significant reduction in the normalized cost between the RND model and all other models. The main reason for this reduction is that in all other models, we generate targets that are similar to earlier targets and not randomly constructed. Further, we observe that endogenous models (MS-strong and MRS-strong) further reduce the cost of the resulting hierarchies. The reason is the large bias for selecting targets that can be constructed with lower (or comparable) cost than the seed targets they evolved from. Thus, introducing tinkering/mutation and  selection both contribute to the emergence of more efficient hierarchies in the Evo-Lexis framework.
\paragraph{\textit{Low-cost design resulting in deeper hierarchies and reuse of more complex modules}}
Having a lower cost hierarchy also means that intermediate nodes are reused more frequently and/or that those intermediate nodes are more complex (i.e., longer strings). We  observe this across models in Fig. \ref{fig:results-2:depth}, \ref{fig:results-2:length}, \ref{fig:results-1:depth} and \ref{fig:results-1:length} -- models with lower normalized cost have deeper Lexis-DAGs and higher intermediate node length. These longer re-used nodes further decrease the cost of the hierarchy. Hence, tinkering/mutation and selection also develop deeper hierarchies with longer intermediate nodes. These two outcomes are ubiquitously observed in both natural and technological systems. Examples include call-graphs and metabolic networks. For instance, for the OpenSSH call-graph and the monkey metabolic network, it has been reported that the underlying dependency networks have an average depth of $10.4$ and $8.1$, respectively \cite{sabrin2016}.

\paragraph{\textit{The recombination mechanism creates target diversity}}
Realistic hierarchies should support a diverse set of requirements or outputs. For example, in network protocol stacks, many different functionalities at the top level of the hierarchy (application layer) are supported by the same hierarchical infrastructure. In our framework, this translates to having a set of targets with high diversity. In Fig. \ref{fig:results-2:diversity} and \ref{fig:results-1:diversity}, we show the target diversity across  different models. The RND model produces the highest target diversity as there are no correlations among the generated targets. In Fig. \ref{fig:results-1:diversity}, we observe that the tinkering/mutation in the M model results in 50\% to 70\% decrease in target diversity. Strong selection in the MS-strong model further decreases the diversity to the point that the targets are almost identical, with only minor variations of the same main string. Such low target diversity is not realistic in natural and technological systems. The reason that the MS-strong model behaves in this manner  is that it generates new targets only through single-character mutations and only when the resulting mutants can be constructed using the existing intermediate nodes (otherwise they would have much higher cost and they would not be selected). Hence, the set of accepted new targets gets very narrow and quite similar to its seed targets.

In biological systems, the evolution of complex species required recombination and sexual reproduction (i.e., crossover). Similarly in the Evo-Lexis framework, the addition of recombination in the MRS model results in increased  target diversity (Fig. \ref{fig:results-2:diversity}) while keeping the earlier properties of the Lexis-DAGs  (i.e., low-cost, large depth, long intermediate nodes).

\paragraph{\textit{Reuse of complex modules in the core set by strong selection}}
Looking at the contents of the core at the 5,000th iteration of all models in Fig. \ref{fig:scatters}, shows that in models without selection, or with weak selection, the core includes only a small number of intermediate nodes.  The reason  is that random mutations make the reuse of longer intermediate nodes unlikely. Note that this does not  mean that long intermediate nodes do not exist in Lexis-DAGs under the M \& MS-weak \& MRS-weak models --  such nodes are less likely however to be reused often. As a result, shorter nodes and mostly sources are more likely to appear in the new targets, and end up in the core set.

On the other hand, models with strong selection (MS  and MRS) limit the locations where the seed(s) can be mutated when generating new targets. This constraint results in reusing longer intermediate nodes. Thus, selection creates a bias towards the reuse of longer intermediate nodes. In the long run, this results in some long nodes dominating the core set in the MS-strong and MRS-strong models (Fig. \ref{fig:scatters:mut+sel-strong} \& \ref{fig:scatters:rec+sel-strong}).

\paragraph{\textit{Emergence of hourglass architecture due to the heavy reuse of  complex intermediate modules in models with strong selection}}
Appearance of longer and heavily reused intermediate nodes in the models with strong selection means that the architecture exhibits the hourglass effect. Indeed, we observe in Fig. \ref{fig:results-2:core} \& \ref{fig:results-1:core} that the core size gets significantly smaller in the presence of strong selection (MS and MRS models). Additionally, Fig. \ref{fig:results-2:hscore} \& \ref{fig:results-1:hscore} show that the MS-strong and MRS-strong models also result in higher H-score values (0.4 and 0.65 on average, respectively). Lexis-DAGs with high H-score values have a small core size with respect to the equivalent flat Lexis-DAG whose core is made up of sources and targets only. 

Overall, the reuse of longer intermediate nodes caused by selection results in  hierarchies with an hourglass architecture. This observation is consistent with a mechanism (known as \emph{Reuse-Preference \cite{sabrin2016}}) that was proposed earlier for the emergence of the hourglass effect in general dependency networks.

\paragraph{\textit{Stability of the core set due to selection}}
Selection also promotes the stability of the core set, as shown in Fig. \ref{fig:results-1:stability} for the MS-strong model. We see an increase in core stability (i.e. similarity of the core during evolution) compared to the MS-weak and M models whose cores mostly consist of sources. Similarly, a stable core is also observed in the MRS-weak and MRS-strong models in Fig. \ref{fig:results-2:stability}. We have already seen that long intermediate nodes appear more often in the core set of models with strong selection. Hence the core stability results show that selection not only contributes to the emergence of a small core, consisting of few highly reused intermediate nodes, but it also promotes the conservation of these core nodes during evolution. This is in agreement with the properties of several systems in which the waist of the hourglass architecture includes critical modules of the system that are highly conserved \cite{akhshabi2011,sabrin2016}. We return to this point, where we further show that this core stability is occasionally interrupted by
major transitions and punctuated equilibria. 

\paragraph{\textit{Fragility caused by stronger selection}}
Fig. \ref{fig:results-2:robustness} and \ref{fig:results-1:robustness} show how the generated hierarchies perform in terms of \emph{robustness}, when we remove
the most central nodes in the system, i.e., the members of the core. Robustness generally relates to the ability to maintain a certain function even when there are internal or external perturbations \cite{sabrin2016}. Fig. \ref{fig:results-1:robustness} and \ref{fig:results-2:robustness} show how the removal of one or more core nodes, in order of importance, contributes to cutting source-target paths in each of the Lexis-DAGs produced (at the 5,000th iteration of each model). 

In hourglass architectures (MS-strong and MRS-strong model), core nodes contribute much more significantly to the overall hierarchy by covering many more source-target paths. Hence, such architectures are fragile if the core nodes are perturbed. This is similar to the concept of removal of hub nodes in scale-free network \cite{barabasi-network}. Weakening selection, reduces the H-score (as in Fig. \ref{fig:results-2:hscore}) and hence, reduces the contribution of core nodes in covering source-target paths. 

Fig. \ref{fig:conclusion} summarizes the  properties of the hierarchies that emerge in the models we described in this section.

%% file: 06-evolvability.tex
\begin{figure}[h]
\centering
\includegraphics[width=0.65\textwidth]{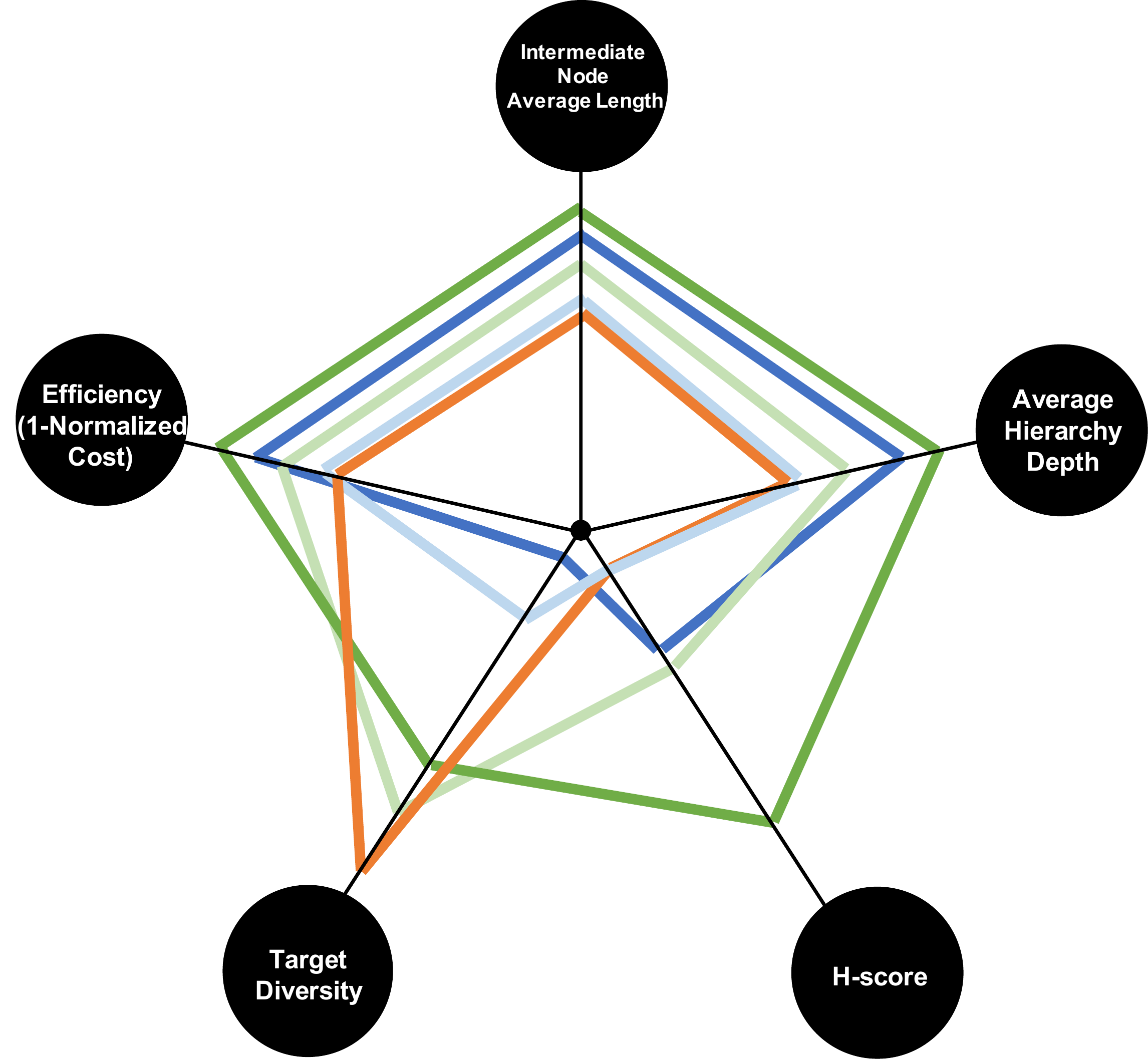}
\includegraphics[trim = 28.5cm 1cm .25cm 4cm, clip, width=0.2\textwidth]{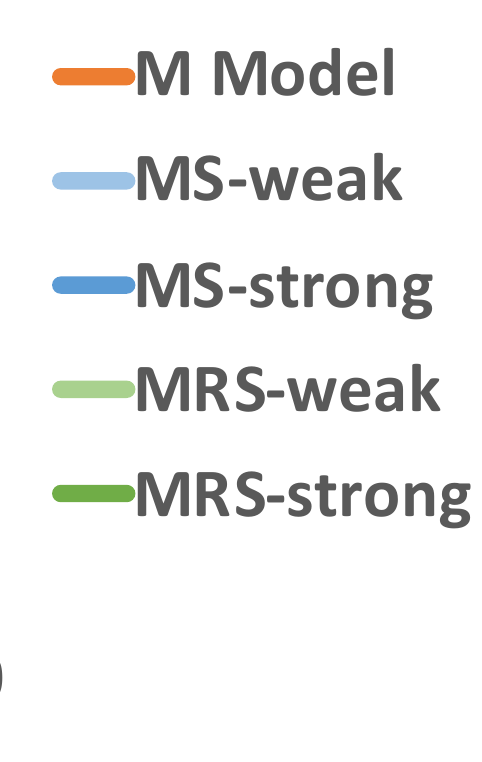}
\caption{Visualizing the various properties of the generated hierarchies that emerge from each model (excluding the RND model). The MRS model produces all properties. This figure shows an approximate value for each metric at the 5,000th iteration of evolution. We define $\text{Efficiency} = 1-\text{Normalized Cost}$. 
\label{fig:conclusion}
}
\end{figure}

\section{Evolvability and the Space of Possible Targets}
\label{evolexis-evolvability}

As shown in the previous section, the MRS-strong model leads to hourglass hierarchies, maintaining at the same time significant target diversity. In this section, we further show that hourglass architectures have two important properties. On the positive side, they are more evolvable in the sense that new targets can be constructed at a low cost, mostly reusing the intermediate modules in the core of the hierarchy. On the negative side however, hourglass architectures only accept a small fraction of the candidate new targets, restricting what a biologist would refer to as the ``phenotypic space'' of the system. This interplay between evolvability and the space of feasible system phenotypes or functions is an important issue in both biological and technological systems (e.g. Internet architecture \cite{rexford}).

We first look at the cost of targets  produced with and without selection. For this purpose, we compare two models: one is the MRS-strong model that acts as an ``endogenous'' target generation process. The other is a variation of MRS without selection that we call {\em MR model} (only mutations and recombination) -- this is an ``exogenous'' target generation process that does not depend on the current state of the hierarchy. The MR model allows us to examine how selection affects the cost and space of acceptable targets with and without the selection constraint. 

In Fig. \ref{fig:pheno-cost}, we calculate the ratio between the average cost of accepted targets per batch in the MRS-strong model over the corresponding cost in the MR model -- we refer to this as \emph{MRS-over-MR per-batch cost-ratio}.
The average and median values of this ratio are 0.53 and 0.52, respectively.
This suggests that the targets generated under stronger selection are of much lower cost (around half) compared to the targets generated without selection. So, the presence of strong selection allows the system to construct new targets at a much lower cost because those selected targets can be constructed mostly reusing the intermediate nodes present in the hierarchy. 

\begin{figure}[h]
\centering
\subfloat[CDF of Per-Batch Cost-Ratio Values of MRS Model\label{fig:pheno-cost}]{
  \includegraphics[trim = 1.8cm 6cm 1.9cm 7.5cm, clip, width=.4\textwidth]{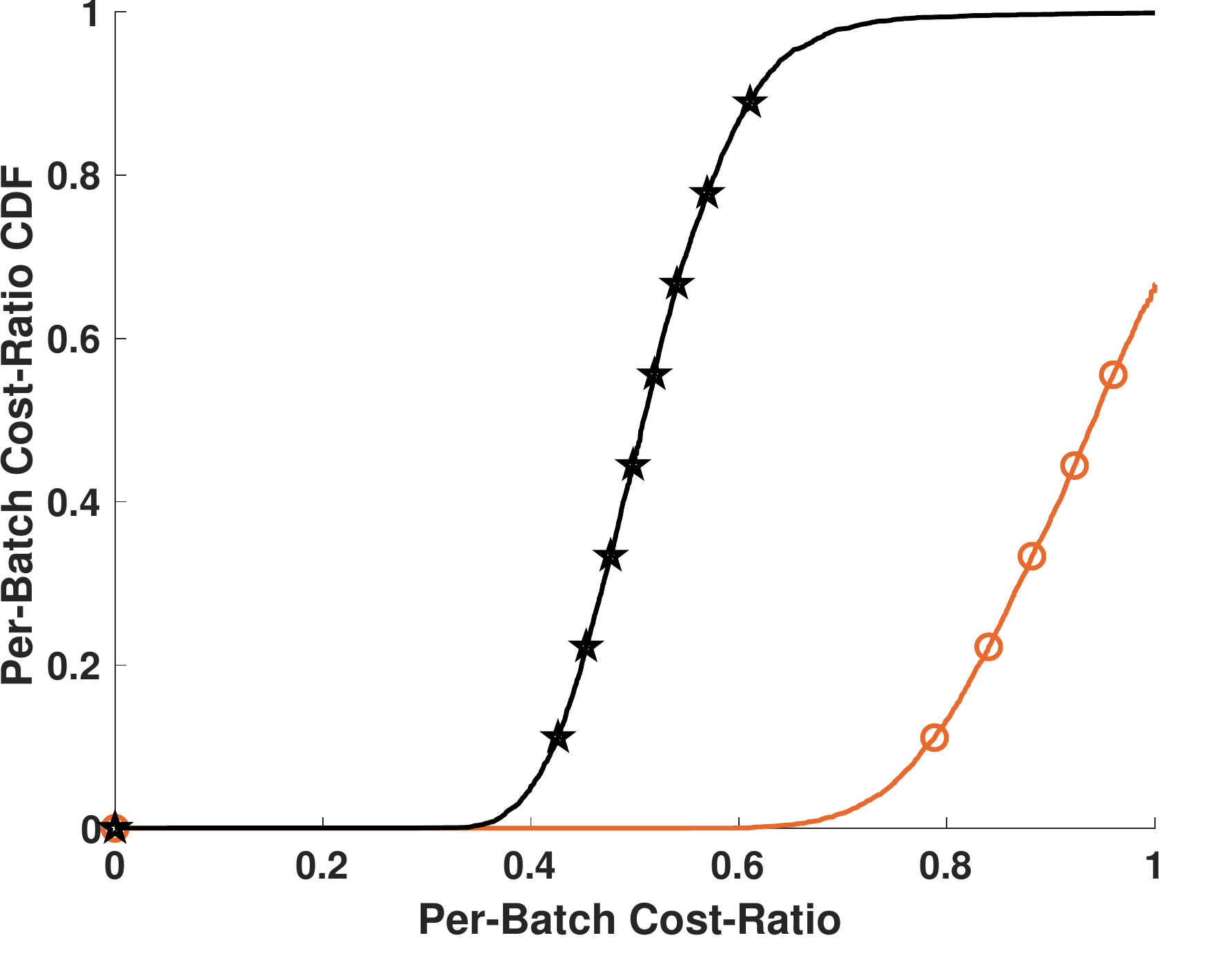}
}
\hspace{1cm}
\subfloat[CDF of Target Acceptance-Likelihoood Values in MRS Model\label{fig:pheno-prob}]{
  \includegraphics[trim = 1.8cm 6cm 1.9cm 7.5cm, clip, width=.4\textwidth]{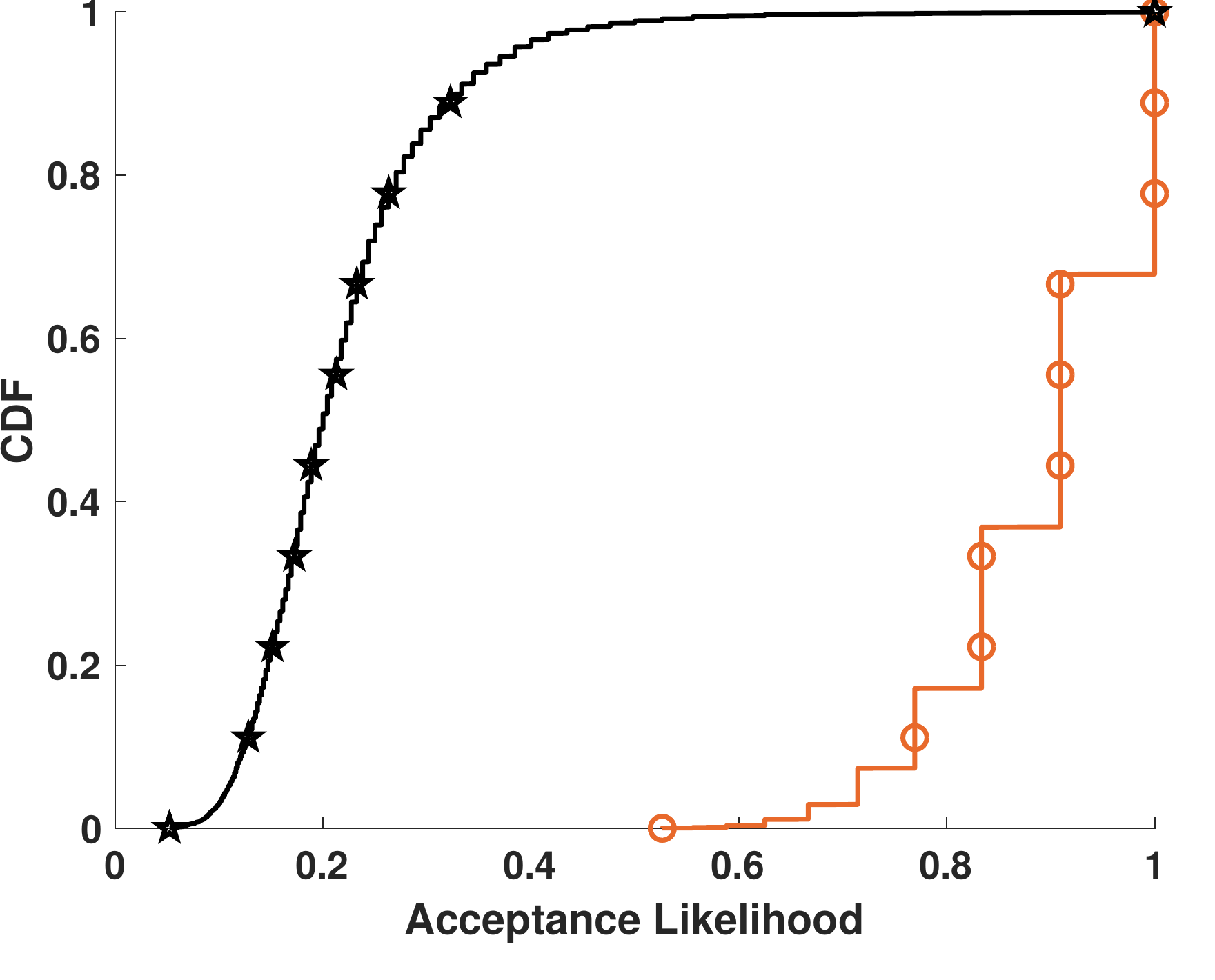}
}
\\
\includegraphics[trim = 7.2cm 14.5cm 3cm 12.8cm, clip, width=.5\textwidth]{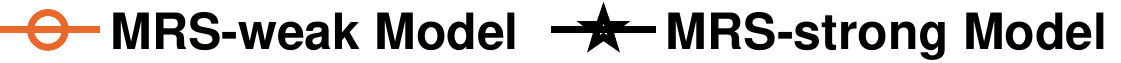}
\caption{{\bf (a)} CDF of MRS-over-MR per-batch cost-ratio, the ratio between the average cost of targets per batch in the MRS model (weak or strong selection) over the average cost of targets per batch in the MR model. {\bf (b)} CDF of the target acceptance-likelihood, i.e., the number of accepted targets generated per-batch in the MRS model divided by the total number of generated targets per batch with the same model. 
}
\label{fig:pheno-cost}
\end{figure}

As a result of strong selection, the \emph{acceptance-likelihood} of new targets generated by the MRS-strong model is much lower than that with the MR model. Specifically, the acceptance-likelihood in Fig. \ref{fig:pheno-prob} is defined as the fraction of accepted targets generated per-batch. The mean and median of this likelihood in the MRS-strong model are equal to 0.2.
In other words, about 80\% of the new targets generated through mutations and recombination are not selected because their cost, given the existing architecture, would be prohibitively high. 

It should be also noted that the MRS-weak model behaves quite similar to the MR baseline in terms of both the MRS-over-MR cost ratio and the target acceptance likelihood.

Overall, the results in this section show that despite having the benefit of lower cost new targets, and thus higher evolvability, selection restricts significantly the phenotypic space of accepted new targets.
Given that the MRS-strong model generates hourglass architectures, we can summarize as follows: hourglass-like hierarchies under the MRS-strong model allow the construction of new functions (accepted targets) at a low cost, by mostly reusing  core modules, but at the same time such architectures significantly restrict which of these functions can be supported. Targets that are quite different than the intermediate modules of the existing hierarchy would most likely not be selected.

%% file: 07-transitions.tex
\section{Major Transitions}
\label{evolexis-transitions}

Major transitions have been an important and interesting phenomenon in both natural and technological evolution. Such transitions create significant shifts in evolutionary trajectories, ecosystems and ``keystone species'' \cite{major-keystone}. There are  many examples of such events in natural systems,  such as the 
``invention'' of sexual reproduction and evolution of multicellularity \cite{major-bio}. In technological evolution, innovations occasionally lead to the emergence of disruptive new technologies, such as the steam engine in the 19th century or air transportation in the 20th century. In the context of computing, the evolution of programming languages has gone through punctuated equilibria, interrupted by  new languages that were developed by tinkering or combining different structural components of older languages  \cite{valverde2015}.  

\begin{figure}[h]
\centering
\begin{flushleft}
\subfloat[MRS-strong Model\label{fig:major-strong}]{
  \includegraphics[trim = 0cm 0cm 0cm 0cm, clip, width=0.75\textwidth]{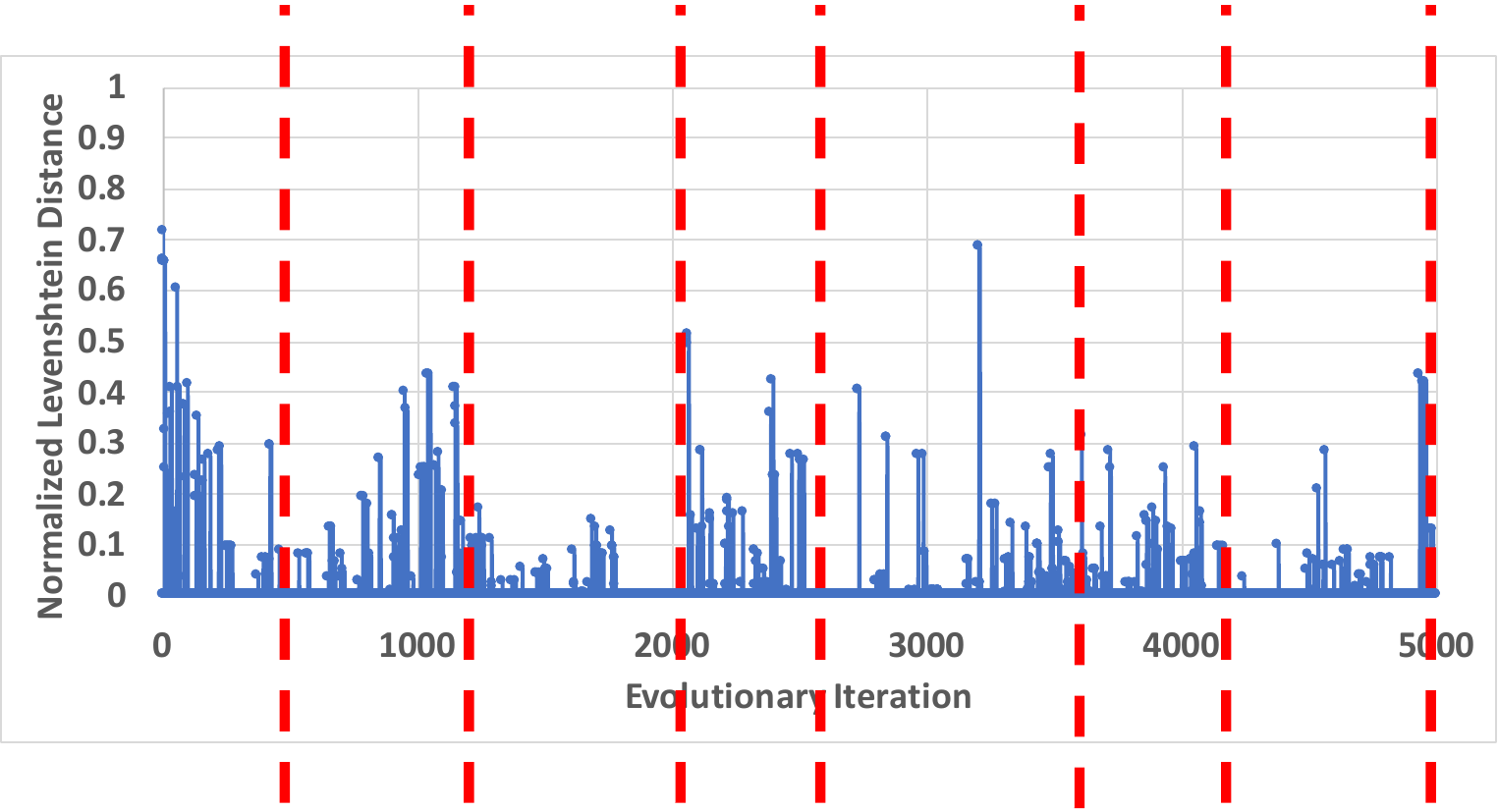}
}
\end{flushleft}
\begin{flushright}
\subfloat[MRS-weak Model\label{fig:major-weak}]{
  \includegraphics[trim = 0cm 0cm 0cm 0cm, clip, width=0.75\textwidth]{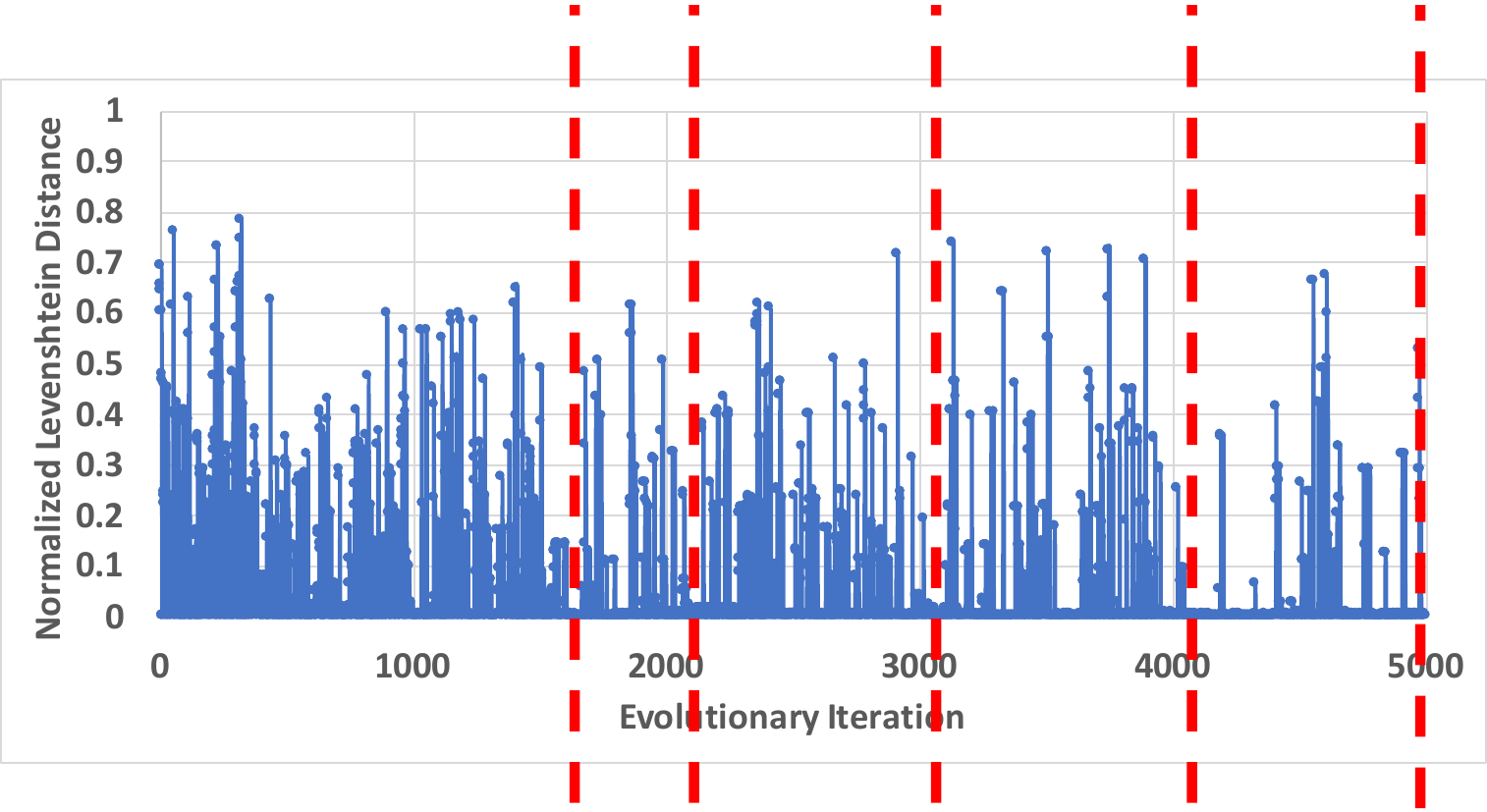}
}
\end{flushright}
\caption{Variability across successive iterations of the top-1 core node (measured using the Levenstein distance) in the MRS model (both strong and weak selection).  The highlighted iterations illustrate some of the stasis periods, in which the top-1 core node remains identical for many iterations. 
}
\label{fig:major-1}
\end{figure}
\begin{figure}[h]
\centering
\begin{flushleft}
\subfloat[${\mu_{LD}} = 0.1$\label{fig:major-strong-stasis}]{
  \includegraphics[trim = 0cm 1.5cm 0cm 0cm, clip, width=0.75\textwidth]{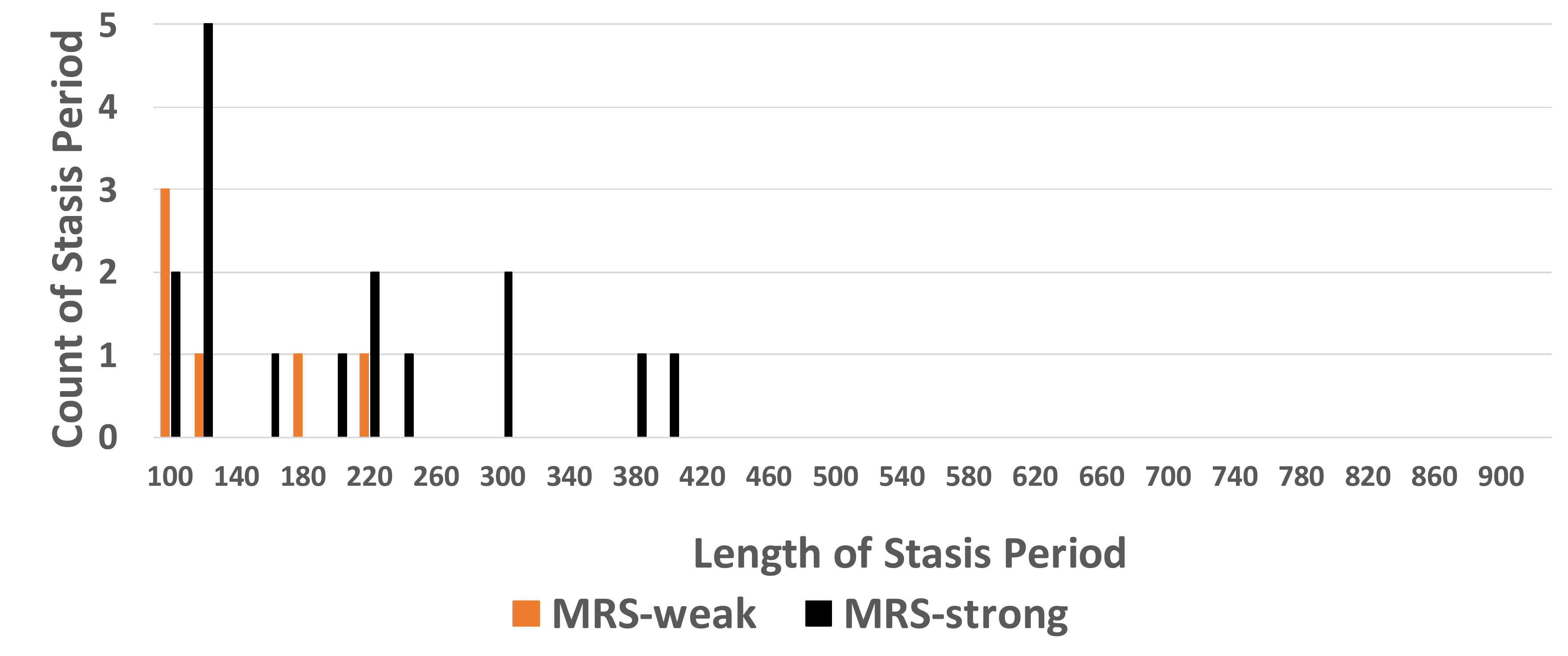}
}
\end{flushleft}
\begin{flushright}
\subfloat[${\mu_{LD}} = 0.2$\label{fig:major-weak-stasis}]{
  \includegraphics[trim = 0cm 1.5cm 0cm 0cm, clip, width=0.77\textwidth]{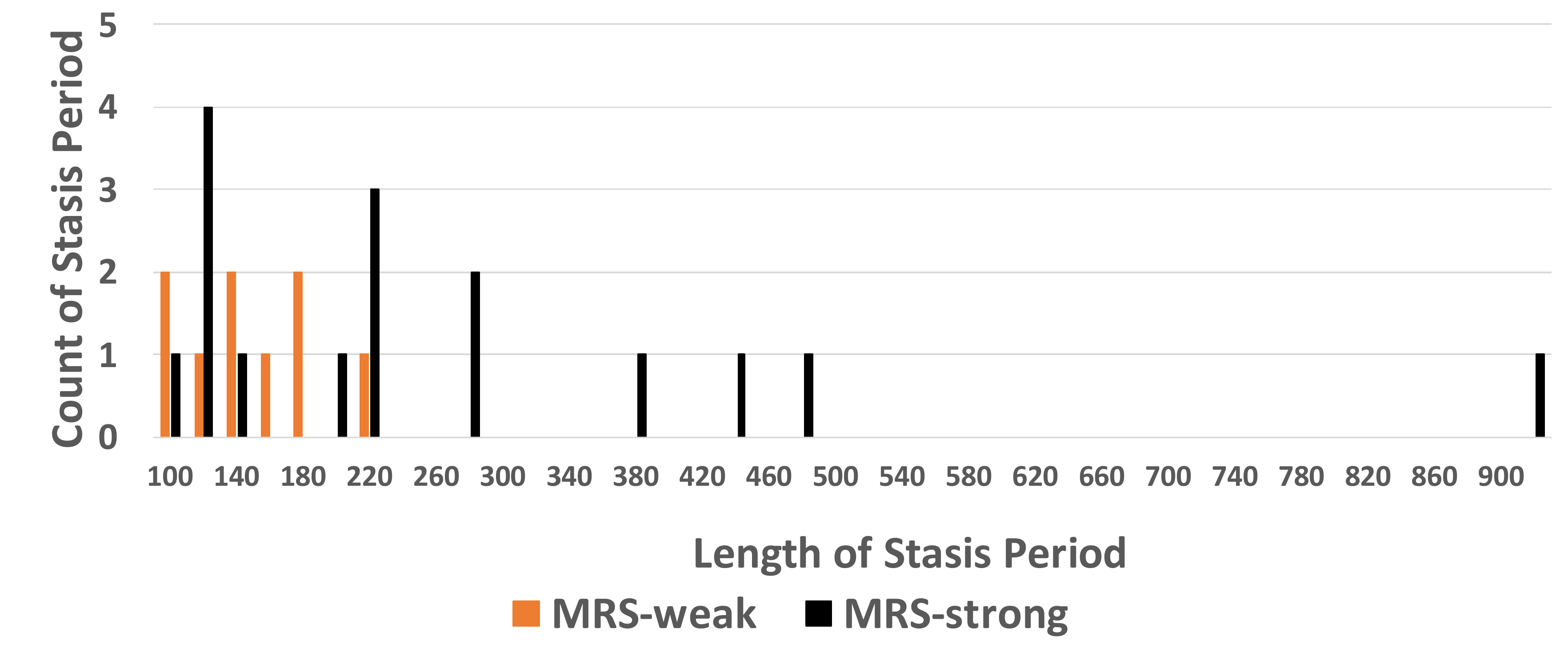}
}
\end{flushright}
  \includegraphics[trim = 0cm 0cm 0cm 13.5cm, clip, width=0.77\textwidth]{images/major-ld2.pdf}
\caption{Count of stasis periods (lasting at least 100 iterations) for two values of the Levenshtein distance threshold, ${\mu_{LD}}$, in Fig. \ref{fig:major-1}. Strong selection leads to longer and more frequent stasis periods. 
}
\label{fig:major-2}
\end{figure}
\begin{figure}[h]
\centering
\subfloat[Top-1 Core Node Changes in MRS-strong\label{fig:major-strong-coretrack}]{
  \includegraphics[trim = 0cm 0cm 0cm 0cm, clip, width=0.49\textwidth]{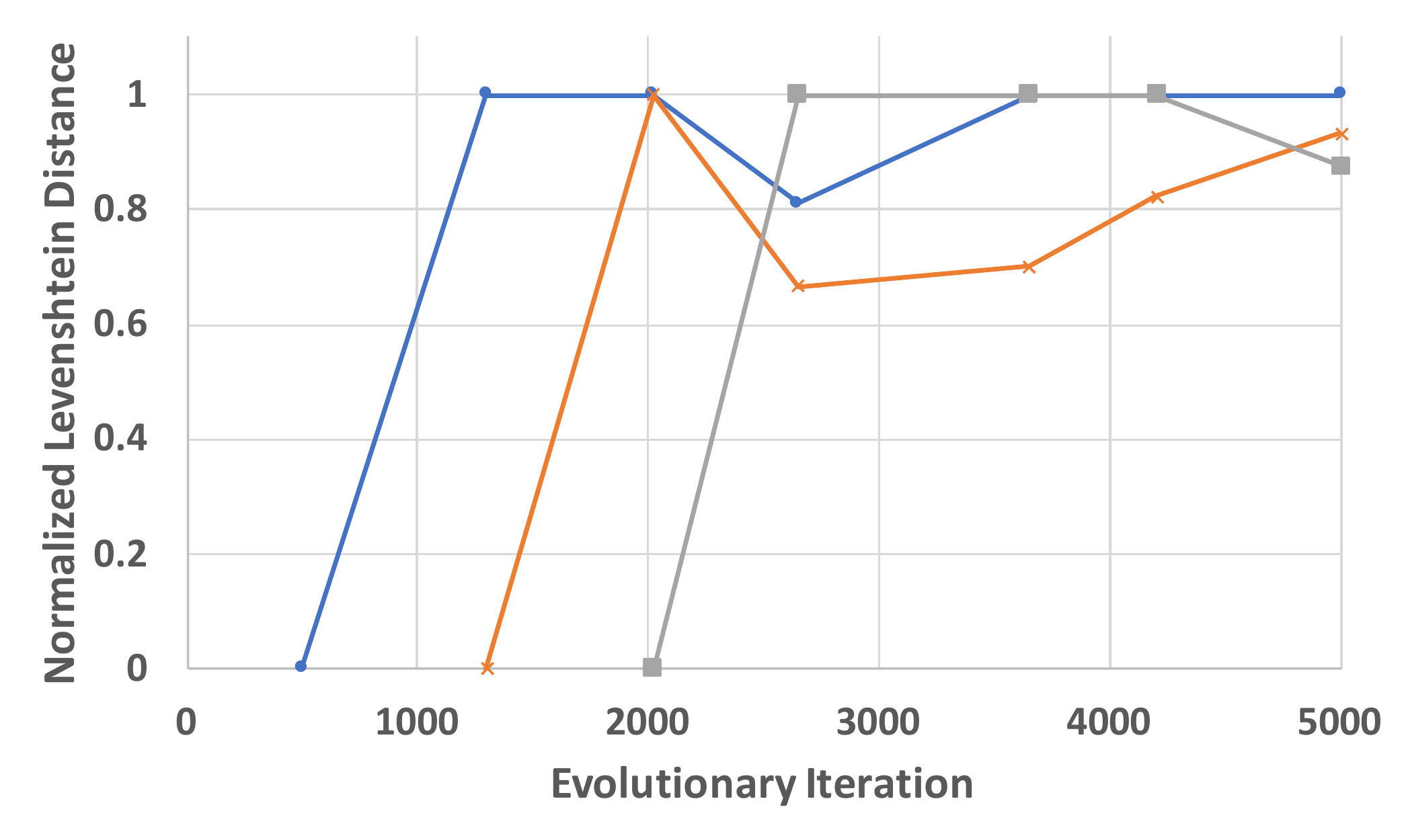}
}
~~
\subfloat[Top-2 Core Node Changes in MRS-strong\label{fig:major-weak-coretrack}]{
  \includegraphics[trim = 0cm 0cm 0cm 0cm, clip, width=0.49\textwidth]{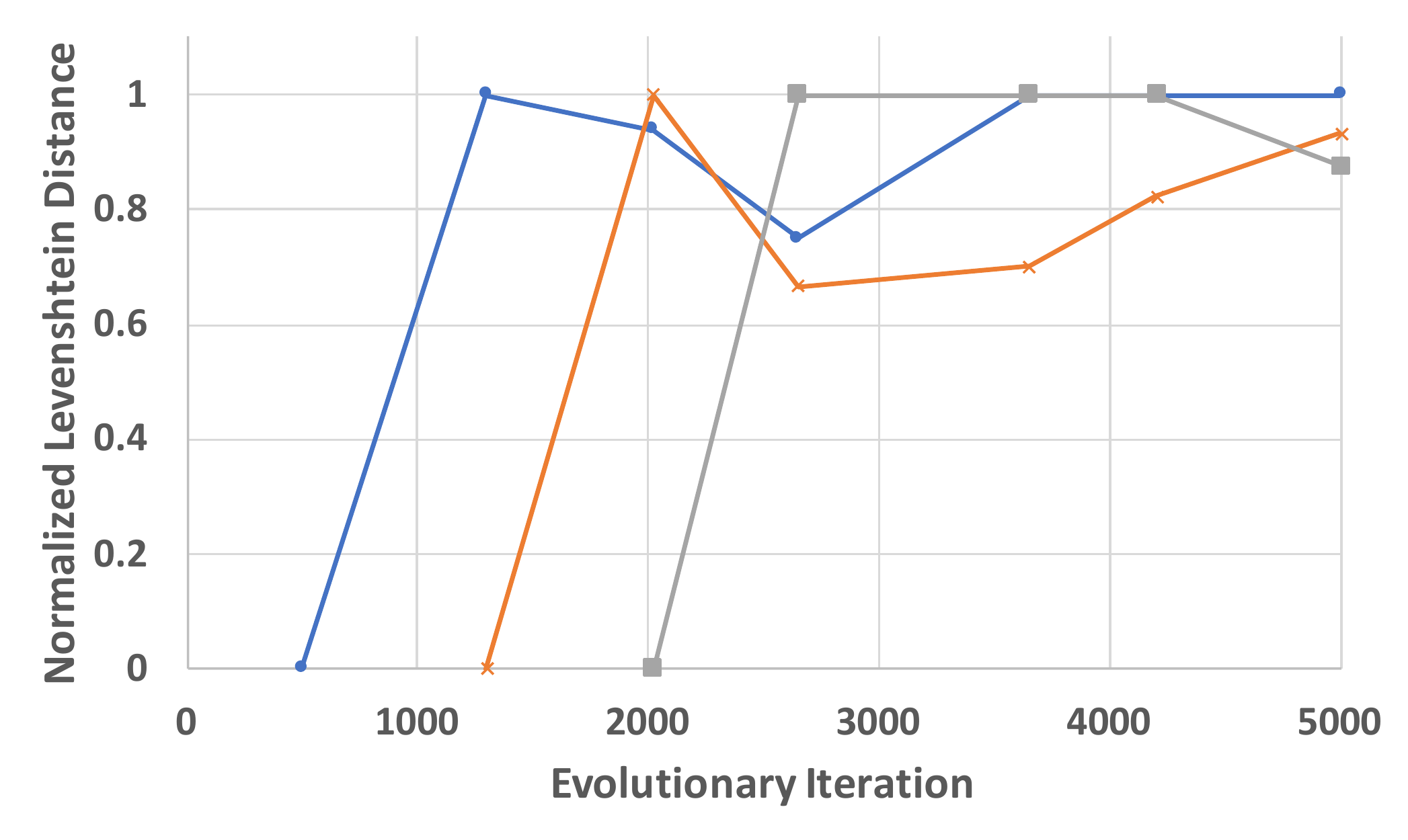}
}
\\
\caption{
Starting from three different stasis periods (with ${\mu_{LD}} = 0.1$), the top-1 and top-2 core node does not stay the same in subsequent stasis periods. The normalized Levenstein distance between the top-1 and top-2 node at the start of each curve and at successive stasis periods is close to 1, suggesting that these nodes have changed. We observed similar results for other core nodes.
}
\label{fig:major-3}
\end{figure}

The results of Fig. \ref{fig:results-2:stability} suggest that the structure of the core is locally stable, when comparing core nodes in adjacent iterations. 
To further investigate the stability of the core during evolution, we focus on the most central node in the core of the Lexis-DAGs, i.e., the core node that covers the largest fraction of source-target paths. We refer to this node of the Lexis-DAG as {\em top-1 core node}.

First, we track the variability of this node locally, by comparing its normalized Levenshtein distance to the top-1 core node in the next iteration.
Fig. \ref{fig:major-1} shows the results of this analysis for both MRS-strong and
MRS-weak. 
In the MRS-strong model, we observe that in most iterations the top-1 core node does not change significantly. Even though there are some spikes in which the Levenshtein distance is larger than 0.2, in 82.6\% 
of the evolutionary iterations the variability of the top-1 core node is less than that. 
Further, there are several {\em stasis periods} in which 
the top-1 core node is practically the same (Levenstein distance lower than 0.1
or even 0). 
In Fig. \ref{fig:major-1} we highlight with red vertical lines a small number of 
stasis periods in which the top-1 core node remains exactly the same for tens of hundreds of iterations. 
On the other hand, the MRS-weak model has significantly higher variability
in the top-1 core node, and fewer/shorter stasis periods. This suggests that
selection is the key factor in generating these long periods of stability
in the core of the hourglass architecture. 

To further quantify this point, we focus on stasis periods that last 
at least 100 iterations (recall that the entire evolutionary paths
in these results consist of 5000 iterations). 
Fig. \ref{fig:major-2} shows that there are fewer and shorter stasis periods in MRS-weak model than in MRS-strong.  
The fraction of iterations that account for stasis conditions is $\frac{478}{5000} \sim 0.095$ in MRS-weak, and $\frac{2928}{5000}\sim 0.585$ in MRS-strong, when the minimum Levenshtein distance is ${\mu_{LD}}=0.1$ (also $\frac{1049}{5000} \sim 0.209$ in MRS-weak and $\frac{4133}{5000}\sim 0.826$ in MRS-strong when ${\mu_{LD}}=0.2$). 

The presence of stasis periods under strong selection suggests that 
the most central intermediate nodes at the waist (or core)
of the hourglass architecture can be quite stable and time-invariant.
What happens however across different stasis periods?
Does that stability persist across different stasis periods, 
or does the architecture exhibit major transitions and punctuated
equilibria?

To answer this question, we focus again on the top-1 core node and 
measure its variability across successive stasis periods. 
In Fig. \ref{fig:major-3}, we consider three
different stasis periods (one curve for each initial stasis period), 
and calculate the normalized Levenstein distance between the top-1 core
node in its initial stasis period and the top-1 core node in subsequent stasis  periods.
Note that the top-1 core node changes significantly across stasis periods.
In fact, the Levenshtein distance is so high (often close to 1), suggesting
that these are completely different core nodes. 
This observation gives more evidence that the top contributors to the core can lose their importance during evolutionary time scales, causing major transitions
in both the core set and, consequently, in the overall hierarchy. 
We have confirmed that this is even more common for lower centrality core nodes too, and it is certainly even more true under weak selection.  

%% file: 08-overhead.tex
\section{Overhead of Incremental Design}
\label{evolexis-overhead}
\begin{figure}[h]
\center
\subfloat[Penalty of Incremental Design \label{fig:results-pid:pid}]{
  \includegraphics[width=.5\textwidth]{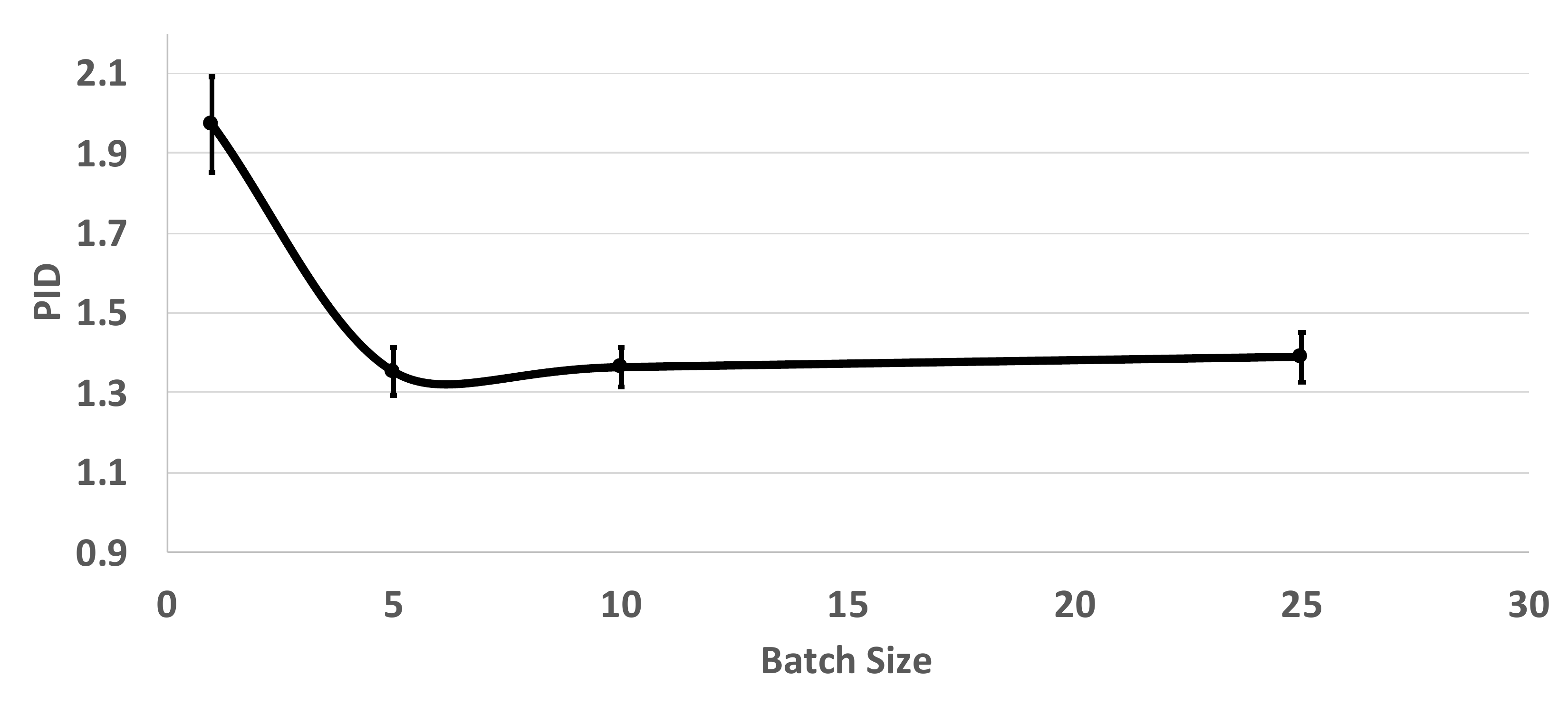}
}
\subfloat[Core Similarity\label{fig:results-pid:sim}]{
  \includegraphics[width=.5\textwidth]{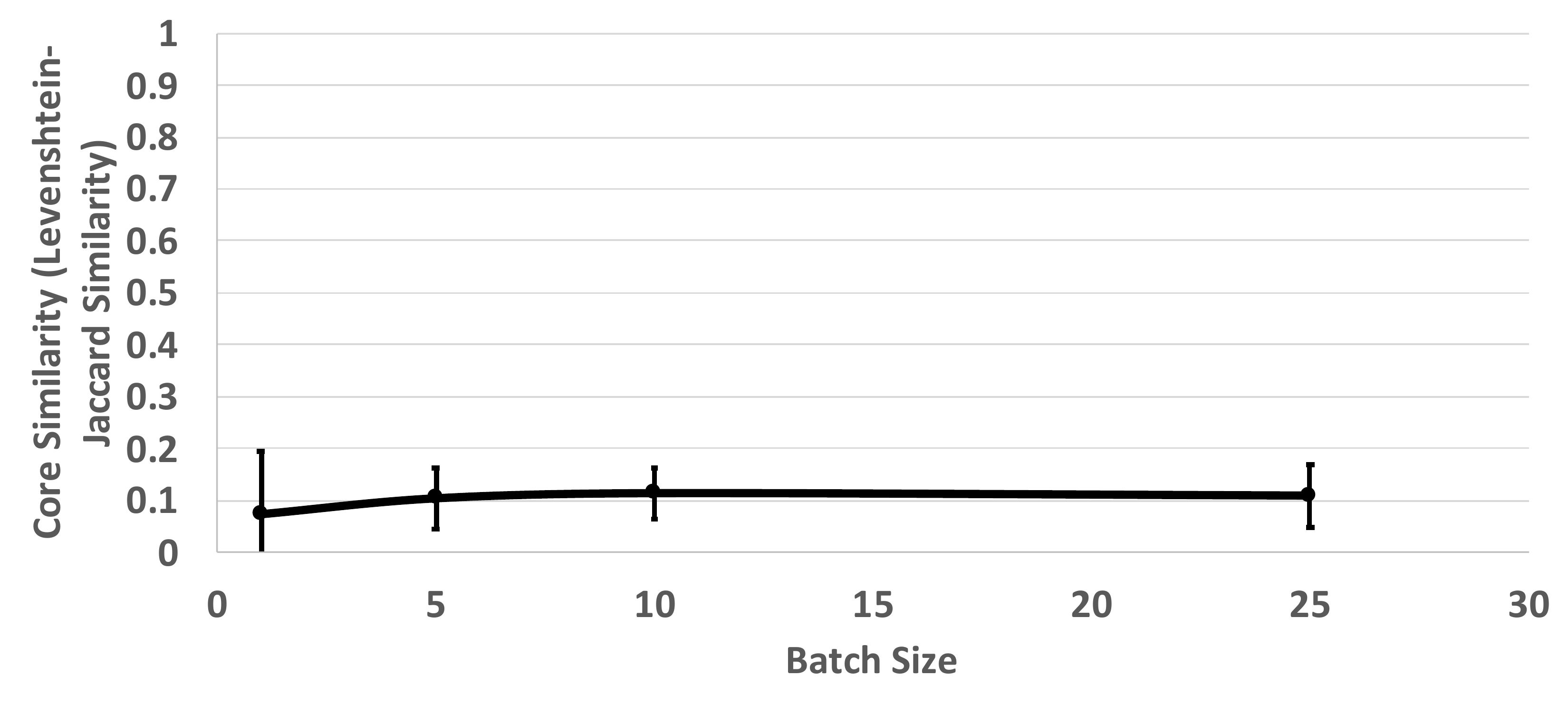}
}
\\
\subfloat[Average Hierarchy  Depth\label{fig:results-pid:depth}]{
  \includegraphics[width=.5\textwidth]{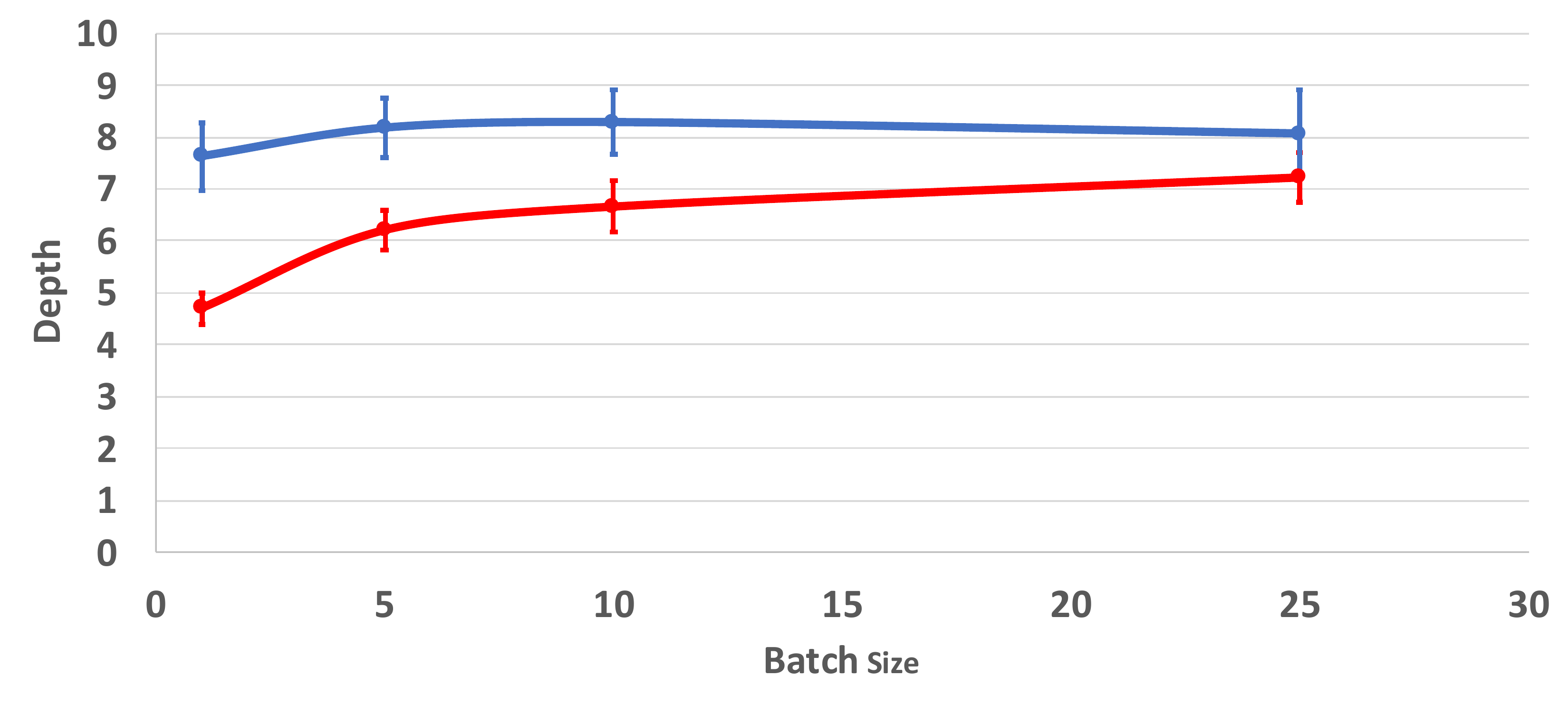}
}
\subfloat[H-Score\label{fig:results-pid:hscore}]{
  \includegraphics[width=.5\textwidth]{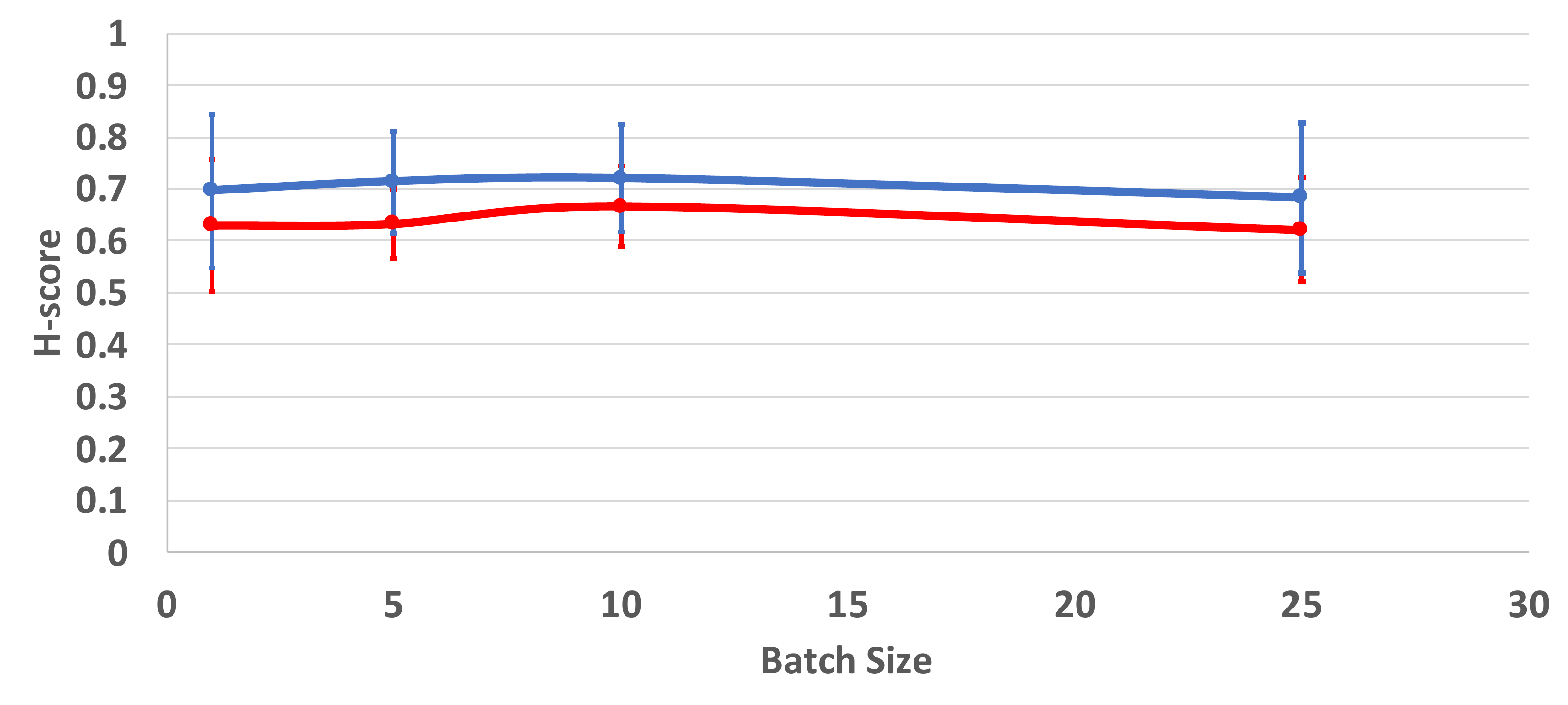}
}
\\
\includegraphics[trim = 5cm .5cm 3cm 13cm, clip,scale=0.3]{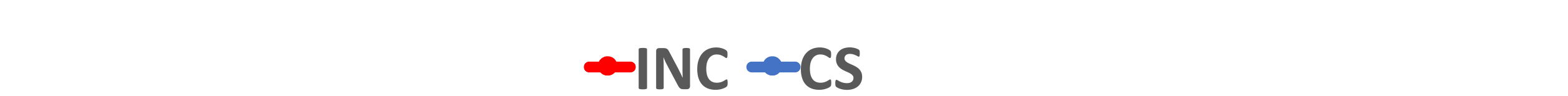}
\caption{Comparison between Incremental (INC) design and Clean-Slate (CS) design, in terms of four metrics and for different batch sizes. For each batch size, the MRS-strong model is run for 5,000 iterations and an average of each metric is taken over 50 distinct iterations. The considered batch sizes are: 1, 5, 10, 25.}
\label{fig:results-pid}
\end{figure}

In this section, we compare the cost and structural characteristics of \emph{Incremental design} (INC) relative to \emph{Clean-Slate} (CS) design, i.e.,
the ideal case in which a new Lexis hierarchy is designed from scratch
every time the set of targets is changed. 
Of course such clean-slate designs are rare or infeasible in practice, 
especially in biological evolution.
CS design is still valuable however as a baseline for evaluating the
cost efficiency of INC, and the hierarchy that is produced by the
latter.

In the Evo-Lexis framework, a key factor that quantifies the difference
between INC and CS design is the \emph{batch size}. 
If the batch size $b$ is equal to the total number of targets in steady state $T_s$, INC and CS are equivalent because the set of targets completely changes 
in each iteration. 
At the other extreme, if the batch size is only one target and $T_s \gg 1$, 
INC performs a minimal adjustment of the hierarchy to support the new target
while CS still redesigns the complete hierarchy. 
In other words, the fraction $b/T_s$ controls the degree of change 
in each evolutionary iteration.
Both in natural and technological systems, evolution proceeds rather 
slowly -- for this reason we only consider the lower range of this ratio, between
1/100 and 25/100. 

In the following we only consider the MRS-strong model (based on the results
of the earlier sections). 
Fig. \ref{fig:results-pid} compares INC and CS in terms of four key metrics. 
The first metric relates to cost: recall that the Penalty of Incremental
Design (PID) is the ratio of the cost of an evolving INC hierarchy over 
the cost of the corresponding CS hierarchy for the same set of targets.
With the exception of the minimum possible batch size ($b$=1), it is 
interesting that INC does {\em not} lead to much less efficient
hierarchies than CS. The PID metric shows that INC is typically around 30\%
more costly than CS for a wide range of batch sizes, suggesting that INC
is  able to often reuse intermediate nodes in constructing the given 
targets, despite the fact that it cannot redesign the complete hierarchy.
The PID is substantially higher when $b$=1 however. The reason is that when the 
INC-Lexis algorithm is given only one new target in every iteration, it is unlikely 
to identify segments of that single target that repeat more than once.
This means that, when $b$=1, INC rarely adds new intermediate nodes in the hierarchy
even though successive targets can be quite similar.
CS, on the other hand, exploits the similarity of the set of targets
in each iteration constructing more intermediate nodes, and reducing 
cost through their reuse. 

Interestingly, even when the INC and CS designs have similar costs, 
they are very different in terms of the nodes that form the core. 
This is shown in Fig. \ref{fig:results-pid:sim}: the similarity of 
the two cores according to the Levenshtein-Jaccard similarity is 
around 0.1. This implies that the two design approaches lead to 
substantially different architectures in terms of the actual intermediate
nodes they reuse.

Additionally, the average hierarchical depth of CS architectures is larger 
(see Fig. \ref{fig:results-pid:depth}) 
because this design approach is able to identify more and longer
intermediate nodes that can be reused to construct the entire set of targets. 
INC, on the other hand, is constrained to not adjust the existing 
portion of the hierarchy, and it can only form new intermediate
nodes when it detects fragments in the set of new targets that are
repeated more than once. So, the INC hierarchies are typically not 
as deep as those in CS. 

Despite their differences, both design approaches lead to 
hourglass architectures when the targets are created with the MRS-strong model.
This is shown in Fig. \ref{fig:results-pid:hscore}, 
and it suggests that even though INC is constrained, as described above, 
it is still able to identify few intermediate nodes that can 
be reused many times to construct the time-varying set of targets.

%% file: 02-related-work.tex
\section{Discussion and Prior Work}
\label{evolexis-related}

The Evo-Lexis model is primarily related to three research themes: 
first, the emergence of modularity and hierarchy in complex systems; second, 
the hourglass architecture in hierarchical networks; 
and lastly, the comparison between offline (or ``clean slate'') design and online (or incremental) design.

\subsection{Modularity and Hierarchy}
The modeling framework of ``Modularly Varying Goals'', by Kashtan and Alon, is a plausible explanation for the 
emergence of modularity \cite{kashtan2005, kashtan2007}. 
By applying incremental changes in logic circuits and evolving neural networks for pattern recognition tasks, they show that modularity in the goals (what we refer to as ``targets'') leads to the emergence of modularity in the organization of the system, whereas randomly varying goals do not lead to modular architectures.
Similarly, 
Arthur et al. focus on the evolution of technology using a simple model of logic circuit gates \cite{arthur2006}. 
Each designed element is a combination of simpler existing elements. 
Their simulation model results in a modularly organized system, in which complex functions are only possible by first creating simpler ones as building blocks. 
These models are similar to Evo-Lexis in the following way:
when the system targets are not randomly constructed but they are generated
through an evolutionary process that involves mutations, recombination and selection, the target functions 
are computed through deep hierarchies that reuse common intermediate components.

Clune et al. show that modularity is a key driver for the evolvability of complex systems \cite{clune2012}. 
The authors demonstrate that selection mechanisms that minimize the cost of connections between nodes in a networked system result in a modular architecture. This is shown by evolving networks that solve pattern recognition tasks and Boolean logic tasks. The inputs sense the environment (e.g. pixels) and produce outputs in a feed-forward manner (e.g. the existence of patterns of interest). In other words, the networks that have evolved for optimizing both performance (accuracy in recognition) and cost (network connections) are more modular and evolvable (in the sense of being adaptable to new tasks) than those optimized for performance only. 
In a follow-up study by Mengistu et al. in \cite{clune2016}, it is shown that the minimization of the cost of connections also promotes the evolution of hierarchy, the recursive composition of sub-modules. When not modeling the cost of connections, even for tasks with hierarchical structure (e.g. a nested boolean function), a hierarchical structure does not emerge.
These modeling frameworks are similar to Evo-Lexis because the latter also aims to minimize the number
of connections in the resulting hierarchical network, and it is this cost minimization that 
provides the incentive for reuse of intermediate components.

At the empirical side, prior work has established that technology evolves similarly to biological evolution,
through tinkering, new combinations of existing components, and selection.  For instance, 
a study of USPTO data gives evidence for the combinatorial evolution of technology \cite{youn2015}. The authors find that the rate of new technological capabilities is slowing down but a huge number of combinations allows for a ``practically infinite space of technological configurations''. By considering technology as a combinatorial process, \cite{kim2016} uses USPTO data to investigate the extent of novelty in patents. They propose a likelihood model for assessing the novelty of combinations of patent codes. Their results show that patents are becoming more conventional (rather than novel) with occasional novel combinations.

\subsection{Hourglass Architecture} 
A property of many hierarchical networks is the \emph{hourglass effect}, which means that the system receives many inputs and produces many outputs through a relatively small number of intermediate modules that are critical for the operation of the entire system \cite{sabrin2016}. This property is also one of the main themes investigated in our work.

Akhshabi et al. studied the \emph{developmental hourglass} which is the pattern of increasing morphological divergence towards earlier and later embryonic development \cite{akhshabi2014}. The authors conclude that 
the main factor that drives the emergence of the hourglass architecture in that context is that the developmental gene regulatory networks become increasingly more specific, and thus sparser, as developmemt progresses.
Earlier, the same authors in \cite{akhshabi2011} were inspired by the hourglass-resemblence of the Internet protocol stack in which the lower and higher layers tend to see frequent innovations, while the protocols at the waist of the hourglass appear to be ``ossified''. The authors present an abstract model, called \emph{EvoArch}, to explain the survival of popular protocols at the waist of the protocol stack.  
The protocols which provide the same functionality in each layer compete with each other and, just as in \cite{akhshabi2014}, the increasing specificity and sparsity is what causes the network to have an hourglass architecture.
The Evo-Lexis model is neither layered, nor probabilistic, 
and so it is fundamentally different than {\em EvoArch}, but it also generates hierarchies in which the 
nodes that represent shorter strings (equivalent to lower-layer nodes in EvoArch) are reused more frequently
and so they have a higher out-degree.  

Friedlander et al. focus on layered networks that perform a linear input-output transformation \cite{alon2015} and
show that in such systems the hourglass architecture emerges when that transformation is compressible. 
In their model, this is interpreted as rank-deficiency of the input-output matrix that describes the function of the system.
A further requirement is that there should be a goal to reduce the number of connections in the network,
similar to Evo-Lexis.
This rank-deficiency in the input-output matrix resembles the case in which Evo-Lexis targets are not constructed
independently but through an evolutionary process that generates significant correlations between different targets. 

The hourglass architecture has been also investigated in general (non-layered) hierarchical dependency networks,
similar to Evo-Lexis, by Sabrin and Dovrolis \cite{sabrin2016}. 
That analysis is based on identifying the core of a dependency network, 
as the minimum set of nodes that cover at least a fraction $\tau$ of all source-to-target dependency paths.
We have adopted that approach, as well as the hourglass metric proposed in \cite{sabrin2016}.
Their study shows the presence of the hourglass property in various technological, natural and information 
systems. The authors also present a model called \emph{Reuse-Preference}, capturing the bias of 
new modules to reuse intermediate modules of similar complexity instead of connecting directly 
to sources or low complexity modules. 

Despite this prior work, the interplay between the emergence of hourglass architectures and cost optimization 
in hierarchical networks has not been explored in previous research. Evo-Lexis identifies the conditions 
under which the hourglass property emerges in optimized dependency networks.

\subsection{Interplay of Design Adaptation and Evolution}
A main theme in our study is the interplay between changes in the environment (the targets that the system has to support) and the internal architecture of the system. 

Bakhshi et al. investigate a network topology design scenario in which the goal is to design a valid communication network between a set of nodes \cite{bakhshi2012}. The authors formulate and compare the consequences of two different optimization scenarios for that goal: \emph{incremental design} in which the modification cost between the two last snapshots of the design is minimized, and \emph{optimized design} in which the total cost of the network is minimized in every increment. Focusing on the case of ring networks, even though the incremental designs are more costly, the relative cost overhead is shown to not increase as the network grows. In a follow-up study, focused on mesh networks, the same observation is made and further, the incremental design is shown to be producing larger density, lower average delay and more robust topologies \cite{bakhshi2013}. 

Incremental design approaches are also considered in other contexts, such as in deep neural networks (DNNs).
Specifically, an important problem in machine learning is how to transfer learned features  
of a deep network from one task to another \cite{dnn}. 
Transfer learning can be considered analogous to the way in which new targets are added in an Evo-Lexis hierarchy:
new targets (output functions) are incrementally included in the Lexis-DAG (incrementally learned), by 
re-using previously constructed intermediate nodes (features of intermediate complexity)
and then optimizing the part of the DAG between those nodes and the new targets (learning the weights between
the existing features and the new outputs). 

The incremental design policies that we consider in this paper are studied in computer science under the 
umbrella of \emph{online algorithms} \cite{online-algs-thesis}: 
an online algorithm finds a sequence of solutions 
based on the inputs it has seen so far, without knowing the entire input sequence in advance. 
The main emphasis of research in online algorithms is to perform \emph{competitive analysis}, i.e., 
to derive worst-case theoretical bounds between of the quality (or cost) of the solution 
of an online algorithm relative to its offline counterpart that knows the entire 
input sequence \cite{borodin}. 
The Incremental Design approach in Evo-Lexis is an online algorithm but our focus is quite different: 
we compare empirically the cost and topological structure of the hierarchies produced by incremental design relative
to an optimized (``clean-slate'') algorithm that designs a minimum-cost hierarchy for the input sequence
that has been seen so far.


\subsection{From abstract modeling to specific evolving systems}
The Evo-Lexis model is a quite general and abstract model and it does not attempt to capture any domain-specific aspects of biological or technological evolution. As such, it makes several assumptions that can be criticized as unrealistic, such as that all targets have the same length, their length stays constant, the fitness of a sequence is strictly based on its hierarchical cost, etc.
We believe that such abstract modeling is still valuable because it can provide insights about the qualitative properties of the resulting hierarchies under different target generation models.
Having said that however, we also believe that the predictions of the Evo-Lexis model should be tested using real data from evolving systems in which the outputs can be well represented by sequences. 

One such system is the iGEM synthetic DNAs dataset \cite{igem}. The target DNA sequences in the iGEM dataset are built from standard ``BioBrick parts'' (more elementary  DNA sequences) that collectively form a library of synthetic DNA sequences. These sequences are submitted to the Registry of Standard Biological Parts in the annual iGEM competition. Previous research in \cite{dnaSynthPaper,siyari2016a} has provided some evidence that these synthetic DNA sequences are designed by reusing existing components, and as such, it has a hierarchical organization. In ongoing work, we investigate how to apply the Evo-Lexis framework in the timeseries of iGEM sequences, and whether the resulting iGEM hierarchies exhibit the same qualitative properties we observed in this study through abstract target generation models. 

%% file: 09-conclusion.tex
\section{Conclusion}
\label{evolexis-conclusion}
We presented Evo-Lexis, an evolutionary framework for modeling the interdependency between an incrementally designed hierarchy and a time-varying set of output functions, or targets, constructed by that hierarchy.
We leveraged the Lexis optimization framework, 
proposed in earlier work \cite{siyari2016a}, 
which allows the design of an optimized hierarchical
network for a given set of sequences.

We developed the optimization framework, evolutionary target generation processes, and evaluation metrics needed to study the emergence and evolution of optimized hierarchies. We summarize the results of our study as follows:
\begin{enumerate}
\item
Tinkering/mutation in the target generation process is found to be a strong initial force for the emergence of low-cost and deep hierarchies. The presence of selection, however, intensifies these properties of the emergent hierarchies.
\item
Selection is also found to enhance the emergence of more complex intermediate modules in optimized hierarchies.
The bias towards reuse of complex modules results in an hourglass architecture in which almost all source-to-target dependency paths traverse a small set of intermediate modules. 
\item
The addition of recombination in the target generation process is essential in providing target diversity in optimized hierarchies. 
\item
Hourglass-shaped optimized hierarchies are found to be fragile if the core nodes (i.e. nodes with highest centrality) are perturbed, similar to the concept of removal of hub nodes in scale-free networks.
\item
We show that an hourglass architecture introduces a trade-off between the cost of introducing new targets and the diversity between selected targets: hourglass architectures are evolvable in the sense that they allow the introduction of new targets at a low cost but they only explore a small part of the ``phenotypic space'' of all possible targets. These are targets that can be constructed at a low cost reusing the larger intermediate modules in the hierarchy.
\item
Our results suggest the existence of major transitions and punctuated equilibria in the evolutionary trajectory of hourglass-shaped hierarchies. The ``extinction'' of central modules is found to be the main factor behind this effect.
\item 
The comparison between incremental design and clean-slate shows that although the former is much more constrained, it has similar cost and it also exhibits the hourglass effect under the proposed evolutionary scenarios. Despite these similarities, each of these design policies results in a very different set of core modules.
\end{enumerate}


\section*{Acknowledgements}
This research was supported by the
National Science Foundation under Grant No. 1319549. We would also like to thank Matthias Gall\'e for his  comments.